\documentclass[a4paper,review]{cas-dc}

\usepackage[numbers]{natbib}
\def\tsc#1{\csdef{#1}{\textsc{\lowercase{#1}}\xspace}}
\tsc{WGM}
\tsc{QE}
\usepackage[utf8]{inputenc}
\usepackage{siunitx}
\usepackage{appendix}

\usepackage{soul}

\usepackage{mathtools}
\usepackage{amsmath,amsfonts,amsthm,amssymb}
\usepackage{empheq}
\usepackage[ruled,vlined]{algorithm2e}
\usepackage{mathalfa, eufrak, yfonts, mathrsfs}

\DeclareMathOperator*{\argmax}{argmax}

\theoremstyle{definition}

\theoremstyle{definition}
\newtheorem*{definition*}{Definition}%[section]

\renewenvironment{proof}{{\bfseries Proof.}}{\qed}

\usepackage{stmaryrd}
\usepackage[ruled,vlined]{algorithm2e}

\theoremstyle{definition}

\newtheorem*{example*}{\small{Example}}
\usepackage{graphicx}
\graphicspath{ {./img/} }
\usepackage{xcolor}
\usepackage{comment}
\usepackage{pgfplots}
\usepackage{tikz}
\usetikzlibrary{decorations.pathreplacing}
\usetikzlibrary{positioning}
\usetikzlibrary{arrows}
\usetikzlibrary{arrows.meta}
\usetikzlibrary{shapes.geometric}
\usetikzlibrary{shapes.misc}
\tikzset{cross/.style={cross out, draw=red, minimum size=40*(#1-\pgflinewidth), inner sep=0pt, outer sep=0pt},
	cross/.default={1pt}}

\makeatletter
\newcommand{\osymbol}[1]{\mathbin{\mathpalette\make@circled#1}}
\newcommand{\make@circled}[2]{
	\ooalign{$\m@th#1\smallbigcirc{#1}$\cr\hidewidth$\m@th#1#2$\hidewidth\cr}
}
\newcommand{\smallbigcirc}[1]{
	\vcenter{\hbox{\scalebox{1}{$\m@th#1\bigcirc$}}}
}
\makeatother

\usepackage{pdfpages}
\usepackage{graphicx}
\usepackage[export]{adjustbox}
\usepackage{subcaption}

\begin{document}
\let\WriteBookmarks\relax
\def\floatpagepagefraction{1}
\def\textpagefraction{.001}

\shorttitle{Decentralized cooperative perception for autonomous vehicles: Learning to value the unknown}    

\shortauthors{Chaveroche et al.}  

\title [mode = title]{Decentralized cooperative perception for autonomous vehicles: Learning to value the unknown}  

\tnotemark[1] 

\tnotetext[1]{This work was carried out and co-funded in the framework of the Labex MS2T and the Hauts-de-France region of France. It was supported by the French Government, through the program ``Investments for the future'' managed by the National Agency for Research (Reference ANR-11-IDEX-0004-02).} 


\author[1]{Maxime Chaveroche}[orcid=0000-0002-0834-4022]

\cormark[1]


\ead{maxime.chaveroche@gmail.com}

%\ead[url]{<URL>}

\credit{Conceptualization, Formal analysis, Investigation, Methodology, Software, Data Curation, Validation, Visualization, Writing - Original Draft, Writing - Review \& Editing}

\affiliation[1]{organization={Alliance Sorbonne Universit\'e, Universit\'e de technologie de Compi\`egne, CNRS, Heudiasyc},
%            addressline={57, avenue de Landshut}, 
%            city={Compi\`egne},
%          citysep={}, % Uncomment if no comma needed between city and postcode
            postcode={CS 60319 - 60203 Compi\`egne Cedex}, 
            state={},
            country={France}}

\author[1]{Franck Davoine}[orcid=0000-0002-8587-6997]

% Footnote of the second author

% Email id of the second author
\ead{franck.davoine@hds.utc.fr}

%\ead[url]{hh}

\credit{Supervision, Writing - Review \& Editing}

\author[1]{V\'eronique Cherfaoui}[orcid=0000-0003-2064-9838]

% Footnote of the second author

% Email id of the second author
\ead{veronique.cherfaoui@hds.utc.fr}

%\ead[url]{hh}

\credit{Supervision, Writing - Review \& Editing}

\cortext[1]{Corresponding author}



\begin{abstract}
		Recently, we have been witnesses of accidents involving autonomous vehicles and their lack of sufficient information. One way to tackle this issue is to benefit from the perception of different view points, namely cooperative perception. 
%While setting extra pieces of road infrastructure to help autonomous vehicles could be imagined, this would require a lot of investments, limit its usage to some areas in the world and even give rise to delays when centralizing perceptions. 
%Talking about centralized collaborative perception in particular, this also features the disadvantage of making the agents broadcast their entire perception, which can be heavy on the means of communication and computation and give rise to delays. 
We propose here a decentralized collaboration, i.e. peer-to-peer, in which the agents are active in their quest for full perception by asking for specific areas in their surroundings on which they would like to know more. %, instead of always broadcasting everything. 
Ultimately, we want to optimize a trade-off between the maximization of knowledge about moving objects and the minimization of the total volume of information received from others, to limit communication costs and message processing time.
%		
%		we tackled the issue of redundancy and irrelevance in decentralized collaborative perception. For this, 
For this, we propose a way to learn a communication policy that reverses the usual communication paradigm by only requesting from other vehicles what is unknown to the ego-vehicle, instead of filtering on the sender side. We tested three different generative models to be taken as base for a Deep Reinforcement Learning (DRL) algorithm, and compared them to a broadcasting policy and a policy randomly selecting areas. More precisely, we slightly modified a state-of-the-art generative model named Temporal Difference VAE (TD-VAE) to make it sequential. We named this variant Sequential TD-VAE (STD-VAE). We also proposed Locally Predictable VAE (LP-VAE), inspired by STD-VAE, designed to enhance its prediction capabilities. We showed that LP-VAE produced better belief states for prediction than STD-VAE, both as a standalone model and in the context of DRL. The last model we tested was a simple state-less model (Convolutional VAE). Experiments were conducted in the driving simulator CARLA, with vehicles exchanging parts of semantic grid maps.
Policies learned based on LP-VAE featured the best trade-off, as long as future rewards were taken into account. 
%However, for our specific problem, the best communication policy was found with only immediate rewards, in a greedy way. Doing so, no prediction capability was needed. Furthermore, the need for a recognition model was alleviated by the use of a memory module that we designed using DST to augment each observation. 
Our best models reached on average a gain of 25\% of the total complementary information, while only requesting about 5\% of the ego-vehicle's perceptual field. We also provided interpretable hyperparameters controlling the reward function, which makes this trade-off adjustable (e.g. allowing greater communication costs).
\end{abstract}


\begin{highlights}
\item Provides a way to learn an efficient decentralized communication policy between autonomous vehicles
\item Proposes a new generative model that learns to build state representations for RL through prediction and reconstruction
\item Proposes a reward function with interpretable parameters to adjust the trade-off between information gain and volume
\item With our experiment parameters, achieved 25\% gain in relevant information, with only 5\% of the total queryable volume
\end{highlights}

% Keywords

\maketitle

% Main text
\section{Introduction}

Recently, we have been witnesses of accidents involving autonomous vehicles and their lack of sufficient information at the right time. One way to tackle this issue is to benefit from the perception of different viewpoints, namely collaborative perception. While setting a multitude of sensors in the road infrastructure could be imagined, this would require a lot of investments and limit its usage to some areas in the world. Instead, we focus on the exchange of information between vehicles about their common environment, where they are the only sources available. 
%We will distinguish two approaches:
%(i) centralized
%and (ii) decentralized.

These communications can simply be centralized by a server that would gather all information from all vehicles to process it and re-distribute it to all, as suggested in \cite{chen2019f}.
% popular \cite{bressonsurvey17}:
%	,saeedisurvey15, cadenasurvey16}:
%gathering all information from every agent, fusing it to build a single perception and then re-distributing it to all agents.
However, this still consists of Vehicle-to-Infrastructure (V2I) communications, which implies (1) an infrastructure cost and the impossibility to share information with other agents when there is no server available nearby. 
%even if it is less pronounced than when setting sensors in the infrastructure.
It also features the disadvantage of (2) making the agents broadcast their entire perception, which can be heavy on the means of communication and computation and give rise to delays. 

In contrast, the decentralized Vehicle-to-Vehicle (V2V) approach \cite{
	%kim16, 
	kim15, 
	li14, 
	zoghby14, 
	%	rauch13, 
	%rauch12, 
	seeliger14a, 
	%seeliger14b,
	%saeedi11, 
	vasic2016system} 
does not require any extra infrastructure to work, i.e. does not implies (1). In this setting, agents directly exchange pieces of information between them. 
It also comes with new problems such as data incest and lower computation capabilities. We will ignore them here as we already tackled the issue of avoiding data incest using Dempster-Shafer Theory (DST) \cite{shafer76} in spite of low computation capabilities with two conference papers \cite{me_gretsi,me} and a journal paper \cite{me_journal}. But V2V communications bring a potentially heavier communication burden as well, due to redundancies. In fact, (2) is worse in this setting than in the centralized one if agents are passive, meaning if they simply broadcast their perception for the others to know, without filtering it beforehand.
Nevertheless, this decentralized approach offers the possibility to make the agents active in their quest for full perception, i.e. making the agents ask for specific areas in their surroundings on which they would like to know more, instead of always broadcasting everything. This is impossible in the centralized setting, as the server decides and thus needs to gather all perceptions beforehand.

Here, we propose such a system, where each agent builds its own local top-down semantic grid and sends specific requests to others in the form of bounding boxes described in the global reference frame. We choose local grid maps for their ability to map an agent's knowledge and to deduce its uncertainties in space.

\def\x{3}
\def\xoffset{0.75}

\section{Related Works}

Since not all uncertain areas are relevant, Active Exploration \cite{stachniss2005information, clemens2016evidential} is not enough; a truly efficient collaboration policy requires some understanding of the scenery \cite{wang2019semantic}, extracted from the spatial arrangement of grid cells and their classes. 
What could lie in the shadows and how to best discover it? If a pedestrian is heading towards an occluded area, we expect the agent to request for this area, as a tracking system. If the agent has no idea of what could be in the unknown, maybe it could ask for some key points to understand the layout of the environment. If an area on the road is near a crowd of people or in the continuity of a pedestrian crossing, ask for it as some unseen-before pedestrians could be crossing, etc. More generally, we would like the agent to know as much as possible about moving objects in its vicinity, while avoiding to request too much information from others.
This represents a complex bounding box selection policy to be learned from pixels. 

Given the long-lasting successes of Deep Learning in such ordeals, it seems natural to consider neural networks for our problem. But, while it is theoretically possible (but practically challenging) to learn our policy in an end-to-end fashion with Model-free Deep Reinforcement Learning (DRL), we choose to first learn a deep generative model to pre-process our inputs. Indeed, training deep neural networks is easier, faster and more stable when the loss on the output is in the form of a well-justified derivable function, which is hard to achieve with reward signals from a RL environment. Building this generative model also allows for more control and insights on what is learned, and reduces the size of the neural networks that are supposed to be trained through model-free DRL. As demonstrated in World Models \cite{ha2018recurrent}, 
%which proposes a system composed of a recurrent generative model and a lightweight controller that learns a policy using this model, 
learning a policy on top of a model can even be achieved with simple heuristics such as Evolution Strategies (ES), with performances equivalent to RL algorithms.

Our model needs to be generative, for inference in unknown areas. In addition, we want it to be predictive, in order to make it understand latent dynamics, anticipating disappearances or inferring hidden road users from the behavior of visible ones. Doing so, it could even eventually compensate for communication latencies.  Such a model would be useful in itself for other tasks as well, e.g. autonomous driving.

%So, decompose our task into 
%(a) the construction of a world model that is both generative, for inference in unknown areas, and predictive, to make it understand latent dynamics, anticipating disappearances or inferring hidden road users from the behavior of visible ones. even eventually compensate for communication latencies and to better model road interactions, and 
%(b) the learning of a bounding box selection policy based on this internal representation. Proceeding with (a) yields a model, which gives more insights and control on what the network is learning, as well as predictions that can be used to learn the policy in (b). Then, as demonstrated in World Models \cite{ha2018recurrent}, which proposes a system composed of a recurrent generative model and a lightweight controller that learns a policy using this model, (b) can be solved with simple heuristics such as Evolution Strategies (ES), instead of some fairly unstable and sample inefficient Reinforcement Learning algorithm.
%This decomposition allows for the exploitation of the world model for other usages in the autonomous vehicle, more insights and control on what the network is learning, less work for the network learning this policy (which is fairly unstable and sample inefficient with a RL algorithm) and finally gives the ability to use the prediction/generation loss as part of the reward function, since this internal world model is in the end what we try to optimize by cooperating with other vehicles. 

Several existing works \cite{
	wirges2018evidential,
	sugiura2019probable,
	hoermann2018dynamic,
	everett2019planning,
	shrestha2019learned} 
employed generative models with convolutional networks in a U-Net architecture in order to augment instantaneous individual grid maps. Some used deterministic networks such as Generative Adversarial Networks (GAN). Others tried to incorporate stochasticity with Monte Carlo Dropout or simply using a Variational Auto-Encoder (VAE). 
Most used occupancy grids as input, but some chose semantic grid maps or DOGMa (occupancy grid with velocities). These inputs were either expressed in a static global reference frame or given to a system that had no prediction capability. Doing so, it appears that none of these approaches really modeled the long-term dynamics of the environment that would be necessary to learn our desired policy. On the other hand, a kind of recurrent generative model inspired by the VAE, namely Temporal Difference VAE (TD-VAE) \cite{gregor2018temporal}, was designed with the specific intent of being taken as base for a reinforcement learning algorithm. It puts an emphasis on the learning of belief states for long-term predictions, which are important for the development of complex strategies. It has been proven in \cite{shaping_belief} that explicitly predicting future states enhances data-efficiency in a number of RL tasks, though they train their model jointly with the policy and do not use the loss defined in \cite{gregor2018temporal}.
Appealed by the theoretical justifications of TD-VAE, its decoupling regarding specific RL tasks (which simplifies the search for good RL hyperparameters) and its demonstrated ability to predict plausible sequences of images in a 3D world at different time horizons and from a variable number of observations, we have implemented and adapted this TD-VAE to our problem. However, correcting some of its weaknesses regarding its actual prediction capability, we finally proposed our own model, called Locally Predictable VAE (LP-VAE).
To learn our communication policy based on this model, we chose the widely used Proximal Policy Optimization (PPO) algorithm \cite{ppo}, which is a fairly stable and simple policy-gradient based DRL algorithm with few hyperparameters. 

Closely related to our goal, other works try to address the problem of efficiently communicating between autonomous vehicles. 
In \cite{v2vnet},
%Tries to maximize transmitted info while minimizing message weight
they used a joint Perception and Prediction (P\&P) model that transforms sensor data into learned features to broadcast to other vehicles. This model also fuses received features with local ones and tries to predict the trajectory of nearby communicating vehicles.
This information compression is also present in our work in the form of a Convolutional VAE preprocessing each observation grid. We go one step further in communication efficiency as our system does not broadcast every piece of information, but chooses instead which one it wishes to receive.
Sending learned features also forces them to make another neural network learn to spatially and temporally transform all pieces of information received from the vehicular network. Even the fusion operation is done by making a neural network learn how to fuse two learned features, without any guarantee on the result. Instead, here we rely on top-down semantic grids, which are simple discretizations of the space around the ego-vehicle. Doing so, we can transform the content of our transmissions using linear transformations. Furthermore, our system keeps its integrity by only fusing probability distributions.

In \cite{drl_vanet}, they used Deep Reinforcement Learning to select only a portion of the perceptive field of an autonomous vehicle to send to others. However, this information filtering is done on the sender side, contrary to our approach that filters on the receiver side. Doing so, their approach still consists in broadcasting pieces of information, regardless of the actual needs of others.

The same can be stated for \cite{higuchi2019value}, where they describe a V2V cooperative perception system in which vehicles exchange object detections. They try to reduce redundancies by estimating the value of a piece of information for a potential receiver. The value here is the novelty, i.e. the probability that the potential receiver is not aware of some object of interest.

%The choice of ES over RL is explained by the fact that it is much simpler to implement, more stable and faster when using to optimize up to a few thousand parameters (even though it is even more sample-inefficient). It has also been proven in World Models that ES is able to reach performance comparable to state-of-the-art RL algorithms.

Section \ref{com:problem} formally introduces our communication problem, justifying the use of a preprocessing generative model.
%Section \ref{com:background} provides some background on PPO and explains why we chose this RL algorithm to learn our communication policy. 
Section \ref{models} formalizes the aforementioned generative model, introducing TD-VAE and LP-VAE. Section \ref{application} presents our deep networks implementing these models. 
%Section \ref{decision-process} formalizes the decision process for which we build our policy. 
Then, section \ref{exp} evaluates and compares the performance of different versions of our models and policy learnings. Finally, we conclude this article with section \ref{com:conclusion}.

\section{Problem formulation}\label{com:problem}

\begin{figure*}[ht]
	\includegraphics[scale=0.31]{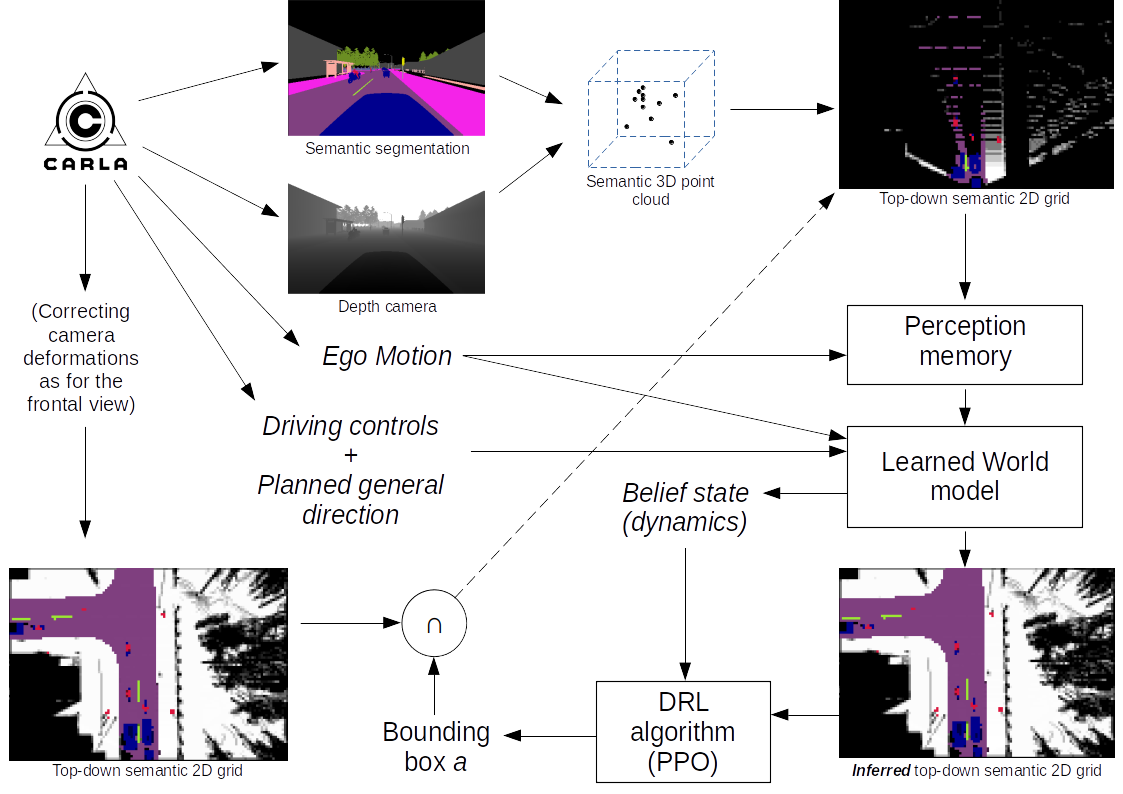}
	\centering
	\caption{Illustration of our application. CARLA provides a semantic segmentation corresponding to a camera attached to the ego-vehicle hood, as well as its corresponding depth (images taken from \cite{carla}). This gives us enough information to create a semantic 3D point cloud, i.e. to scatter all pixels in space according to their depth and image coordinates (and the camera deformation). From it, we project these pixels back into a 2D plane (i.e. a grid), but from a top-down point of view (and without camera deformations). In parallel, we get the ego-vehicle motion since the previous time step in order to update a \textit{perception memory} containing 2D points from previous time steps. We add the current semantic grid to this memory and give the resulting augmented grid to our \textit{learned world model} (STD-VAE or LP-VAE), along with the ego-motion and driving policy commands. In turn, this model tries to guess what is hidden in occluded areas and provides a belief state about latent dynamics. These outputs are then given to a \textit{DRL algorithm} that chooses a grid area to request to the world. This area is extracted at the next time step from a grid generated by a camera above the ego-vehicle. Finally, this information is fused at the next time step with the ego-vehicle perception.
	}
	\label{fig:overview}
\end{figure*}

We formulate our communication problem as a Markov Decision Process (MDP). Fig. \ref{fig:overview} gives an overview of it, working with the driving simulator CARLA \cite{carla} for our experiments.

\subsection{State space}\label{com:state}

\begin{figure}[t]
	\includegraphics[scale=0.3]{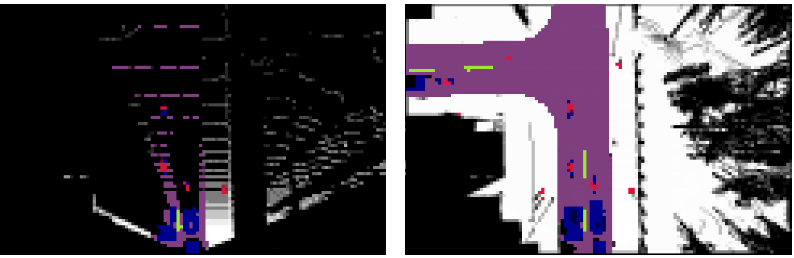}
	\centering
	\caption{\textbf{Left:} Illustration of an instance of top-down semantic grid $G_t$ corresponding to a partial observation $x_t$ in our model. Red is for \textit{pedestrians}, blue is for \textit{cars}, yellow is for \textit{road lines}, purple is for \textit{road}, white is for \textit{other} and black is for ignorance. The displayed class is the one with the greatest mass. The intensity of its color depends on its mass: the closer to 0, the darker.
		Notice all the occlusions due to walls or other road users, in addition to the limited distance of perception of the ego-vehicle. \textbf{Right:} Instance of top-down semantic grid corresponding to a complete observation $y_t$ in our model. Actually, this view is obtained with a facing ground camera above the ego-vehicle. Doing so, it contains itself some occlusions due to trees, poles, buildings, etc. Thus, it is rather a hint about the true $y_t$. This view can also be obtained by the fusion of multiple view points, from autonomous vehicles or infrastructure sensors.
	}
	\label{grid}
\end{figure}

We assume the existence of a driving policy from which we only know the actions taken at each time step: ego-vehicle controls (acceleration and steering angle, each ranging in $[-1, 1]$) and global direction (average of the next 10 equally-spaced points the planner set to visit in meters relative to the ego-vehicle's reference). This driving policy influences the road environment in which the ego-vehicle is moving. This is not the case with the communication environment that we consider in this MDP. 
Each observation is a tuple $(G_t, C_t, V_t)$, where $G_t$ is an ego-centered semantic grid, $C_t$ represents the actions taken by the driving policy at a given instant $t$ (which influence $G_{t+1}$) and $V_t$ is the motion of the ego-vehicle between $t-1$ and $t$. Each semantic grid $G_t$ is a top-down 6-channels pseudo-Bayesian mass grid corresponding to the five classes of the frame of discernment $\Omega = \{ \textit{pedestrian}, \textit{car}, \textit{road lines},\textit{road}, \textit{other} \}$. The class \textit{car} actually contains any type of vehicle, even bikes. The class \textit{road lines} contains any road marking: road lines, arrows, painted stop signs, etc. The class \textit{other} contains the rest of the static objects perceivable by the agent, such as vegetation, sidewalks, buildings, etc. The last channel represents ignorance, i.e. the mass put on $\Omega$. This means that $G_t \geq 0$ and, for any cell index $i$ of $G_t$, we have $\sum_{k=1}^{6} G_t[i][k] = 1.$ These cells are distributed as a matrix (grid) of 80 rows and 120 columns, i.e. $G_t$ is analog to a $80\times120\times6$ image of values in $[0, 1]$.
%Recall that the contour function \text{pl} is defined by $$\forall \omega \in \Omega,~ \text{pl}(\omega) = \text{Pl}(\{\omega\}),$$ where Pl is a plausibility function. 
See Fig. \ref{grid} for a visualization of this semantic grid.

These observations constitute a very large and complex space which would be hard to transform into exploitable neural network features without a derivable loss function. Thus, we will first build a generative model of the driving environment (implicitly including the agent's driving policy). Besides, learning this model beforehand will give us more control on the information flow that should be considered by the communication policy. 
Therefore, the state space of our MDP is made of learned features from this generative model. Several versions of this generative model are proposed in Section \ref{models}.

% represented by the latent states $Z$ introduced in the last section. However, we consider the inference of these controls implicit in predictions. So, the latent dynamics of the ego-vehicle itself is supposed to be contained in $Z$ as well as the ones of the other agents. We only give these controls as part of the belief state $B_t$ for the inference of $Z_t$. 
%There are $2.N+2$ continuous actions: $N$ ground 2D-points indicating the next $N$ locations planned by the local planner, expressed relatively to the ego-vehicle, and two immediate control commands that are the acceleration and steering angle, each ranging in $[-1, 1]$.

%\subsection{Decision Process}\label{decision-process}

%Predicting future perception grids is essential to find a suitable policy as there are communication latencies, putting a delay between the choice of a bounding box and its effects in the environment (which is the perception grid itself).
%To ease the prediction process, we make the assumption that each state follows the Markov property, i.e. only depending on the previous one.
%Following our PO-DTMC modeling of section \ref{ai-model}, we formulate our decision process as a partially observable Markov decision process (POMDP). 
%This means that our model will try to infer what is unknown at each time step both in terms of dynamics and occlusions.
%One way to achieve this and to ease the prediction process is to model our problem as a partially observable Markov decision process (POMDP), so that the latent state at some instant is only dependent from the latent state at the previous time step.

\subsection{Action space}
Our MDP has 4 continuous actions that each ranges in $[0, 1]$, defining a bounding box in the local grid $G_t$ of the ego-vehicle at time $t$:
%, where $D$ is the number of timesteps delaying the effect of the choice of a bounding box: 
width, height, column and row. 
This bounding box is supposed to represent an area in the ego-vehicle's future surroundings.
%, in global coordinates so that other agents can transmit their information about this area.
%In parallel, the relative pose (i.e. motion and orientation) of the ego-vehicle is estimated by a simple Multilayer perceptron (MLP) taking as input $Z_{t-1}$ and $Z_t$, in order to be able to describe the outputted future bounding box $A_{t+1}$ in the reference of the ego-vehicle at time $t$. It is assumed that the ego-vehicle has access to its global coordinates, which can thus be used to describe this bounding box in the global reference. 

\begin{figure}[t]
	\includegraphics[scale=0.3]{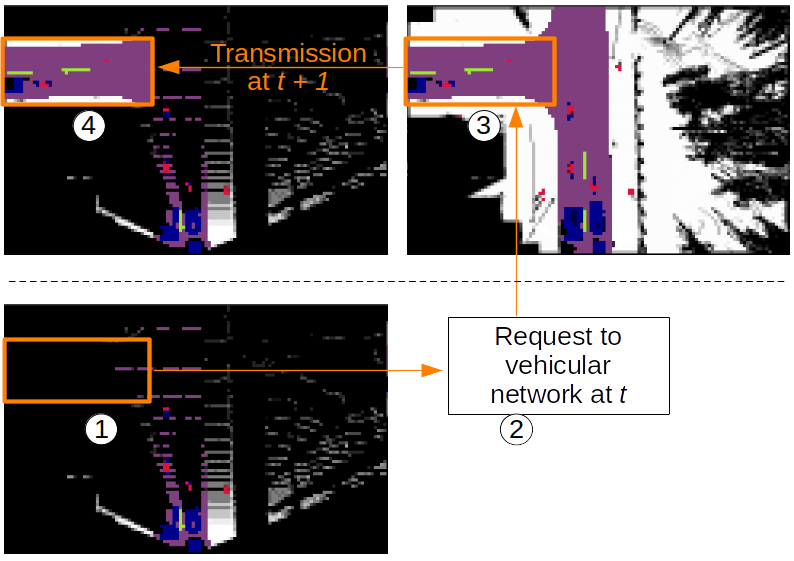}
	\centering
	\caption{Illustration of our decision process: 1) Based on what is known at time $t$, select a bounding box where there is high uncertainty and high probability to discover road users. 2) Send this request in global coordinates to the vehicular network (which may consists of both infrastructure sensors and other autonomous vehicles). 3) At time $t+1$, we expect some vehicles to transmit their perception of this area. In our implementation, complete perceptions are simply obtained by a camera above the ego-vehicle since we focus on the selection of bounding boxes, i.e. 1). 4) The transmitted partial perception is fused with the one of the ego-vehicle at time $t+1$.
	}
	\label{fig:decision_process}
\end{figure}

\subsection{Transition function}

Transitions from a state-action pair to a new state depend also on the driving environment, i.e. CARLA. First, this environment generates a new partial grid $G_{t+1}$ and other observations already described.
The bounding box described by the action given at time $t$ is then translated into an area of $G_{t+1}$ filled with complete information. Fig. \ref{fig:decision_process} illustrates this process.

In addition, a visual memory mechanism, specific to our MDP, makes perceptions persist for a few time steps, discounted a little more every time. This implements short-term memory, so that we only consider as unknown what has not been perceived in a long time (or never). This also has the effect of giving consequences to past actions, since bounding boxes in the same area will have close to no potential information gain for a few time steps.

\newcommand{\plG}[1]{\sloppy\text{$G^{\textit{Pl}}_{#1}$}}
\newcommand{\pl}[1]{\sloppy\text{$\textit{Pl}_{#1}$}}
\newcommand{\plv}[2]{\sloppy\text{$\pl{#1}\left(#2\right)$}}
\newcommand{\plnorm}[1]{\sloppy\text{$\overline{{Pl}^c_{#1}}$}}
\newcommand{\plnormv}[2]{\sloppy\text{$\plnorm{#1}\left(#2\right)$}}

\subsection{Rewards}
Finally, let us define a reward function for our MDP.
%(also called \textit{fitness function} for ES) 
%to be maximized by our policy. 
Let $r_t$ be a reward density, defined for each cell $i$ of $G_{{t+1}}$ as:
\begin{align}\label{reward_density}
	r_t(i) &= -\eta . r_{\min} ~+~ S[i] . \displaystyle\sum_{k=1}^{5} r_{\textit{obj}}[k].~ \max\left(0,~\vphantom{G_{{t+1}}[i][k] - \widetilde{G}_{{t+1}}[i][k]}\right.\nonumber\\ &\quad\qquad\qquad\qquad\qquad \left.G_{{t+1}}[i][k] - \widetilde{G}_{{t+1}}[i][k]\right)^w
\end{align}
%\begin{align*}
%r_t(i) = G_{{t+1}}[i][6] . \displaystyle\sum_{k=1}^{5} r_{\textit{obj}}[k].~ G^M_{{t+1}}[i][k],
%\end{align*}
%\begin{align*}
%r_t(i) = \displaystyle\sum_{\substack{S\subseteq \Omega\\S\neq \emptyset}} \underset{s\in S}{\mathbb{E}}\left[r_{\textit{obj}}(s)\right].~ m_{{t+1}}[i]\left(S\right)~.\log_{|\Omega|} \left(\frac{|\Omega|}{|S|}\right),
%\end{align*}
where $w \in \mathbb{R}^{+*}$, $\eta\in [0,1]$ and $\widetilde{G}_{{t+1}}$ is the grid before fusion with the grid $G^M_{t+1}$ corresponding to $M_{t+1}$. The quantity $r_{\textit{obj}}$ is a nonnegative reward per object pixel (only null for the static class \textit{other}, i.e. $r_{\textit{obj}}[5]=0$) such that $r_{\textit{obj}}[k] \geq r_{\textit{obj}}[k+1]$. Indeed, pedestrian are the smallest identifiable objects among our classes and so must have the highest reward per pixel. The quantity $r_{\min}$ is equal to the least positive reward per pixel, i.e. $r_{\min} = r_{\textit{obj}}[4]$. It is used to discourage the selection of uninteresting cells. The coefficient $\eta$ that multiplies it represents the minimum informational gain that is needed to consider this cell worth to be requested. For some value of $\eta$, this minimum gain applies to the class with the least reward, while it becomes virtually more and more forgiving as the class has a greater reward per cell.
%The exponent $w \in \mathbb{R}^{+*}$ controls the weight of this gain in mass on the reward density. 
Moreover, notice that $\max(0,~ G_{{t+1}}[i][k] - \widetilde{G}_{{t+1}}[i][k]) \in [0, 1]$, which implies that $\max(0,~ G_{{t+1}}[i][k] - \widetilde{G}_{{t+1}}[i][k])^w \in [0, 1]$. This means that $w$ only alters the significance of some gain in mass: for $w\in(0,1)$, $\max(0,~ G_{{t+1}}[i][k] - \widetilde{G}_{{t+1}}[i][k])$ will be greater than for $w=1$, while for $w\in(1,+\infty)$, $\max(0,~ G_{{t+1}}[i][k] - \widetilde{G}_{{t+1}}[i][k])$ will be less. In other words, if $w\in(1,+\infty)$, then the gain will have to be more important to have an impact on $r_t(i)$.
Finally, $S$ represents a spatial filter to account for the fact that we are not equally interested everywhere in discovering road users. For example, a road user very far ahead is not as valuable an information as a road user just around the corner. We defined a forward filter $S_F$ and a lateral filter $S_L$, such that $S = S_F . S_L$. We set
\begin{align*}
	S_F[i] = 1 - \left[\frac{\beta_F}{1-\alpha} . \max\left(0,~ \frac{F(i)}{\max(F)} - \alpha\right)\right]
\end{align*}
where $\alpha \in [0,1)$ and $\beta_F \in [0,1]$. The quantity $F(i)$ is the forward distance (number of rows from the row in which the center of the ego-vehicle is) corresponding to cell $i$. The greater the parameter $\beta_F$, the less the farest cells are valued. The greater the parameter $\alpha$, the farer from the ego-vehicle the decrease in value starts. 

\begin{figure}[t]
	\includegraphics[scale=0.4]{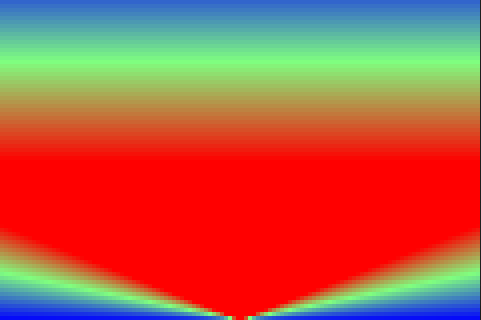}
	\centering
	\caption{Heatmap illustrating our spatial filter $S$ for $\alpha=0.5$, $\beta_F=0.8$, $\beta_L=1$ and $\zeta=0.01$. Deep blue is 0, while deep red is 1, which means that the reward in a cell located in a blue region will be 0, no matter what is inside. The center of the ego-vehicle is in the middle of the first row starting from bottom.}
	\label{spatial-filter}
\end{figure}

The second filter is defined as
\begin{align*}
	S_L[i] = 1 - \frac{\beta_L}{\zeta} . \max\left(0,~ \zeta - \left|\cos\left(\text{arctan2}\left(L(i), F(i)\right)\right)\right|~\right)
\end{align*}
where $\zeta \in (0,1]$. The quantity $L(i)$ is the lateral distance (number of columns from the column in which the center of the ego-vehicle is) corresponding to cell $i$. This filter describes a cone in front of the ego-vehicle (and symmetrically at the back of it) in which the cells are the most valued. The greater the parameter $\zeta$, the narrower this cone. The greater the parameter $\beta_L$ is, the less the cells outside the cone (i.e. on the sides of the ego-vehicle) are valued. Fig. \ref{spatial-filter} provides a visualization of $S$.
%extracted from $Y_{t+1}$ according to $A_{t+1}$, i.e. the grid corresponding to $M_{t+1}$.

The reward associated with some action $a_t$ is defined as
% For some sub-grid $B$ of $G_{t+D}$ defined by the bounding box chosen at time $t$, we define our reward function as:
\begin{align}\label{reward}
	R_t(a_t) = - K. (1-\eta) . r_{\min} + &\displaystyle\sum_{i \in I(a_t)} r_t(i),
\end{align}
where $K$ is the minimum number of interesting cells that must be entirely discovered in order to make the request worthwhile, $I(a) = [v(a), v(a)+h(a)]\times[u(a), u(a)+w(a)]$ and $u(a)$, $v(a)$, $w(a)$, $h(a)$ are respectively the column index, row index, width and height indicated by some action $a$.

\subsubsection{Grid fusion}

In order to produce $G_{t}$ from $\widetilde{G}_t$ and the grid $G^M_t$ corresponding to $M_t$ in Eq. (\ref{reward_density}), we need to define a fusion procedure. As each cell $i$ in both $\widetilde{G}_t$ and $G^M_t$ is a mass function, we know that:
$$
G_t[i][6] = \widetilde{G}_t[i][6] ~.~ G^M_t[i][6],
$$ 
where 6 is the channel corresponding to the mass on $\Omega$.
Furthermore, we can get the contour functions of these pseudo-Bayesian mass functions simply by adding the mass on $\Omega$ to the mass on each of our 5 classes. Then, a simple pointwise multiplication of these two contour functions produces the contour function corresponding to $G_t$. This also implies a mass on $\emptyset$, which is caused by conflicting pieces of evidence between the two mass functions. Since we are not interested in this level of conflict, we choose to renormalize masses as in Dempster's combination rule \cite{dempster68}. Unlike Dempster's rule however, we only distribute this conflict on singletons $G_t[i][1:5]$ and keep the true value $G_t[i][6]$, as the distinction between ignorance and conflict is crucial to our communication policy. Algorithm \ref{algo:fusion} details this procedure.

\begin{comment}\label{algo:pltom}
	
	\KwIn{Contour function \textit{pl}}
	\KwOut{mass on singletons only $m$, masses on a chain $u$ of size $|\Omega|-1$}
	\vspace{0.25cm}
	
	$N \leftarrow \text{len(\textit{pl})}$\;
	$m \leftarrow \text{sort}(\textit{pl})$\;
	$u \leftarrow \text{zeros}(\textit{N-1})$\;
	$u_{\textit{acc}} \leftarrow 0$\;
	
	\For{$i = 0$ \KwTo $N-2$}{
		$s \leftarrow \sum_{j=0}^{N-1} m[j]$\;
		$x \leftarrow \max(0, \frac{s + u_{\textit{acc}}-1}{N-1-i})$\;
		$x \leftarrow \min(x, m[i])$\;
		$m[i:] \leftarrow m[i:] - x$\;
		$u[N-2-i] \leftarrow x$\;
		$u_{\textit{acc}} \leftarrow u_{\textit{acc}} + x$\;
	}
	\textbf{Return} $m, u$\;
	
	\caption{Computation of the least committed pseudo-consonant mass function corresponding to a contour function \textit{pl}.}
\end{comment}

\begin{algorithm}
	\KwIn{Two pseudo-Bayesian mass functions $m_1$, $m_2$}
	\KwOut{The fused mass function $m_{12}$}
	\vspace{0.25cm}
	
	$N \leftarrow \text{len($m_1$)}$\;
	$m_{12}[N] \leftarrow m_{1}[N] ~.~ m_{2}[N]$\;
	$m_{12}[1:N-1] \leftarrow (m_{1}[1:N-1] + m_{1}[N]) ~.~ (m_{2}[1:N-1] + m_{2}[N]) - m_{12}[N]$\;
	$s \leftarrow \text{sum} (m_{12}[1:N-1])$\;
	\If{$s > 0$}{
		$m_{12}[1:N-1] \leftarrow (1-m_{12}[N])~.~\frac{m_{12}[1:N-1]}{s}$\;
	}
	\textbf{Return} $m_{12}$\;
	
	\caption{Fusion procedure for two pseudo-Bayesian mass functions $m_1$ and $m_2$.}\label{algo:fusion}
\end{algorithm}

\section{Models}\label{models}

%\subsection{Model}\label{models}
In this section, we will present several versions of the generative model mentioned in section \ref{com:state}, namely STD-VAE and LP-VAE. In the end, this generative model will provide us with learned features describing the state of the environment related to the MDP presented in section \ref{com:problem}, in order to reduce the size of the network optimized through DRL and to control what is kept in the information flow. 
We will start by formalizing in section \ref{ai-model} a draft of this model that ignores the actions the agent takes at each time step. Then, we will briefly introduce in section \ref{td-vae} the original TD-VAE \cite{gregor2018temporal}. Following that, we will propose in section \ref{comm:stdvae} our sequential variant of TD-VAE, i.e. STD-VAE. Inspired by this model, we will then propose LP-VAE in section \ref{lp-vae}. Finally, section \ref{full-model} will demonstrate with LP-VAE how to modify this generative model to incorporate the actions chosen by the agent.

\subsection{Action-independent modeling}\label{ai-model}

As a vehicle clearly cannot access the complete state of its surroundings through its sole perception, we can model our problem as a 
Partially Observable Discrete-Time Markov Chain (PO-DTMC),
%Partially Observable Markov Decision Process (POMDP), 
where $X_t$ and $Z_t$ denote random variables representing respectively a partial observation and a latent state at time $t$. However, we consider that $Z_t$ and $X_t$ are in different spaces, the latent space 
%being a lower-dimensional space than the observation space, while 
describing the whole environment and containing information about object dynamics and trajectories allowing for predictions. More precisely, $X_t$ corresponds to the sole perception of the ego-vehicle at time $t$, without memory of the past. We also introduce a third random variable $Y_t$ which represents the spatially complete observation corresponding to $Z_t$ in the space of $X_t$. In other words, $X_t$ is a partial observation of $Y_t$ which is itself a partial observation of $Z_t$.

So, let $\theta$ be a set containing the parameters of a generative model that projects a latent state $Z_t$ onto the observation space as $(X_t,~ Y_t)$. We choose to implement this generative model as a deep neural network and we set the following Gaussian distributions as constraints, for numerical stability and simplicity: 
\begin{itemize}
	\item $Z_i \sim \mathcal{N}(0, I_d)$ 
	\item $p_{Z_{i+1}|Z_i}(\cdot | z_{t}; \theta) = \mathcal{N}(\mu_{z}(z_{t};\theta),~ \sigma^2_{z}(z_{t};\theta) . I_d)$
	\item $p_{Y_i|Z_i}(\cdot | z_{t}; \theta) = \mathcal{N}(\mu_{y}(z_{t}; \theta),~ \alpha_y . I_{|X_t|})$
	\item $p_{X_i|Y_i,Z_i}(\cdot | y_t, z_{t}; \theta) = \mathcal{N}(\mu_{x}(y_t, z_{t};\theta),~ \alpha_x . I_{|X_t|})$
\end{itemize}
where $\mu_{z}$, $\sigma_{z}$, $\mu_{x}$ and $\mu_{y}$ are all deep neural networks taking their parameters in $\theta$, where $d$ is an arbitrary number of dimensions for $Z_t$, where $z_t$ is a realization of $Z_t$ for some $t \in [1, T]$ and where $\alpha_\cdot \in \big[\frac{1}{2\pi}, +\infty\big)$. This last constraint implies that the generative model recreates independently each dimension of $X_t$ from a latent state $z_t$ with the same fixed precision. Moreover, the PO-DTMC formulation implies that each pair of observations $(X_t, Y_t)$ is only dependent on $Z_t$, i.e. 
\begin{align*}
	p_{X, Y | Z}\left(x, y ~|~ z; \theta\right)  = \prod_{t =1}^{T} p_{X_i, Y_i|Z_i}(x_t, y_t ~|~ z_t; \theta),
	%	p_{X, Y | Z}  = \prod_{t =1}^{T} p_{X_t|Z_t} ~.~ p_{Y_t|Z_t},
\end{align*}
and that the Markovian property holds in latent space, i.e.
\begin{align*}
	p_{Z}(z;\theta) = p_{Z_i}(z_1).\prod_{t=2}^{T} p_{Z_{i+1}|Z_{i}}(z_t ~|~ z_{t-1};\theta).
	%	p_{Z} = p_{Z_1}.\prod_{t=2}^{T} p_{Z_t|Z_{t-1}}.
\end{align*}

\begin{figure}[h]
	\centering
	\begin{tikzpicture}[scale=0.58, every node/.style={transform shape},
		node/.style={draw, dot,minimum size=0.2cm, inner sep=0pt},
		det/.style={draw, diamond,minimum size=1.2cm, inner sep=1pt},
		rect/.style={draw, rectangle,minimum size=1.1cm, inner sep=2pt},
		var/.style={draw, circle, minimum size=1.25cm}
		]
		
		%	\node[draw, dashed, rectangle,minimum size=(7cm,8cm)] at (0, 0) {};
		\draw[draw=black, rounded corners] (-\x-11.75, -3) rectangle (-\x+0.5 ,3);
		\node (name) at (-\x+0.1, -2.6) {$N$};
		
		\node[var] (theta) at (-\x-13, 0) {$\theta$};
		\node (proxy) at (-\x-11.65, 0) {};
		
		\node[var] (b0) at (-\x-9.5, 0) {$Y_{1}$};
		\node[var] (b1) at (-\x-7, 0) {$Y_{2}$};
		\node[var] (bt1prev) at (-\x-3, 0) {$Y_{T-1}$};
		\node[var] (bt1) at (-\x-0.5, 0) {$Y_{T}$};
		\node[var] (x0) at (-\x-10.5, 2) {$X_{1}$};
		\node[var] (x1) at (-\x-8, 2) {$X_{2}$};
		\node[var] (xt1prev) at (-\x-4, 2) {$X_{T-1}$};
		\node[var] (xt1) at (-\x-1.5, 2) {$X_{T}$};
		
		\node[var] (z0S) at (-\x-10.5, -2) {$Z_{1}$};
		\node[var] (zt1prevS) at (-\x-4, -2) {$Z_{T-1}$};
		\node[circle, minimum size=1.1cm] (zt0S) at (-\x-6, -2) {$...$};
		\node[var] (z1S) at (-\x-8, -2) {$Z_{2}$};
		\node[var] (zt1S) at (-\x-1.5, -2) {$Z_{T}$};
		
		\draw[->,>=latex] (z0S) to (z1S);
		\draw[->,>=latex] (z1S) to (zt0S);
		\draw[->,>=latex] (zt0S) to (zt1prevS);
		\draw[->,>=latex] (zt1prevS) to (zt1S);
		
		\draw[->,>=latex] (z0S) to (x0);
		\draw[->,>=latex] (z1S) to (x1);
		\draw[->,>=latex] (zt1prevS) to (xt1prev);
		\draw[->,>=latex] (zt1S) to (xt1);
		
		\draw[->,>=latex] (z0S) to (b0);
		\draw[->,>=latex] (z1S) to (b1);
		\draw[->,>=latex] (zt1prevS) to (bt1prev);
		\draw[->,>=latex] (zt1S) to (bt1);
		
		\draw[->,>=latex] (b0) to (x0);
		\draw[->,>=latex] (b1) to (x1);
		\draw[->,>=latex] (bt1prev) to (xt1prev);
		\draw[->,>=latex] (bt1) to (xt1);
		
		\draw[->,>=latex] (theta) to (proxy);
		
		%	\draw[->,>=latex] (theta) to [out=40,in=-150] (x0);
		%	\draw[->,>=latex] (theta) to [out=40,in=-150] (x1);
		%	\draw[->,>=latex] (theta) to [out=40,in=-150] (xt1prev);
		%	\draw[->,>=latex] (theta) to [out=40,in=-150] (xt1);
		%	
		%	\draw[->,>=latex] (theta) to [out=0,in=-150] (b0);
		%	\draw[->,>=latex] (theta) to [out=0,in=-150] (b1);
		%	\draw[->,>=latex] (theta) to [out=0,in=-160] (bt1prev);
		%	\draw[->,>=latex] (theta) to [out=0,in=-160] (bt1);
		%	
		%	\draw[->,>=latex] (theta) to [out=-40,in=-150] (z0S);
		%	\draw[->,>=latex] (theta) to [out=-40,in=-150] (z1S);
		%	\draw[->,>=latex] (theta) to [out=-40,in=-150] (zt1prevS);
		%	\draw[->,>=latex] (theta) to [out=-40,in=-150] (zt1S);
		
		%	\draw[->,>=latex] (cell0h1) to [out=0,in=90] (fcn);
		%	\draw[->,>=latex] (ct0) to [out=20,in=-190] (cell1c0);
	\end{tikzpicture}
	\caption{\small
		Bayesian network of our generative model of parameters in $\theta$. We have $N$ replications of this model, corresponding to the $N$ sequences of length $T$ in our dataset. The parameter set $\theta$ influences the inference of all variables in the model for the $N$ sequences we have.
	}\label{fig:bayesnet-model}
\end{figure}
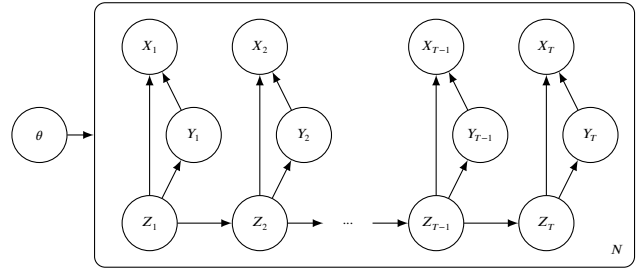

Fig. \ref{fig:bayesnet-model} provides the Bayesian network corresponding to our model.

Thus, based on a dataset of $N$ independent sequences of partial and complete observations $D = (x_{1:T}, ~y_{1:T})_{1:N}$, we want to optimize the parameters $\theta$ so that the probability that the model generates the sequences of $D$ is maximal under its constraints. In other words, we want to find the parameters $\theta$ that maximize $p_{(X,Y)^{(1)}, \dots, (X,Y)^{(N)}}(D;\theta)$, which is the same as finding $\theta$ maximizing $\log p_{(X,Y)^{(1)}, \dots, (X,Y)^{(N)}}(D;\theta)$. We have:
\begin{align*}
	\log p_{(X,Y)^{(1)}, \dots, (X,Y)^{(N)}}(D;\theta) &= \sum_{(x, y) \in D} \log p_{X,Y}(x, y;\theta)%\\
	%&= \sum_{(x_{1:T}, y_{1:T})_k \in X} \log \int_{z} p_\theta((x_{1:T}, y_{1:T})_k ~|~ z_{1:T}) ~.~ p_\theta(z_{1:T}) ~dz_{1:T}\\
	%\underset{\text{(Jensen's inequality)}}{\geq}& \sum_{(x_{1:T}, y_{1:T})_k \in X} \int_{z} \log p_\theta((x_{1:T}, y_{1:T})_k ~|~ z_{1:T}) + \log p_\theta(z_{1:T}) ~dz_{1:T}
\end{align*}
where
\begin{align*}
	&p_{X,Y}(x, y;\theta) \\&= \int p_{X,Y|Z}(x, y ~|~ z;\theta) ~.~ p_{Z}(z; \theta) ~dz\\
	&= \int \cdots \int p_{Z_i}(z_1) . \prod_{t =1}^{T} p_{X_i, Y_i|Z_i}(x_t, y_t ~|~ z_t;\theta) \\
	&\qquad\qquad\qquad\qquad~.~ \prod_{t=2}^{T} p_{Z_{i+1}|Z_{i}}(z_t ~|~ z_{t-1};\theta) ~\prod_{t=1}^{T} dz_{t}
\end{align*}
which is intractable, due to the fact that $\mu_{z}$, $\sigma_{z}$, $\mu_{x}$ and $\mu_{y}$ are multi-layers neural networks with nonlinearities. This intractability is amplified by the fact that we work with sequences of $T$ non-independent continuous latent states, which implies a multiple integral over $\mathbb{R}^{T\times d}$.
% makes the number of outcomes to evaluate grow exponentially.
This means that we cannot evaluate or differentiate the marginal likelihood $p_{X,Y}(x, y;\theta)$. For the same reasons, the posterior distribution 
\begin{align*}
	p_{Z|X,Y} (\cdot | x, y;\theta) &= \frac{p_{X,Y|Z} (x, y| \cdot~;\theta) . p_Z(\cdot~;\theta)}{p_{X,Y} (x, y;\theta)},
\end{align*}
%where
%\begin{align*}
%p_\theta (z_{t} ~| x_{1:t}, y_{1:t}, z_{t+1}) &= \frac{p_\theta (x_{1:t}, y_{1:t}~| z_{t}) .  p_\theta(z_{t+1} ~| z_t) . p_\theta(z_{t})}{\int p_\theta (x_{1:t}, y_{1:t} ~| z_{t}) . p_\theta(z_{t+1} ~| z_t) . p_\theta(z_t) ~dz_t},
%\end{align*}
is intractable, which implies that methods based on the posterior distribution such as the Expectation-Maximization (EM) algorithm cannot be employed either. So, let us adopt the Variational Bayesian (VB) approach by introducing a variational distribution dependent on a parameter set $\phi$ to approximate $p_{Z|X,Y} (\cdot | x, y;\theta)$. 
%However, contrary to classic VB methods, we will not approximate $p_\theta (z_{t} ~| x_{1:T}, y_{1:T})$ but $p_\theta(z_t ~| x_{1:t})$, as 
But, more than just a mathematical trick, we want this variational distribution to actually be a recognition model such that it is able to infer latent states only given past partial observations, in order to infer $y$ and to be able to generate plausible next observations.

\subsection{TD-VAE model}\label{td-vae}

TD-VAE \cite{gregor2018temporal} is a variant of the original VAE \cite{kingma2014VAE} for temporal sequences which features the particularity to separate \textit{belief states} from latent states. A belief state $b_t$ is a statistics describing $x_{1:t}$ such that $p_{Z_t|X_{1:t}}(\cdot| x_{1:t}; \theta) \approx p_{Z_t|B_t}(\cdot| b_t; \theta)$. The end goal motivating this distinction, aside theoretical accuracy, is to learn a model able to deterministically aggregate observations by updating a statistics $b_t$ that contains enough information to infer some latent state $z_t$, avoiding the accumulation of estimation errors on $z_{1:t-1}$. Since $z_t$ alone allows for predictions of next latent states, $b_t$ constitutes a belief on plausible latent dynamics that is simply updated with each new observation. This feature is important for model-based RL. 

In \cite{gregor2018temporal}, they chose additionally to make their model provide \textit{jumpy} predictions, i.e. directly predicting a latent state $z_{t+\delta}$ from some $z_{t}$ where $\delta$ is not precisely known, in order to abstract latent dynamics for the benefit of computational efficiency. Formally, they seek to optimize $\theta$ so that it maximizes the expression 
\begin{align}\label{tdvae_obj}
	\underset{\delta \sim \mathcal{U}_{[\delta_i, \delta_s]}}{\mathbb{E}}\left[\underset{t \sim \mathcal{U}_{[1, T-\delta]}}{\mathbb{E}}\left[\log p_{X_{t+\delta}|B_t}\left(x_{t+\delta}|b_{t}; \theta \right)\right]\right],
\end{align}
where $\mathcal{U}_{[a,b]}$ is the uniform distribution on the interval $[a,b]$ and $B_t = \text{RNN}(X_t, B_{t-1}; \phi)$. This cannot be optimized directly, as showed in the previous section. However, we can maximize a lower bound of this expression by introducing a variational distribution.

Let $Q_{t,\delta}(\phi) = q_{Z_t, Z_{t+\delta}|B_t, B_{t+\delta}}(\cdot | b_{t}, b_{t+\delta};\phi)$ be this variational distribution, dependent on a parameter set $\phi$, such that
\begin{align*}
	q_{Z_{t}, Z_{t+\delta} | B_t, B_{t+\delta}}\left(\cdot | b_{t}, b_{t+\delta};\phi \right) \approx p_{Z_{t}, Z_{t+\delta} | B_{t}, X_{t+\delta}}\left(\cdot | b_{t}, x_{t+\delta}; \theta \right)
\end{align*}
where it is important to notice that
\begin{align*}
	p_{Z_{t}, Z_{t+\delta} | B_{t}, X_{t+\delta}}\left(\cdot | b_{t}, x_{t+\delta}; \theta \right) &= \frac{p_{X_{t+\delta}, Z_t, Z_{t+\delta} | B_t}\left(x_{t+\delta}, \cdot | b_{t}; \theta \right)}{p_{X_{t+\delta}|B_t}\left(x_{t+\delta}|b_{t}; \theta \right)} \\&= \frac{P_{t,\delta}(\theta)}{p_{X_{t+\delta}|B_t}\left(x_{t+\delta}|b_{t}; \theta \right)}.
\end{align*}
To find the optimal parameters $\phi$ that minimize its approximation error, we can optimize $\phi$ so that it minimizes through gradient descent the following average Kullback-Leibler (KL) divergence:
\begin{align*}%\label{tdvae-dkl}
	%	&D_{KL}\left(q_{Z_{t}, Z_{t+\delta} | X_{1:t}, X_{t+\delta}}\left(\cdot | x_{1:t}, x_{t+\delta};\phi \right) ~\bigg|\bigg|~ p_{Z_{t}, Z_{t+\delta} | X_{1:t}, X_{t+\delta}}\left(\cdot | x_{1:t}, x_{t+\delta}; \theta \right)\right)\\
	\underset{\delta \sim \mathcal{U}_{[\delta_i, \delta_s]}}{\mathbb{E}}&\left[\underset{t \sim \mathcal{U}_{[1, T-\delta]}}{\mathbb{E}}\left[\vphantom{\frac{P_{t,\delta}(\theta)}{p_{X_{i+\delta}|B_i}\left(x_{t+\delta}|b_{t}; \theta \right)}}\right.\right. \\
	&\left.\left. D_{KL}\left(Q_{t,\delta}(\phi) ~\bigg|\bigg|~ \frac{P_{t,\delta}(\theta)}{p_{X_{i+\delta}|B_i}\left(x_{t+\delta}|b_{t}; \theta \right)}\right)\right]\right],
\end{align*}
%\begin{align*}
%	Q_{t,\delta}(\phi) &= q_{Z_t, Z_{t+\delta} | B_t,B_{t+\delta}}\left(\cdot | b_{t}, b_{t+\delta};\phi \right),
%	\\
%	P_{t,\delta}(\theta) &= p_{X_{t+\delta}, Z_t, Z_{t+\delta} | B_t}\left(x_{t+\delta}, \cdot | b_{t}; \theta \right).
%\end{align*}
This cannot be optimized directly either. Yet, it can be shown that we can equivalently minimize this divergence, while also maximizing a lower bound of (\ref{tdvae_obj}),
%$$\underset{\delta \sim \mathcal{U}_{[\delta_i, \delta_s]}}{\mathbb{E}}\left[\underset{t \sim \mathcal{U}_{[1, T-\delta]}}{\mathbb{E}}\left[\log p_{X_{t+\delta}|B_t}\left(x_{t+\delta}|b_{t}; \theta \right)\right]\right],$$ 
by minimizing the following loss w.r.t. $\phi$ and $\theta$:
\begin{align*}
	%	&D_{KL}\big(q_\phi\left(z_{t_1}, z_{t_2} | x_{\leq t_1}, x_{t_2}\right) ~||~ p_\theta\left(z_{t_1}, z_{t_2} | x_{\leq t_1}, x_{t_2}\right)\big)\\
	%	&= \underset{(z_{t_1}, z_{t_2}) \sim q_\phi\left(z_{t_1}, z_{t_2} | x_{\leq t_1}, x_{t_2}\right)}{\mathbb{E}} \big[ \log q_\phi\left(z_{t_1}, z_{t_2} | x_{\leq t_1}, x_{t_2}\right) - \log p_\theta\left(z_{t_1}, z_{t_2} | x_{\leq t_1}, x_{t_2}\right)\big]
	&\mathcal{L_{\text{TD-VAE}}}(x; \theta, \phi)\\&= \underset{\delta \sim \mathcal{U}_{[\delta_i, \delta_s]}}{\mathbb{E}}\left[\underset{t \sim \mathcal{U}_{[1, T-\delta]}}{\mathbb{E}}\left[
%	\right.\right.\\
%	&\qquad\qquad\left.\left. 
	D_{KL}\left(Q_{t,\delta}(\phi) ~||~ P_{t,\delta}(\theta)\right)\right] \vphantom{\underset{t \sim \mathcal{U}_{[1, T-\delta]}}{\mathbb{E}}}\right]
\end{align*}
where
\begin{align*}
	&D_{KL}\left(Q_{t,\delta}(\phi) ~||~ P_{t,\delta}(\theta)\right)
	\\
	&= \underset{Z_{t}, Z_{t+\delta} \sim Q_{t,\delta}(\phi)
		%	q_{Z, Z | B,B}\left(\cdot | b_t, b_{t+\delta};\phi \right)
	}{\mathbb{E}} \left[ \log q_{Z_i | B_i}\left(z_{t+\delta} | b_{t+\delta};\phi \right) \right.\nonumber\\
	&\qquad\left.+ \log q_{Z_t | B_t, B_{t+\delta}, Z_{t+\delta}}\left(z_{t} | b_{t}, b_{t+\delta}, z_{t+\delta} ;\phi \right)\right.\\
	&\qquad\left. - \log p_{Z_i|B_i}\left(Z_{t} | b_{t};\theta\right) - \log p_{Z_{+\delta}|Z}\left(Z_{t+\delta} | Z_{t};\theta\right) \vphantom{\underset{t \sim \mathcal{U}_{[1, T-\delta]}}{\mathbb{E}}}\right.\nonumber\\
	&\qquad\left.-\log p_{X_i|Z_i}\left(x_{t+\delta} | Z_{t+\delta};\theta\right)
	\right]\nonumber.
	%&\left.+ \log q_{Z|B}\left(Z_{t+\delta} | b_{t+\delta};\phi\right) + \log q_{Z-|B, B, Z}\left(Z_{t} | b_{t}, b_{t+\delta}, Z_{t+\delta};\phi\right) \right]\nonumber
\end{align*}
In complement, 
%\begin{align*}
%q_{Z, Z | B,B}\left(z_{t}, z_{t+\delta} | b_{t}, b_{t+\delta};\phi \right) = q_{Z | B}\left(z_{t+\delta} | b_{t+\delta};\phi \right) ~.~ q_{Z- | B, B, Z}\left(z_{t} | b_{t}, b_{t+\delta}, z_{t+\delta} ;\phi \right),
%\end{align*}
the authors of \cite{gregor2018temporal} had to make the strong assumption that $p_{Z_i|B_i}\left(\cdot | b_{t};\theta\right) = q_{Z_i|B_i}\left(\cdot | b_{t};\phi\right)$ for any $\theta, \phi$. They also set $p_{Z_{+\delta}|Z}\left(\cdot | z_{t};\theta\right)$ as a multivariate normal distribution with diagonal covariance matrix, corresponding to the distribution of latent states at any instants in $[t+\delta_i, t+\delta_s]$. This is in contradiction with our sequential latent model $p_{Z_{i+1}|Z_i}\left(\cdot | z_{t};{\theta}\right)$, which is itself a multivariate normal distribution with diagonal covariance matrix. In this regard, $p_{Z_{+\delta}|Z}\left(\cdot | z_{t};\theta\right)$ can be seen as a rough approximation.

This abstraction of latent dynamics may be useful in some cases where precision is not needed and the variability of observations $x_{t:t+\delta}$ gathered in a \textit{moment} can be summarized in latent space by smooth transitions between states corresponding to dataset samples. However, we argue that models of complex environments, in which the observation space is combinatorially extremely large and in which multiple agents interact with each other, require precise learning signals to \textit{understand} latent dynamics and so to generalize well outside the training set. More importantly, TD-VAE cannot consider the actions taken by the observing agent between $t$ and $t+\delta$. Yet, learning the link between actions and observations is central in RL. 
%This is why we slightly modify TD-VAE to propose a sequential variant, namely STD-VAE, in section \ref{comm:stdvae}.
%
%However, from a practical point of view, directly predicting $z_{t_2}$ from $z_{t_1}$, where $t_2 - t_1$ varies, completely abstracts latent dynamics and can only give a vague sense of what could be next rather than a real prediction, leveraging network memory of the dataset --- what is commonly considered as overfitting. While this could be useful in some cases, as advertised in their paper, ours requires robust and accurate step-by-step predictions. As suggested in the original paper, we could simply fix $t_2 - t_1 = 1$ to get a sequential model. 
%
%[LA RAISON POUR LAQUELLE LES EXPERIENCES DE TD VAE DONNENT TOUT DE MEME DE BONNES PREDICTIONS EST PROBABLEMENT PARCE QU'ILS ENCODENT TOUTE LA SÉQUENCE (Y COMPRIS LA PARTIE QUI DOIT ÊTRE PRÉDITE) AVANT DE LA "PRÉDIRE". CECI EXPLIQUE POURQUOI ILS UTILISENT DES MODELES À NOMBREUSES COUCHES (16 COUCHES DE TD VAE LIÉES PAR UN LSTM POUR LEUR ENVIRONEMENT 3D). SOIT C'EST PARCE QU'ILS CONSERVENT L'INFO DE LA SÉQUENCE, SOIT ILS ESSAIENT RÉELLEMENT DE PRÉDIRE ET ONT AINSI BESOIN DE TOUTE CETTE MACHINERIE POUR APPRENDRE PAR COEUR.]

\subsection{Our Sequential variant STD-VAE of the TD-VAE model}\label{comm:stdvae}

\begin{figure}[t]
	\centering
	\begin{tikzpicture}[scale=0.69, every node/.style={transform shape},
		node/.style={draw, dot,minimum size=0.2cm, inner sep=0pt},
		det/.style={draw, diamond,minimum size=1.2cm, inner sep=1pt},
		rect/.style={draw, rectangle,minimum size=1.1cm, inner sep=2pt},
		var/.style={draw, circle, minimum size=1.2cm}
		]
		
		\draw[draw=black, rounded corners] (-\x-12.5, -3) rectangle (-\x-0.5 ,3);
		\node (name) at (-\x-12.1, -2.6) {$N$};
		
		\node[var] (b0) at (-\x-11.5, 0) {$B_{1}$};
		\node[var] (b1) at (-\x-9, 0) {$B_{2}$};
		\node[circle, minimum size=1.1cm] (bt0) at (-\x-7, 0) {$...$};
		\node[var] (bt1prev) at (-\x-5, 0) {$B_{T-1}$};
		\node[var] (bt1) at (-\x-2.5, 0) {$B_{T}$};
		\node[var] (x0) at (-\x-10.5, 2) {$X_{1}$};
		\node[var] (x1) at (-\x-8, 2) {$X_{2}$};
		\node[var] (xt1prev) at (-\x-4, 2) {$X_{T-1}$};
		\node[var] (xt1) at (-\x-1.5, 2) {$X_{T}$};
		
		\node[var] (z0S) at (-\x-10.5, -2) {$Z_{1}$};
		\node[var] (zt1prevS) at (-\x-4, -2) {$Z_{T-1}$};
		\node[circle, minimum size=1.1cm] (zt0S) at (-\x-6, -2) {$...$};
		\node[var] (z1S) at (-\x-8, -2) {$Z_{2}$};
		\node[var] (zt1S) at (-\x-1.5, -2) {$Z_{T}$};
		
		\draw[->,>=latex, dashed] (x0) to [out=-135,in=80] (b0);
		\draw[->,>=latex, dashed] (x1) to [out=-135,in=80] (b1);
		\draw[->,>=latex, dashed] (xt1prev) to [out=-135,in=80] (bt1prev);
		\draw[->,>=latex, dashed] (xt1) to [out=-135,in=80] (bt1);
		
		\draw[->,>=latex, dashed] (b0) to [out=30,in=155] (b1);
		\draw[->,>=latex, dashed] (b1) to [out=30,in=155] (bt0);
		\draw[->,>=latex, dashed] (bt0) to [out=30,in=155] (bt1prev);
		\draw[->,>=latex, dashed] (bt1prev) to [out=30,in=155] (bt1);
		
		\draw[->,>=latex, dashed] (b0) to [out=-100,in=150] (z0S);
		\draw[->,>=latex, dashed] (b1) to [out=-100,in=150] (z1S);
		\draw[->,>=latex, dashed] (bt1prev) to [out=-100,in=150] (zt1prevS);
		\draw[->,>=latex, dashed] (bt1) to [out=-100,in=150] (zt1S);
		
		\draw[->,>=latex, dashed] (zt1S) to [out=-145,in=-30] (zt1prevS);
		\draw[->,>=latex, dashed] (zt1prevS) to [out=-145,in=-30] (zt0S);
		\draw[->,>=latex, dashed] (zt0S) to [out=-145,in=-30] (z1S);
		\draw[->,>=latex, dashed] (z1S) to [out=-145,in=-30] (z0S);
		
		\draw[->,>=latex] (z0S) to (z1S);
		\draw[->,>=latex] (z1S) to (zt0S);
		\draw[->,>=latex] (zt0S) to (zt1prevS);
		\draw[->,>=latex] (zt1prevS) to (zt1S);
		
		\draw[->,>=latex] (z0S) to (x0);
		\draw[->,>=latex] (z1S) to (x1);
		\draw[->,>=latex] (zt1prevS) to (xt1prev);
		\draw[->,>=latex] (zt1S) to (xt1);
		
		%	\draw[->,>=latex] (cell0h1) to [out=0,in=90] (fcn);
		%	\draw[->,>=latex] (ct0) to [out=20,in=-190] (cell1c0);
	\end{tikzpicture}
	\caption{\small
		Bayesian networks corresponding to STD-VAE. Solid lines represent the Bayesian network of our generative model (without $Y_t$) of parameters in $\theta$. Dashed lines represent the Bayesian network of the recognition model of parameters in $\phi$ proposed by TD-VAE.
		Parameter dependencies are not represented for the sake of clarity. Only $B_t$ is not directly influenced by $\theta$, while only variables at the end of a dashed arrow are influenced by $\phi$.
		We have $N$ replications of this model, corresponding to the $N$ sequences of length $T$ in our dataset.
	}\label{fig:bayesnet-model_phi}
\end{figure}
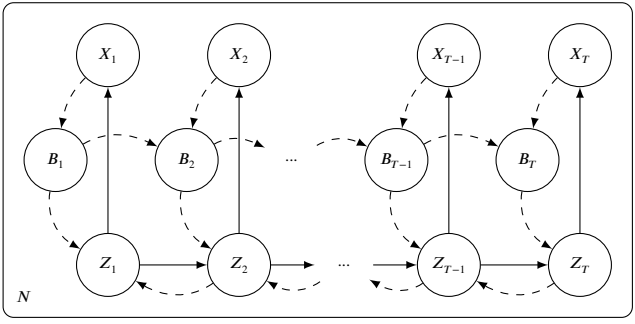

The authors of \cite{gregor2018temporal} also proposed a sequential version of their model. Its corresponding Bayesian network is given in Fig. \ref{fig:bayesnet-model_phi}.
They chose to train its parameters as a particular case of the jumpy one, simply taking $\delta = 1$. 
Yet, this would only maximize a lower bound of the probability to observe $x_{t+1}$ after $b_t$, i.e.
%\begin{align*}
$\underset{t \sim \mathcal{U}_{[1, T-1]}}{\mathbb{E}}\left[\log p_{X_{t+1}|B_t}\left(x_{t+1}|b_{t}; \theta \right)\right]$, instead of the whole future sequence $x_{t+1:T}$ after $b_t$, i.e. $\underset{t \sim \mathcal{U}_{[1, T-1]}}{\mathbb{E}}\left[\log p_{X_{t+1:T}|B_t}\left(x_{t+1:T}|b_{t}; \theta \right)\right]$.
%\end{align*}

From a practical point of view, this would prove to be computationally heavy if done multiple times per sequence and would not learn from the accumulation of prediction errors: particularly in a stochastic network such as TD-VAE and with a time step small enough, the network will tend to optimize weights such that the predicted next state looks almost identical to the initial state. It is only by chaining these predictions that their errors become significant. 
Thus, we choose a slightly different variational distribution. Let $Q_{t}(\phi) = q_{Z_{t:T} | B_{t:T}}\left(\cdot | b_{t:T};\phi \right)$ be this variational distribution, dependent on a parameter set $\phi$, such that
\begin{align*}
	q_{Z_{t:T} | B_{t:T}}\left(\cdot | b_{t:T};\phi \right) \approx p_{Z_{t:T}| B_{t},X_{t+1:T}}\left(\cdot | b_t, x_{t+1:T}; \theta \right)
\end{align*}
where it is important to notice that
\begin{align*}
	&p_{Z_{t:T}| B_{t},X_{t+1:T}}\left(\cdot | b_t, x_{t+1:T}; \theta \right) \\
	&= \frac{p_{X_{t+1:T}, Z_{t:T} | B_t}\left(x_{t+1:T}, \cdot | b_{t}; \theta \right)}{p_{X_{t+1:T}|B_{t}}\left(x_{t+1:T}|b_{t}; \theta \right)} \\&= \frac{P_{t}(\theta)}{p_{X_{t+1:T}|B_{t}}\left(x_{t+1:T}|b_{t}; \theta \right)}.
\end{align*}
To find the optimal parameters $\phi$ that minimize its approximation error, we can optimize $\phi$ so that it minimizes through gradient descent the following average Kullback-Leibler (KL) divergence:
\begin{align*}%\label{tdvae-dkl}
	%	&D_{KL}\left(q_{Z_{t}, Z_{t+\delta} | X_{1:t}, X_{t+\delta}}\left(\cdot | x_{1:t}, x_{t+\delta};\phi \right) ~\bigg|\bigg|~ p_{Z_{t}, Z_{t+\delta} | X_{1:t}, X_{t+\delta}}\left(\cdot | x_{1:t}, x_{t+\delta}; \theta \right)\right)\\
	&\underset{t \sim \mathcal{U}_{[1, T-1]}}{\mathbb{E}}\left[\vphantom{\frac{P_{t,\delta}(\theta)}{p_{X_{i+\delta}|B_i}\left(x_{t+\delta}|b_{t}; \theta \right)}}
	D_{KL}\left(Q_{t}(\phi) ~\bigg|\bigg|~ \frac{P_{t}(\theta)}{p_{X_{t+1:T}|B_{t}}\left(x_{t+1:T}|b_{t}; \theta \right)}\right)\right],
\end{align*}
It can be shown that we can equivalently minimize this divergence, while also maximizing a lower bound of $$\underset{t \sim \mathcal{U}_{[1, T-1]}}{\mathbb{E}}\left[\log p_{X_{t+1:T}|B_t}\left(x_{t+1:T}|b_{t}; \theta \right)\right],$$
by minimizing the following loss w.r.t. $\phi$ and $\theta$:
\begin{align*}
	%	&D_{KL}\big(q_\phi\left(z_{t_1}, z_{t_2} | x_{\leq t_1}, x_{t_2}\right) ~||~ p_\theta\left(z_{t_1}, z_{t_2} | x_{\leq t_1}, x_{t_2}\right)\big)\\
	%	&= \underset{(z_{t_1}, z_{t_2}) \sim q_\phi\left(z_{t_1}, z_{t_2} | x_{\leq t_1}, x_{t_2}\right)}{\mathbb{E}} \big[ \log q_\phi\left(z_{t_1}, z_{t_2} | x_{\leq t_1}, x_{t_2}\right) - \log p_\theta\left(z_{t_1}, z_{t_2} | x_{\leq t_1}, x_{t_2}\right)\big]
	&\mathcal{L_{\text{STD-VAE}}}(x; \theta, \phi)= \underset{t \sim \mathcal{U}_{[1, T-1]}}{\mathbb{E}}\left[D_{KL}\left(Q_{t}(\phi) ~||~ P_{t}(\theta)\right)\right]
\end{align*}
where
\begin{align}
	%	&D_{KL}\big(q_\phi\left(z_{t_1}, z_{t_2} | x_{\leq t_1}, x_{t_2}\right) ~||~ p_\theta\left(z_{t_1}, z_{t_2} | x_{\leq t_1}, x_{t_2}\right)\big)\\
	%	&= \underset{(z_{t_1}, z_{t_2}) \sim q_\phi\left(z_{t_1}, z_{t_2} | x_{\leq t_1}, x_{t_2}\right)}{\mathbb{E}} \big[ \log q_\phi\left(z_{t_1}, z_{t_2} | x_{\leq t_1}, x_{t_2}\right) - \log p_\theta\left(z_{t_1}, z_{t_2} | x_{\leq t_1}, x_{t_2}\right)\big]
	&D_{KL}\left(Q_{t}(\phi) ~||~ P_{t}(\theta)\right)\nonumber\\
	&= \underset{Z_{t:T} \sim Q_t(\phi)}{\mathbb{E}} \left[ \vphantom{\sum_{k=t}^{T-1}}  \log q_{Z_i|B_i}\left(Z_{T} | b_{T};\phi\right) \right.\nonumber\\
	&\quad\left.+ \sum_{k=t}^{T-1} \log q_{Z_i|B_i, Z_{i+1}}\left(Z_{k} | b_{k}, Z_{k+1};\phi\right) \right.\nonumber\\
	&\quad\left. - \log p_{Z_i|B_i}\left(Z_{t} | b_{t};\theta\right) - \sum_{k=t+1}^{T}\log p_{Z_{i+1}|Z_i}\left(Z_{k} | Z_{k-1};\theta\right) \vphantom{\underset{t \sim \mathcal{U}_{[1, T-\delta]}}{\mathbb{E}}}\right.\nonumber\\
	&\quad\left. -\sum_{k=t}^{T}\log p_{X_i|Z_i}\left(x_{k} | Z_{k};\theta\right) \right]\vphantom{\underset{t \sim \mathcal{U}_{[1, T-\delta]}}{\mathbb{E}}}\label{stdvae-loss}
\end{align}

\begin{comment}
	In addition, as mentioned in the previous section, we need to also maximize $\\\underset{t \sim \mathcal{U}_{[1, T-1]}}{\mathbb{E}}\left[\log p_{X_{1:t}}(x_{1:t};\theta)\right]$, since our model is intended to do more than predictions. This means that we want to simultaneously minimize
	\begin{align}\label{seqtdvae-dkl2}
		&\underset{t \sim \mathcal{U}_{[1, T-1]}}{\mathbb{E}}\left[D_{KL}\left(q_{Z_{1:t} | B_{1:t}}\left(\cdot | b_{1:t};\phi \right) ~\bigg|\bigg|~ p_{Z_{1:t}| X_{1:t}}\left(\cdot | x_{1:t}; \theta \right)\right)\right].
	\end{align}
\end{comment}

Fig. \ref{fig:seqtdvae} visually explains the process of evaluating (\ref{stdvae-loss}),
%the two losses derived from (\ref{seqtdvae-dkl}) and (\ref{seqtdvae-dkl2}), 
which is very similar to the original TD-VAE. 
%All distributions are assumed as Gaussian.
The belief network aggregates observations such that each belief $b_t$ is assumed to be a sufficient statistics for $x_{1:t}$. The smoothing network, knowing what the final latent state $z_{T}$ is, given observations $x_{1:T}$, infers what should have been latent states $z_{t:T-1}$. This gives us two different distributions for the inference of $z_{t}$: one given only observations $x_{1:t}$, and the other given all observations $x_{1:T}$. In the learning phase, we measure the divergence between these two distributions as a loss to prompt correct dynamics recognition and consistency in the belief network. Then, the Markovian transition model infers the next state from the current one. We infer the Gaussian parameters of the next state for each latent state inferred by the smoothing network and measure as loss the divergence between the distribution inferred by the smoothing network and the one inferred by the transition model.  
%		starts with the latent state inferred by the smoothing network, which is supposed to be the closer to the true latent state since it has been inferred from all observations up to $t_2$.
Finally, for each latent state $z_k$ sampled from the smoothing network, we infer the Gaussian parameters describing the observation $x_k$ with the decoding network and compute the negative log-likelihood of $x_k$ given these parameters as loss.

However, our preliminary experiments on this model with a dataset acquired in CARLA \cite{carla} revealed very poor prediction quality when $z_t$ is sampled from $q_{Z_t|B_t}(\cdot|b_t;\phi)$, while providing very good predictions when $z_t$ is sampled from $q_{Z_t|B}(\cdot|b_{t:T};\phi)$, i.e. from the smoothing network. 
In fact, this seems obvious considering that the prediction part of this model is trained with the latent states sampled from the variational distribution $q_{Z_{t:T} | B_{t:T}}\left(\cdot | b_{t:T};\phi \right)$ and not $q_{Z_{t:T} | B_{t}}\left(\cdot | b_{t};\phi \right)$.
This is what motivates the introduction in the next section of a local predictability constraint, allowing us to train our model on samples from $q_{Z_{t:T} | B_{t}}\left(\cdot | b_{t};\phi \right)$. This will also allow us to keep the idea of predicting distant latent states from current observations while avoiding the strong assumption that $p_{Z|B}\left(\cdot | b_{t};\theta\right) = q_{Z|B}\left(\cdot | b_{t};\phi\right)$.

\def\x{3}
\def\xoffset{0.75}
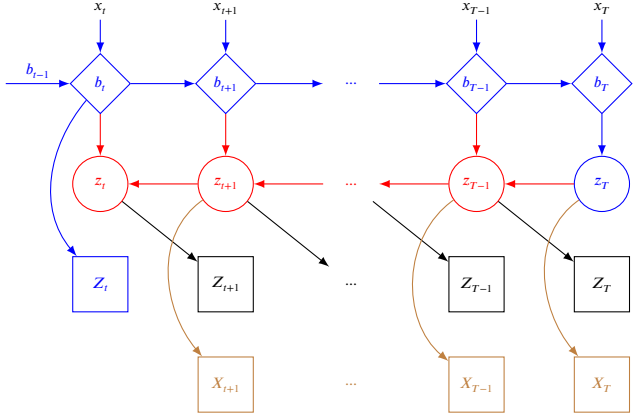
\begin{figure}[t]
	\centering
	\begin{tikzpicture}[scale=0.66, every node/.style={transform shape},
		node/.style={draw, dot,minimum size=0.2cm, inner sep=0pt},
		det/.style={draw, diamond,minimum size=1.2cm, inner sep=1pt},
		rect/.style={draw, rectangle,minimum size=1.1cm, inner sep=2pt},
		var/.style={draw, circle, minimum size=1.1cm}
		]
		
		%	\node[draw, dashed, rectangle,minimum size=(7cm,8cm)] at (0, 0) {};
		%	\draw[draw=black, dashed] (-5.5, -1.5) rectangle (4.5 ,1.5);
		%	\node (name) at (2.25, 1.75) {Self-correcting LSTM cell};
		\node (bt1prev) at (-\x-4, 0) {};
		\node[det, blue] (bt1) at (-\x-2, 0) {$b_{t}$};
		\node[det, blue] (bt1next) at (-\x+0.5, 0) {$b_{t+1}$};
		\node[circle, minimum size=1.1cm, blue] (bt) at (0, 0) {$...$};
		\node[det, blue] (bt2prev) at (\x-0.5, 0) {$b_{T-1}$};
		\node[det, blue] (bt2) at (\x+2, 0) {$b_{T}$};
		\node (bt2next) at (\x+4, 0) {};
		\node (xt1) at (-\x-2, 1.5) {$x_{t}$};
		\node (xt1next) at (-\x+0.5, 1.5) {$x_{t+1}$};
		\node (xt2prev) at (\x-0.5, 1.5) {$x_{T-1}$};
		\node (xt2) at (\x+2, 1.5) {$x_{T}$};
		
		\node[var, red] (zt1S) at (-\x-2, -2) {$z_{t}$};
		\node[var, blue] (zt2) at (\x+2, -2) {$z_{T}$};
		\node[var, red] (zt2prevS) at (\x-0.5, -2) {$z_{T-1}$};
		\node[circle, minimum size=1.1cm, red] (ztS) at (0, -2) {$...$};
		\node[var, red] (zt1nextS) at (-\x+0.5, -2) {$z_{t+1}$};
		
		%	\node (A) at (-\x-2, -3.25) {$A$};
		%	\node[rect, blue] (zt1) at (-\x-2, -4) {$\substack{\mu_{z_{t_1}} | b_{t_1}; \theta\\\sigma_{z_{t_1}} | b_{t_1}; \theta}$};
		%	\node[rect, green] (zt2T) at (\x+2, -4) {$\substack{\mu_{z_{t_2}} | z_{t_2-1}; \theta\\\sigma_{z_{t_2}} | z_{t_1-1}; \theta}$};
		%	\node[rect, green] (zt2prevT) at (\x-0.5, -4) {$\substack{\mu_{z_{t_2-1}} | z_{t_2-2}; \theta\\\sigma_{z_{t_2-1}} | z_{t2-2}; \theta}$};
		%	\node[circle, minimum size=1.1cm, green] (ztT) at (0, -4) {$...$};
		%	\node[rect, green] (zt1nextT) at (-\x+0.5, -4) {$\substack{\mu_{z_{t_1+1}} | z_{t_1}; \theta\\\sigma_{z_{t_1+1}} | z_{t_1}; \theta}$};
		\node[rect, blue] (zt1) at (-\x-2, -4) {$Z_{t}$};
		\node[rect] (zt2T) at (\x+2, -4) {$Z_{T}$};
		\node[rect] (zt2prevT) at (\x-0.5, -4) {$Z_{T-1}$};
		\node[circle, minimum size=1.1cm] (ztT) at (0, -4) {$...$};
		\node[rect] (zt1nextT) at (-\x+0.5, -4) {$Z_{t+1}$};
		
		%	\node (B) at (-\x-2, -5.25) {$B$};
		%	\node[rect, brown, dashed] (xt1D) at (-\x-2, -6) {$\substack{\mu_{x_{t_1}} | z_{t_1}; \theta}$};
		\node[rect, brown] (xt1nextD) at (-\x+0.5, -6) {$X_{t+1}$};
		\node[rect, brown] (xt2prevD) at (\x-0.5, -6) {$X_{T-1}$};
		\node[rect, brown] (xt2D) at (\x+2, -6) {$X_{T}$};
		\node[circle, minimum size=1.1cm, brown] (xtD) at (0, -6) {$...$};
		
		\draw[->,>=latex, blue] (xt1) to (bt1);
		\draw[->,>=latex, blue] (xt1next) to (bt1next);
		\draw[->,>=latex, blue] (xt2prev) to (bt2prev);
		\draw[->,>=latex, blue] (xt2) to (bt2);
		\draw[->,>=latex, blue] (bt1) to (bt1next);
		\draw[->,>=latex, blue] (bt1next) to (bt);
		\draw[->,>=latex, blue] (bt) to (bt2prev);
		\draw[->,>=latex, blue] (bt2prev) to (bt2);
		\draw[->,>=latex, blue] (bt1) to [out=-130,in=130] (zt1);
		\draw[->,>=latex, blue] (bt2) to (zt2);
		%	\draw[->,>=latex, blue] (bt2) to node[font=\small, auto] {$b_{T+1}$} (bt2next);
		\draw[->,>=latex, blue] (bt1prev) to node[font=\small, auto] {$b_{t-1}$} (bt1);
		\draw[->,>=latex, red] (zt2) to (zt2prevS);
		%	\draw[->,>=latex, red] (bt2) to (zt2prevS);
		\draw[->,>=latex, red] (bt2prev) to (zt2prevS);
		\draw[->,>=latex, red] (zt2prevS) to (ztS);
		%	\draw[->,>=latex, red] (bt2prev) to (ztS);
		\draw[->,>=latex, red] (ztS) to (zt1nextS);
		%	\draw[->,>=latex, red] (bt) to (zt1nextS);
		\draw[->,>=latex, red] (bt1next) to (zt1nextS);
		\draw[->,>=latex, red] (zt1nextS) to (zt1S);
		%	\draw[->,>=latex, red] (bt1next) to (zt1S);
		\draw[->,>=latex, red] (bt1) to (zt1S);
		\draw[->,>=latex] (zt1S) to (zt1nextT);
		\draw[->,>=latex] (zt1nextS) to (ztT);
		\draw[->,>=latex] (ztS) to (zt2prevT);
		\draw[->,>=latex] (zt2prevS) to (zt2T);
		%	\draw[->,>=latex, brown, dashed] (zt1S) to [out=-145,in=130] (xt1D);
		\draw[->,>=latex, brown] (zt1nextS) to [out=-145,in=130] (xt1nextD);
		\draw[->,>=latex, brown] (zt2prevS) to [out=-145,in=140] (xt2prevD);
		\draw[->,>=latex, brown] (zt2) to [out=-145,in=130] (xt2D);
		
		%	\draw[->,>=latex] (cell0h1) to [out=0,in=90] (fcn);
		%	\draw[->,>=latex] (ct0) to [out=20,in=-190] (cell1c0);
	\end{tikzpicture}
	\caption{\small
		Illustration of the forward computations allowing for the evaluation of the STD-VAE loss (\ref{stdvae-loss}).
		%		the two losses derived from (\ref{seqtdvae-dkl}) and (\ref{seqtdvae-dkl2}).
		%		mixing their jumpy prediction learning scheme with their step-wise sequential loss. 
		A diamond indicates a deterministically inferred variable. A rectangle indicates the deterministic inference of distribution parameters. A circle indicates the deterministic inference of distribution parameters and a sample from this distribution.
		The blue network is the belief network. The red network is the smoothing network. The black network is the Markovian transition model. The brown network is the decoding network. 
		%		The dashed element with label A only exists when considering the loss derived from (\ref{seqtdvae-dkl}), while the dashed element with label B only exists when considering the loss derived from (\ref{seqtdvae-dkl2}).
	}\label{fig:seqtdvae}
\end{figure}

\subsection{Our Locally Predictable VAE (LP-VAE) model}\label{lp-vae}

First, we put a local predictability constraint for the model to be able to predict multiple time steps into the future: 
%\begin{align}\label{predictability}
%p_{Z_{+H}}(\cdot|~z_{t_1};\theta) \approx p_{Z_{+H}|X}(\cdot|~z_{t_1}, (x, y)_{t_1+[1:H]};\theta)
%\end{align}
%\begin{align}\label{predictability}
%p_{Z_{+H}}(\cdot|~x_{1:t_1};\theta) \approx p_{Z_{+H}|X}(\cdot|~(x, y)_{1:H};\theta)
%\end{align}
\begin{align}\label{predictability}p_{Z|X_{1:t}}(\cdot|~x_{1:t};\theta) \approx p_{Z|X,Y}(\cdot|~x, y;\theta)
\end{align}
for any instant $t \geq t_{\text{min}}$. This means that there must be some instant $t_{\text{min}}$ such that the partial observations $x_{1:t_{\text{min}}}$ are sufficient to recognize the latent dynamics of the whole sequence, i.e. such that all observations $y_{1:T}$ and all subsequent partial observations $x_{t_{\text{min}}+1:T}$ bring negligible additional information in the recognition of these latent dynamics. Notice that 
\begin{align*}
	p_{Z|X,Y}(\cdot|~x, y;\theta) = \frac{p_{X,Y,Z}(x, y, \cdot~;\theta)}{p_{X,Y}(x, y;\theta)} = \frac{P(\theta)}{p_{X,Y}(x, y;\theta)},
\end{align*}
and let us note $P_t(\theta) = p_{Z|X_{1:t}}(\cdot|~x_{1:t};\theta)$.
To enforce Eq. (\ref{predictability}), we want to minimize the average KL divergence
\begin{align*}
	&\underset{t \sim~ \mathcal{U}_{[t_{\text{min}}, ~T-1]}}{\mathbb{E}}\left[D_{KL}\left( P_t(\theta) ~\bigg |\bigg |~ \frac{P(\theta)}{p_{X,Y}(x, y;\theta)} \right) \right] \\
	%&=\underset{t \sim~ \mathcal{U}_{[t_{\text{min}}, ~T-1]}}{\mathbb{E}}\left[ D_{KL}\left( p_{Z|X_{1:t}}(\cdot|~x_{1:t};\theta) ~\bigg |\bigg |~ \frac{p_{X,Y,Z}(x, y, \cdot~;\theta)}{p_{X,Y}(x, y;\theta)} \right) \right]\\
	&= \log p_{X,Y}(x, y;\theta) + \underset{t \sim~ \mathcal{U}_{[t_{\text{min}}, ~T-1]}}{\mathbb{E}}\left[ D_{KL}\left( P_t(\theta) ~ | |~ P(\theta) \right) \right],
\end{align*}
which we cannot minimize directly, due to the intractability of $p_{X,Y}(x, y;\theta)$ and $p_{Z|X_{1:t}}(\cdot|~x_{1:t};\theta)$. However, we have:
\begin{align}\label{UB}
	&\underset{t \sim~ \mathcal{U}_{[t_{\text{min}}, ~T-1]}}{\mathbb{E}}\left[ D_{KL}\left( P_t(\theta) ~ | |~ P(\theta) \right) \right]\nonumber\\
	&= -\log p_{X,Y}(x, y;\theta)\nonumber \\
	&\qquad+ \underset{t \sim~ \mathcal{U}_{[t_{\text{min}}, ~T-1]}}{\mathbb{E}}\left[ D_{KL}\left( P_t(\theta) ~\bigg |\bigg |~ \frac{P(\theta)}{p_{X,Y}(x, y;\theta)} \right) \right]\nonumber\\
	&\geq -\log p_{X,Y}(x, y;\theta),
\end{align}
since the KL divergence is always nonnegative for two probability distributions.
So, by optimizing $\theta$ to minimize $\underset{t \sim~ \mathcal{U}_{[t_{\text{min}}, ~T-1]}}{\mathbb{E}}\left[ D_{KL}\left( P_t(\theta) ~ | |~ P(\theta) \right) \right]$, 
%we minimize $\underset{t_1 \sim~ \mathcal{U}_{[t_{\text{min}}, ~T-1]}}{\mathbb{E}}\left[D_{KL}\left( p_{Z|X_{1:t_1}}(\cdot|~x_{1:t_1};\theta) ~\bigg |\bigg |~ p_{Z|X,Y}(\cdot|~x, y;\theta) \right)\right]$ or 
we maximize a lower bound of $p_{X,Y} (x, y;\theta)$, 
%where $q_{\text{enc}}(t_1, x_{1:t_1}) = p_{Z_{i+[1:T-t_1]}|Z_{i}}\left(\cdot | \cdot~;\theta\right).\displaystyle\prod_{t=1}^{t_1} q_{Z_t|X_{1:t}}(\cdot | x_{1:t};\phi)$, 
which is our primary goal. Thus, 
%to also minimize $\underset{t_1 \sim~ \mathcal{U}_{[t_{\text{min}}, ~T-1]}}{\mathbb{E}}\left[ D_{KL}\left( p_{Z|X_{1:t_1}}(\cdot|~x_{1:t_1};\theta) ~\bigg |\bigg |~ p_{Z|X,Y}(\cdot|~x, y;\theta) \right) \right]$, 
we can simply introduce a variational distribution to approximate $p_{Z|X_{1:t}}(\cdot|~x_{1:t};\theta)$ as long as we simultaneously minimize the aforementioned KL divergence. Such a variational distribution corresponds to a recognition model that tries to predict the next latent states in addition to recognizing the current and past ones, which is more useful than one that would directly approximate $p_{Z|X,Y}(\cdot|~x, y;\theta)$.

%Of course, $p_{Z|X_{1:t_1}}(\cdot|~x_{1:t_1};\theta)$ is also intractable. However, 
Notice that:
\begin{align}
	%&p_{Z|X_{1:t}}(z|~x_{1:t};\theta)\nonumber \\
	%&= p_{Z_{1:t}|X_{1:t}}(z_{1:t}|~x_{1:t};\theta)\nonumber\\
	%&\qquad~.~ p_{Z_{t+1:T}|Z_{1:t}, X_{1:t}}(z_{t+1:T}|z_{1:t}, x_{1:t};\theta)\nonumber\\
	%&= p_{Z_{1:t}|X_{1:t}}(z_{1:t}|~x_{1:t};\theta) ~.~ \prod_{k=t+1}^{T} p_{Z_{i+1}|Z_i}(z_{k}|z_{k-1};\theta)\nonumber\\
	%&= p_{Z_t|X_{1:t}}(z_{t}|~x_{1:t};\theta) ~.~ p_{Z_{1:t-1}|Z_t, X_{1:t}}(z_{1:t-1}|z_{t}, x_{1:t};\theta) \nonumber\\
	%&\qquad~.~ \prod_{k=t+1}^{T} p_{Z_{i+1}|Z_i}(z_{k}|z_{k-1};\theta)\nonumber\\
	%&= p_{Z_t|X_{1:t}}(z_{t}|~x_{1:t};\theta) ~.~ \prod_{k=1}^{t-1} p_{Z_{k}|Z_{k+1},X_{1:k}}(z_{k}|z_{k+1}, x_{1:k};\theta) \nonumber\\
	%&\qquad~.~ \prod_{k=t+1}^{T} p_{Z_{i+1}|Z_i}(z_{k}|z_{k-1};\theta)\label{lpvae-var},
	&p_{Z|X_{1:t}}(z|~x_{1:t};\theta)\nonumber \\
	%&= p_{Z|X}(z_{1:t}|~x_{1:t};\theta)~.~ p_{Z|Z, X}(z_{t+1:T}|z_{1:t}, x_{1:t};\theta)\nonumber\\
	%&= p_{Z|X}(z_{1:t}|~x_{1:t};\theta) ~.~ \prod_{k=t+1}^{T} p_{Z|Z}(z_{k}|z_{k-1};\theta)\nonumber\\
	&= p_{Z|X}(z_{t}|~x_{1:t};\theta) ~.~ p_{Z|Z, X}(z_{1:t-1}|z_{t}, x_{1:t};\theta)\nonumber \\
	&\qquad\qquad\qquad\qquad\qquad~.~ 
	p_{Z|Z, X}(z_{t+1:T}|z_{1:t}, x_{1:t};\theta)
	%\prod_{k=t+1}^{T} p_{Z|Z}(z_{k}|z_{k-1};\theta)
	\nonumber\\
	&= p_{Z|X}(z_{t}|~x_{1:t};\theta) ~.~ \prod_{k=1}^{t-1} p_{Z|Z,X}(z_{k}|z_{k+1}, x_{1:k};\theta) \nonumber\\
	&\qquad\qquad\qquad\qquad\qquad~.~ \prod_{k=t+1}^{T} p_{Z|Z}(z_{k}|z_{k-1};\theta)\label{lpvae-var},
\end{align}
omitting variable indices in distribution indices for the sake of clarity. Based on this decomposition, let us introduce two variational distributions $Q^1_{t}(\phi) = q_{Z_t|X_{1:t}}(\cdot| x_{1:t};\phi)$ and $Q^2_{t}(\phi) = q_{Z_t|X_{1:t}, Z_{t+1}}(\cdot| x_{1:t}, z_{t+1};\phi)$ taking their parameters in the parameter set $\phi$ such that:
\begin{align*}
	q_{Z_t|X_{1:t}}(\cdot| x_{1:t};\phi) &\approx p_{Z_t|X_{1:t}}(\cdot| x_{1:t};\theta)\\%\label{qb_approx}\\
	q_{Z_t|X_{1:t}, Z_{t+1}}(\cdot| x_{1:t}, z_{t+1};\phi) &\approx p_{Z_t|X_{1:t}, Z_{t+1}}(\cdot| x_{1:t}, z_{t+1};\theta).%\label{qs_approx}
\end{align*}
We assume that both $p_{Z_t|X_{1:t}}(\cdot| x_{1:t};\theta)$ and $\\p_{Z_t|X_{1:t}, Z_{t+1}}(\cdot| x_{1:t}, z_{t+1};\theta)$ have an approximate Gaussian form with an approximately diagonal covariance matrix, i.e.
\begin{align*}
	Q^1_{t}(\phi) &= \mathcal{N}(\mu_{b}(x_{1:t};\phi), ~\sigma_{b}(x_{1:t};\phi).I_d)\\
	Q^2_{t}(\phi) &= \mathcal{N}(\mu_{s}(x_{1:t}, z_{t+1};\phi), ~\sigma_{s}(x_{1:t}, z_{t+1};\phi).I_d),
\end{align*}
where $\mu_b$, $\sigma_b$, $\mu_s$ and $\sigma_s$ are deep neural networks taking their parameters in the parameter set $\phi$. 
Taking back Eq. (\ref{lpvae-var}), we get:
\begin{align*}
	&p_{Z|X_{1:t}}(z|~x_{1:t};\theta)\\
	&\approx q_{Z|X}(z_{t}|~x_{1:t};\phi) ~.~ \prod_{k=1}^{t-1} q_{Z|Z,X}(z_{k}|z_{k+1}, x_{1:k};\phi) \\
	&\qquad\qquad\qquad\qquad\qquad~.~ \prod_{k=t+1}^{T} p_{Z|Z}(z_{k}|z_{k-1};\theta)\\
	&= q_{Z|X}(z_{1:t}| x_{1:t};\phi) ~.~ p_{Z|Z}(z_{t+1:T}|~z_{t};\theta)\\
	&= q_{Z|X_{1:t}}(z|~x_{1:t};\theta, \phi) = Q_t(\theta, \phi),
\end{align*}
which means that posing our two variational distributions $Q^1_t(\phi)$ and $Q^2_t(\phi)$ is equivalent to posing the variational distribution $Q_t(\theta, \phi) \approx p_{Z|X_{1:t}}(\cdot|~x_{1:t};\theta)$.

Therefore, we want to optimize $\phi$ and $\theta$ to minimize 
$$
\underset{t \sim~ \mathcal{U}_{[t_{\text{min}}, ~T-1]}}{\mathbb{E}}\left[ D_{KL}\left( Q_t(\theta, \phi) ~\bigg |\bigg |~ \frac{P(\theta)}{p_{X,Y}(x, y;\theta)} \right) \right]
$$ 
while optimizing $\phi$ to minimize
$$
\underset{t \sim~ \mathcal{U}_{[t_{\text{min}}, ~T-1]}}{\mathbb{E}}\left[ D_{KL}\left( Q_t(\theta, \phi) ~||~ P_t(\theta) \right) \right].
$$
Actually, to achieve both these objectives, we only need to minimize 
\begin{align}\label{lpvae-loss_general}
	&\mathcal{L_{\text{LP-VAE}}}(x,y; \theta, \phi) = \underset{t \sim~ \mathcal{U}_{[t_{\text{min}}, ~T-1]}}{\mathbb{E}}\left[ D_{KL}\left( Q_t(\theta, \phi) ~ ||~ P(\theta) \right) \right]
\end{align}
w.r.t. both $\phi$ and $\theta$. See Appendix \ref{app:lpvae-loss} for more details.
Developing the KL divergence of Eq. (\ref{lpvae-loss_general}) to make our recurrent distributions appear, we finally obtain:
\begin{align}
	&D_{KL}\left( Q_t(\theta, \phi) ~ ||~ P(\theta) \right)\nonumber\\
	&=\underset{Z \sim Q_t(\theta, \phi)}{\mathbb{E}}\left[ \log q_{Z_{1:t}|B_{1:t}}(Z_{1:t}| b_{1:t};\phi)\right. \nonumber\\
	&\left.\qquad\qquad+ \log p_{Z_{t+1:T} | Z_{t}}(Z_{t+1:T}|~Z_{t};\theta) \right]\nonumber\\
	&\qquad- \underset{Z \sim Q_t(\theta, \phi)}{\mathbb{E}}\left[ \log p_{Z_{1:t}}(Z_{1:t}~;\theta) \right.\nonumber\\
	&\left.\qquad\qquad\qquad+ \log p_{Z_{t+1:T} | Z_{t}}(Z_{t+1:T} | Z_{t}~;\theta) \right.\nonumber\\
	&\left.\qquad\qquad\qquad+ \log p_{X,Y|Z}(x, y | Z~;\theta) \vphantom{p_{Z_{1:t}}(Z_{1:t}~;\theta)}\right]\nonumber\\
	&=D_{KL}\left(q_{Z_{1:t}|B_{1:t}}(\cdot| b_{1:t};\phi) ~||~ p_{Z_{1:t}} (\cdot~;\theta)\right) \nonumber\\
	&\qquad- \underset{Z \sim Q_t(\theta, \phi)}{\mathbb{E}}\left[\log p_{X,Y|Z} (x, y |Z;\theta)\right]\label{reco-gene}
\end{align}
which leads to
\begin{align}\label{lpvae-loss}
	&D_{KL}\left( Q_t(\theta, \phi) ~ ||~ P(\theta) \right)
	%= \underset{t_1 \sim~ \mathcal{U}_{[t_{\text{min}}, ~T-1]}}{\mathbb{E}}\left[ D_{KL}\left( q_{Z|X_{1:t_1}}(\cdot|~x_{1:t_1};\theta, \phi) ~\bigg |\bigg |~ p_{X,Y,Z}(x, y, \cdot~;\theta) \right) \right]
	\nonumber\\
	&= \underset{Z \sim Q_t(\theta, \phi)}{\mathbb{E}} \left[ \vphantom{\sum_{k=1}^{T}} \log q_{Z_i|B_i}\left(Z_{t} | b_{t};\phi\right) \right.\nonumber\\
	&\qquad\left.+ \sum_{k=1}^{t-1} \log q_{Z_{i}|B_i, Z_{i+1}}\left(Z_{k} | b_{k}, Z_{k+1};\phi\right)\right.\nonumber\\
	&\qquad\left. - \log p_{Z_i}\left(Z_{1};\theta\right) - \sum_{k=2}^{t} \log p_{Z_{i+1}|Z_i}\left(Z_{k} | Z_{k-1};\theta\right) \right.\nonumber\\
	&\qquad\left. -\sum_{k=1}^{T} \log p_{X_i, Y_i|Z_i}\left(x_{k}, y_k | Z_{k};\theta\right) \right]
\end{align}

\begin{figure*}[t]
	\centering
	\begin{tikzpicture}[scale=0.75, every node/.style={transform shape},
		node/.style={draw, dot,minimum size=0.2cm, inner sep=0pt},
		det/.style={draw, diamond,minimum size=1.2cm, inner sep=1pt},
		rect/.style={draw, rectangle,minimum size=1.1cm, inner sep=2pt},
		var/.style={draw, circle, minimum size=1.1cm}
		]
		
		%	\node[draw, dashed, rectangle,minimum size=(7cm,8cm)] at (0, 0) {};
		%	\draw[draw=black, dashed] (-5.5, -1.5) rectangle (4.5 ,1.5);
		%	\node (name) at (2.25, 1.75) {Self-correcting LSTM cell};
		\node[det, blue] (b0) at (-\x-10.5, 0) {$b_{1}$};
		\node[det, blue] (b1) at (-\x-8, 0) {$b_{2}$};
		\node[circle, minimum size=1.1cm, blue] (bt0) at (-\x-6, 0) {$...$};
		\node[det, blue] (bt1prev) at (-\x-4, 0) {$b_{t-1}$};
		\node[det, blue] (bt1) at (-\x-1.5, 0) {$b_{t}$};
		\node (x0) at (-\x-10.5, 1.5) {$x_{1}$};
		\node (x1) at (-\x-8, 1.5) {$x_{2}$};
		\node (xt1prev) at (-\x-4, 1.5) {$x_{t-1}$};
		\node (xt1) at (-\x-1.5, 1.5) {$x_{t}$};
		
		\node[var, red] (z0S) at (-\x-10.5, -2) {$z_{1}$};
		\node[var, red] (zt1prevS) at (-\x-4, -2) {$z_{t-1}$};
		\node[circle, minimum size=1.1cm, red] (zt0S) at (-\x-6, -2) {$...$};
		\node[var, red] (z1S) at (-\x-8, -2) {$z_{2}$};
		
		\node[var, blue] (zt1S) at (-\x-1.5, -2) {$z_{t}$};
		\node[var] (zt2) at (\x+1.5, -2) {$z_{T}$};
		\node[var] (zt2prevS) at (\x-1, -2) {$z_{T-1}$};
		\node[circle, minimum size=1.1cm] (ztS) at (0, -2) {$...$};
		\node[var] (zt1nextS) at (-\x+1, -2) {$z_{t+1}$};
		
		\node[rect] (z1T) at (-\x-8, -4) {$Z_{2}$};
		\node[rect] (zt1prevT) at (-\x-4, -4) {$Z_{t-1}$};
		\node[circle, minimum size=1.1cm] (zt0T) at (-\x-6, -4) {$...$};
		\node[rect] (zt1T) at (-\x-1.5, -4) {$Z_{t}$};
		
		\node[rect, brown] (x0D) at (-\x-10.5, -6) {$XY_{1}$};
		\node[rect, brown] (x1D) at (-\x-8, -6) {$XY_{2}$};
		\node[rect, brown] (xt1prevD) at (-\x-4, -6) {$XY_{t-1}$};
		\node[rect, brown] (xt1D) at (-\x-1.5, -6) {$XY_{t}$};
		\node[rect, brown] (xt1nextD) at (-\x+1, -6) {$XY_{t+1}$};
		\node[rect, brown] (xt2prevD) at (\x-1, -6) {$XY_{T-1}$};
		\node[rect, brown] (xt2D) at (\x+1.5, -6) {$XY_{T}$};
		\node[circle, minimum size=1.1cm, brown] (xtD) at (0, -6) {$...$};
		\node[circle, minimum size=1.1cm, brown] (xtD) at (-\x-6, -6) {$...$};
		
		\draw[->,>=latex, blue] (x0) to (b0);
		\draw[->,>=latex, blue] (x1) to (b1);
		\draw[->,>=latex, blue] (xt1prev) to (bt1prev);
		\draw[->,>=latex, blue] (xt1) to (bt1);
		
		\draw[->,>=latex, blue] (b0) to (b1);
		\draw[->,>=latex, blue] (b1) to (bt0);
		\draw[->,>=latex, blue] (bt0) to (bt1prev);
		\draw[->,>=latex, blue] (bt1prev) to (bt1);
		
		\draw[->,>=latex, red] (b0) to (z0S);
		\draw[->,>=latex, red] (b1) to (z1S);
		\draw[->,>=latex, red] (bt1prev) to (zt1prevS);
		\draw[->,>=latex, blue] (bt1) to (zt1S);
		
		\draw[->,>=latex, red] (zt1S) to (zt1prevS);
		\draw[->,>=latex, red] (zt1prevS) to (zt0S);
		\draw[->,>=latex, red] (zt0S) to (z1S);
		\draw[->,>=latex, red] (z1S) to (z0S);
		
		\draw[->,>=latex] (zt2prevS) to (zt2);
		\draw[->,>=latex] (ztS) to (zt2prevS);
		\draw[->,>=latex] (zt1nextS) to (ztS);
		\draw[->,>=latex] (zt1S) to (zt1nextS);
		
		\draw[->,>=latex] (z0S) to (z1T);
		\draw[->,>=latex] (zt0S) to (zt1prevT);
		\draw[->,>=latex] (zt1prevS) to (zt1T);
		
		\draw[->,>=latex, brown] (z0S) to (x0D);
		\draw[->,>=latex, brown] (z1S) to [out=-40,in=45] (x1D);
		\draw[->,>=latex, brown] (zt1prevS) to [out=-40,in=45] (xt1prevD);
		\draw[->,>=latex, brown] (zt1S) to [out=-40,in=45] (xt1D);
		\draw[->,>=latex, brown] (zt1nextS) to (xt1nextD);
		\draw[->,>=latex, brown] (zt2prevS) to (xt2prevD);
		\draw[->,>=latex, brown] (zt2) to (xt2D);
		
		%	\draw[->,>=latex] (cell0h1) to [out=0,in=90] (fcn);
		%	\draw[->,>=latex] (ct0) to [out=20,in=-190] (cell1c0);
	\end{tikzpicture}
	\caption{\small
		Illustration of the forward computations allowing for the evaluation of the LP-VAE loss.
		%		mixing their jumpy prediction learning scheme with their step-wise sequential loss. 
		A diamond indicates a deterministically inferred variable. A rectangle indicates the deterministic inference of distribution parameters. A circle indicates the deterministic inference of distribution parameters and a sample from this distribution. 
		The blue network is the belief network. The red network is the smoothing network. The black network is the Markovian transition model. The brown network is the decoding network. 
	}\label{fig:lpvae}
\end{figure*}
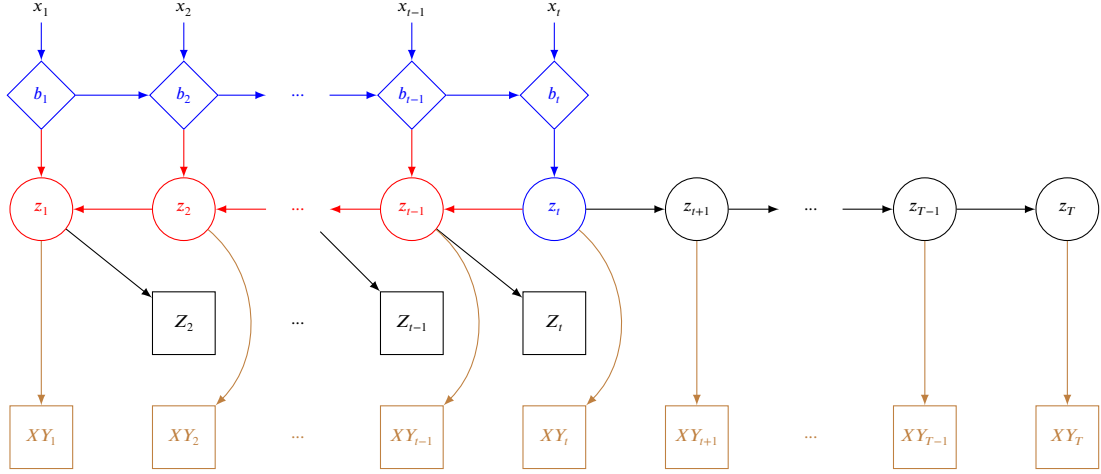

Fig. \ref{fig:lpvae} illustrates the process of evaluating (\ref{lpvae-loss}).
We can easily give an interpretation to this loss:
%noticing that
%\begin{align}
%&D_{KL}\left( Q_t(\theta, \phi) ~ ||~ P(\theta) \right)\nonumber\\
%&=\underset{Z \sim Q_t(\theta, \phi)}{\mathbb{E}}\left[ \log q_{Z_{1:t}|B_t}(Z_{1:t}| b_{t};\phi)\right. \nonumber\\
%&\left.\qquad\qquad+ \log p_{Z_{t+1:T} | Z_{1:t}}(Z_{t+1:T}|~Z_{t};\theta) \right]\nonumber\\
%&\qquad- \underset{Z \sim Q_t(\theta, \phi)}{\mathbb{E}}\left[ \log p_{Z_{1:t}}(Z_{1:t}~;\theta) \right.\nonumber\\
%&\left.\qquad\qquad\qquad+ \log p_{Z_{t+1:T} | Z_{1:t}}(Z_{t+1:T} | Z_{1:t}~;\theta) \right.\nonumber\\
%&\left.\qquad\qquad\qquad+ \log p_{X,Y|Z}(x, y | Z~;\theta)\right]\nonumber\\
%&=D_{KL}\left(q_{Z_{1:t}|B_{t}}(\cdot| b_{t};\phi) ~||~ p_{Z_{1:t}} (\cdot~;\theta)\right) \nonumber\\
%&\qquad- \underset{Z \sim Q_t(\theta, \phi)}{\mathbb{E}}\bigg[\log p_{X,Y|Z} (x, y |Z;\theta)\bigg]\label{reco-gene}
%\end{align}
%From there, 
we can identify two global objectives in Eq. (\ref{reco-gene}) that are reminiscent of the original VAE \cite{kingma2014VAE} in terms of interpretation: the $D_{KL}$ term is an encoder loss for the recognition model of parameters $\phi$, while the second term is a decoder loss for the generative model of parameters $\theta$. It can be viewed as a precision loss (second term) optimized against a regularization (first term) to prevent from overfitting.

We can even go deeper in interpretation to highlight what differs from the original VAE. Contrary to the original VAE, our model generates a sequence of observations instead of an isolated one. Doing so, we have a Markovian transition model that predicts a latent state from the previous one with its own set of parameters separated from the decoder ones. 
Therefore, it seems natural to have a third loss term for prediction. We can make it appear by splitting the second term of Eq. (\ref{reco-gene}), i.e.:
\begin{align*}
	&D_{KL}\left( Q_t(\theta, \phi) ~ ||~ P(\theta) \right)\nonumber\\
	&=D_{KL}\left(q_{Z_{1:t}|B_{1:t}}(\cdot| b_{1:t};\phi) ~||~ p_{Z_{1:t}} (\cdot~;\theta)\right) \nonumber\\
	&\quad- \underset{Z \sim Q_t(\theta, \phi)}{\mathbb{E}}\bigg[\log p_{(X,Y)_{1:t}|Z_{1:t}} ((x, y)_{1:t} |Z_{1:t};\theta)\bigg]\nonumber\\%\label{reco-dec}\\
	&\quad- \underset{Z \sim Q_t(\theta, \phi)}{\mathbb{E}}\bigg[\log p_{(X,Y)_{t+1:T}|Z_{t+1:T}} ((x, y)_{t+1:T} |Z_{t+1:T};\theta)\bigg]
	%\label{pred}
\end{align*}
The first term is an encoder loss. The second term is a decoder loss. The third term is a prediction loss.
This prediction loss can also be viewed as a loss optimized against a regularization since the $D_{KL}$ term affects the inference of $Z_{t}$ by the recognition model from which the next latent states are predicted.

\subsection{LP-VAE with actions}\label{full-model}

The models we described up to this point represents the environment evolving around the observing agent. However, our agent also acts on this environment and influences the observations gathered to train our model. Thus, we need to modify it in order to integrate this subtlety. 

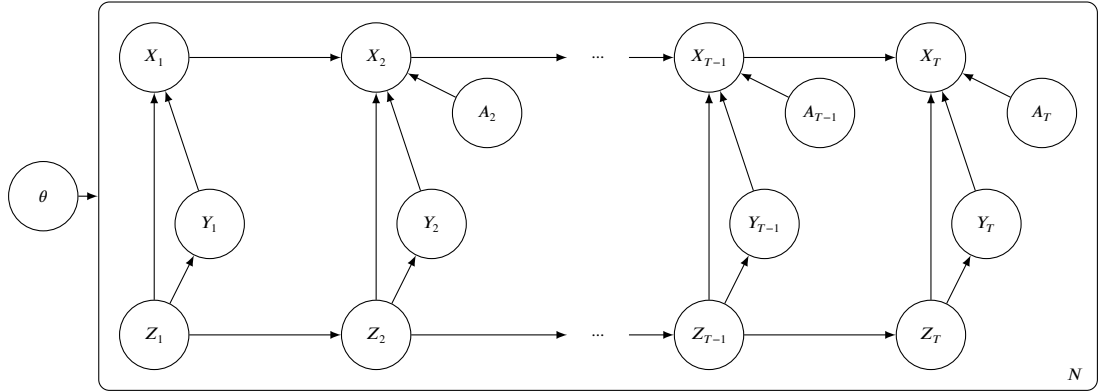
\begin{figure*}[t]
	\centering
	\begin{tikzpicture}[scale=0.73, every node/.style={transform shape},
		node/.style={draw, dot,minimum size=0.2cm, inner sep=0pt},
		det/.style={draw, diamond,minimum size=1.2cm, inner sep=1pt},
		rect/.style={draw, rectangle,minimum size=1.1cm, inner sep=2pt},
		var/.style={draw, circle, minimum size=1.25cm}
		]
		
		%	\node[draw, dashed, rectangle,minimum size=(7cm,8cm)] at (0, 0) {};
		\draw[draw=black, rounded corners] (-\x-11, -3) rectangle (-\x+7 ,4);
		\node (name) at (-\x+6.6, -2.7) {$N$};
		
		\node[var] (theta) at (-\x-12, 0.5) {$\theta$};
		\node (proxy) at (-\x-10.9, 0.5) {};
		
		\node[var] (b0) at (-\x-9, 0) {$Y_{1}$};
		\node[var] (b1) at (-\x-5, 0) {$Y_{2}$};
		\node[var] (bt1prev) at (-\x+1, 0) {$Y_{T-1}$};
		\node[var] (bt1) at (-\x+5, 0) {$Y_{T}$};
		%	\node[var] (xi0) at (-\x-10, 4) {$\widetilde{X}_{1}$};
		%	\node[var] (xi1) at (-\x-6, 4) {$\widetilde{X}_{2}$};
		%	\node[var] (xit1prev) at (-\x, 4) {$\widetilde{X}_{T-1}$};
		%	\node[var] (xit1) at (-\x+4, 4) {$\widetilde{X}_{T}$};
		
		\node[var] (m1) at (-\x-4, 2) {$A_{2}$};
		\node[var] (mt1prev) at (-\x+2, 2) {$A_{T-1}$};
		\node[var] (mt1) at (-\x+6, 2) {$A_{T}$};
		
		\node[var] (x0) at (-\x-10, 3) {$X_{1}$};
		\node[var] (x1) at (-\x-6, 3) {$X_{2}$};
		\node[circle, minimum size=1.1cm] (xt0) at (-\x-2, 3) {$...$};
		\node[var] (xt1prev) at (-\x, 3) {$X_{T-1}$};
		\node[var] (xt1) at (-\x+4, 3) {$X_{T}$};
		
		%	\node[var] (x0) at (-\x-8, 4) {$X_{1}$};
		%	\node[var] (x1) at (-\x-4, 4) {$X_{2}$};
		%	\node[circle, minimum size=1.1cm] (xt0) at (-\x-2, 4) {$...$};
		%	\node[var] (xt1prev) at (-\x+2, 4) {$X_{T-1}$};
		%	\node[var] (xt1) at (-\x+6, 4) {$X_{T}$};
		
		%	\node[var] (a1) at (-\x-4, -1.15) {$A_{2}$};
		%	\node[var] (at1prev) at (-\x+2, -1.15) {$A_{T-1}$};
		%	\node[var] (at1) at (-\x+6, -1.15) {$A_{T}$};
		
		\node[var] (z0S) at (-\x-10, -2) {$Z_{1}$};
		\node[var] (zt1prevS) at (-\x, -2) {$Z_{T-1}$};
		\node[circle, minimum size=1.1cm] (zt0S) at (-\x-2, -2) {$...$};
		\node[var] (z1S) at (-\x-6, -2) {$Z_{2}$};
		\node[var] (zt1S) at (-\x+4, -2) {$Z_{T}$};
		
		\draw[->,>=latex] (z0S) to (z1S);
		\draw[->,>=latex] (z1S) to (zt0S);
		\draw[->,>=latex] (zt0S) to (zt1prevS);
		\draw[->,>=latex] (zt1prevS) to (zt1S);
		
		%	\draw[->,>=latex] (z0S) to (xi0);
		%	\draw[->,>=latex] (z1S) to (xi1);
		%	\draw[->,>=latex] (zt1prevS) to (xit1prev);
		%	\draw[->,>=latex] (zt1S) to (xit1);
		
		\draw[->,>=latex] (z0S) to (x0);
		\draw[->,>=latex] (z1S) to (x1);
		\draw[->,>=latex] (zt1prevS) to (xt1prev);
		\draw[->,>=latex] (zt1S) to (xt1);
		
		\draw[->,>=latex] (z0S) to (b0);
		\draw[->,>=latex] (z1S) to (b1);
		\draw[->,>=latex] (zt1prevS) to (bt1prev);
		\draw[->,>=latex] (zt1S) to (bt1);
		
		\draw[->,>=latex] (b0) to (x0);
		\draw[->,>=latex] (b1) to (x1);
		\draw[->,>=latex] (bt1prev) to (xt1prev);
		\draw[->,>=latex] (bt1) to (xt1);
		
		%	\draw[->,>=latex] (b0) to (xi0);
		%	\draw[->,>=latex] (b1) to (xi1);
		%	\draw[->,>=latex] (bt1prev) to (xit1prev);
		%	\draw[->,>=latex] (bt1) to (xit1);
		
		\draw[->,>=latex] (x0) to (x1);
		\draw[->,>=latex] (x1) to (xt0);
		\draw[->,>=latex] (xt0) to (xt1prev);
		\draw[->,>=latex] (xt1prev) to (xt1);
		
		%	\draw[->,>=latex] (x0) to (xi1);
		%	\draw[->,>=latex] (x1) to (xt0);
		%	\draw[->,>=latex] (xt0) to (xit1prev);
		%	\draw[->,>=latex] (xt1prev) to (xit1);
		
		%	\draw[->,>=latex] (xi0) to (x0);
		%	\draw[->,>=latex] (xi1) to (x1);
		%	\draw[->,>=latex] (xit1prev) to (xt1prev);
		%	\draw[->,>=latex] (xit1) to (xt1);
		
		\draw[->,>=latex] (m1) to (x1);
		\draw[->,>=latex] (mt1prev) to (xt1prev);
		\draw[->,>=latex] (mt1) to (xt1);
		
		%	\draw[->,>=latex] (b1) to (m1);
		%	\draw[->,>=latex] (bt1prev) to (mt1prev);
		%	\draw[->,>=latex] (bt1) to (mt1);
		
		%	\draw[->,>=latex] (a1) to (m1);
		%	\draw[->,>=latex] (at1prev) to (mt1prev);
		%	\draw[->,>=latex] (at1) to (mt1);
		
		\draw[->,>=latex] (theta) to (proxy);
		
		%	\draw[->,>=latex] (theta) to [out=40,in=-150] (x0);
		%	\draw[->,>=latex] (theta) to [out=40,in=-150] (x1);
		%	\draw[->,>=latex] (theta) to [out=40,in=-150] (xt1prev);
		%	\draw[->,>=latex] (theta) to [out=40,in=-150] (xt1);
		%	
		%	\draw[->,>=latex] (theta) to [out=0,in=-150] (b0);
		%	\draw[->,>=latex] (theta) to [out=0,in=-150] (b1);
		%	\draw[->,>=latex] (theta) to [out=0,in=-160] (bt1prev);
		%	\draw[->,>=latex] (theta) to [out=0,in=-160] (bt1);
		%	
		%	\draw[->,>=latex] (theta) to [out=-40,in=-150] (z0S);
		%	\draw[->,>=latex] (theta) to [out=-40,in=-150] (z1S);
		%	\draw[->,>=latex] (theta) to [out=-40,in=-150] (zt1prevS);
		%	\draw[->,>=latex] (theta) to [out=-40,in=-150] (zt1S);
		
		%	\draw[->,>=latex] (cell0h1) to [out=0,in=90] (fcn);
		%	\draw[->,>=latex] (ct0) to [out=20,in=-190] (cell1c0);
	\end{tikzpicture}
	\caption{\small
		Bayesian network of our generative model of parameters in $\theta$. We have $N$ replications of this model, corresponding to the $N$ sequences of length $T$ in our dataset. The parameter set $\theta$ influences the inference of all variables in the model for the $N$ sequences we have.
	}\label{fig:bayesnet-model-actions}
\end{figure*}

Let $A_t$ be the action applied at time $t$ on perceptions. This action describes a mask on the information contained in $Y_t$. This partial information is then transmitted to the observing agent, influencing $X_t$. It has no influence on the environment evolving around the agent, only on its perception of it. This means that $Y_t$ and $Z_t$ are not affected by $A_t$.
%Let $\widetilde{X}_t$ be a random variable corresponding to the ego-vehicle perception at time $t$ combined with the memory of previous observations.
Moreover, we will now consider that the random variable $X_t$ is the ego-vehicle perception at time $t$, eventually augmented with information from $Y_t$, in accordance with $A_t$, and combined with the discounted memory of the previous partial observations $X_{1:t-1}$.
%the observation $\widetilde{X}_t$ updated by $M_t$. 
Fig. \ref{fig:bayesnet-model-actions} provides the corresponding Bayesian network.

%In addition, let $M_t$ be the random variable representing $Y_t$ masked by $A_t$. 

We set the following constraints:
\begin{itemize}
	\item $Z_i \sim \mathcal{N}(0, I_d)$ 
	\item $p_{Z_{i+1}|Z_i}(\cdot | z_{t}; \theta) = \mathcal{N}(\mu_{z}(z_{t};\theta),~ \sigma^2_{z}(z_{t};\theta) . I_d)$
	\item $p_{Y_i|Z_i}(\cdot | z_{t}; \theta) = \mathcal{N}(\mu_{y}(z_{t}; \theta),~ \alpha_y . I_{|X_t|})$
	%	\item $p_{Y_i|X_i, Z_i}(\cdot | x_t, z_{t}; \theta) = \mathcal{N}(\mu_{c}(x_t, z_{t}; \theta),~ \alpha_c . I_{|X_t|})$
	%	\item $p_{\widetilde{X}_i|X_{i-1}, Y_i,Z_i}(\cdot | x_{t-1}, y_t, z_{t}; \theta) = \mathcal{N}(\mu_{\tilde{x}}(x_{t-1}, y_t, z_{t};\theta),~ \alpha_{\tilde{x}} . I_{|X_t|})$
	%	\item $p_{M_i|A_i, Y_i}(\cdot | a_t, y_t; \theta) = \mathcal{N}(\mu_{m}(a_t, y_t;\theta),~ \alpha_m . I_{|X_t|})$
	\item $p_{X_i|X_{i-1}, Y_t, Z_t, A_t}(\cdot | x_{t-1}, y_t, z_t, a_t; \theta) \\= \mathcal{N}(\mu_{x}(x_{t-1}, y_t, z_t, a_t;\theta),~ \alpha_x . I_{|X_t|})$
\end{itemize}
where all parameters $\mu_{\cdot}$ and $\sigma_{\cdot}$ are deep neural networks taking their parameters in $\theta$, and $\alpha_\cdot \in \big[\frac{1}{2\pi}, +\infty\big)$.

Our dataset $D$ is composed of $N$ independent sequences of partial and complete observations with a randomly chosen bounding box $A_t$, i.e. $D = (x_{1:T}, ~y_{1:T}, a_{2:T})_{1:N}$.
Fortunately, Eq. (\ref{lpvae-var}) still holds in this new model. Moreover, we know that the environment does not depend on the actions $A_{2:T}$ taken on its perception of it and that the actions only mask regions of $Y_t$ while not altering the remaining. Finally, since $X_t$ contains the information transmitted from $Y_t$ in accordance with $A_t$, the actions $A_{2:t}$ do not bring any information for the inference of the latent states $Z_{1:t}$. 
%Even if both $X$ and $Y$ are not known, giving $A$ brings negligible information for the inference of $Z$. 
Given the Bayesian network in Fig. \ref{fig:bayesnet-model-actions}, the actions $A$ without knowing $X_{t+1:T}$ do not bring any information for the inference of the latent states $Z_{t+1:T}$ either. We have:
\begin{align*}
	p_{Z|X_{1:t}, A}(\cdot|~x_{1:t}, a;\theta) = p_{Z|X_{1:t}}(\cdot|~x_{1:t};\theta)
\end{align*}
Thus, we keep the LP-VAE variational distributions 
\begin{align*}
	Q^1_{t}(\phi) &= q_{Z_t|X_{1:t}}(\cdot| x_{1:t};\phi),\\
	Q^2_{t}(\phi) &= q_{Z_t|X_{1:t}, Z_{t+1}}(\cdot| x_{1:t}, z_{t+1};\phi),\\
	Q_t(\theta, \phi) &\approx p_{Z|X_{1:t}}(\cdot|~x_{1:t};\theta).
\end{align*}
Then, for our local predictability constraint (See Eq. (\ref{predictability})), we consider $p_{Z|X,Y,A}(\cdot|~x, y,a;\theta)$ instead of $p_{Z|X,Y}(\cdot|~x, y;\theta)$. Notice that
\begin{align*}
	p_{Z|X,Y,A}(\cdot|~x, y,a;\theta) &= \frac{p_{X,Y,Z|A}(x, y, \cdot~|a;\theta)}{p_{X,Y|A}(x, y ~|a;\theta)} \\
	&= \frac{P(\theta)}{p_{X,Y|A}(x, y ~|a;\theta)}
	%&= \frac{p_{X_{2:T},Y,M,Z|X_1,A}(x_{2:T}, y, m, \cdot~|x_1, a;\theta) . p_{X_1|A}(x_1~|a;\theta)}{p_{X,Y,M|A}(x, y, m |a;\theta)} \\
	%&= \frac{p_{X_{2:T},Y,M,Z|X_1,A}(x_{2:T}, y, m, \cdot~|x_1, a;\theta)}{p_{X_{2:T},Y,M|X_1,A}(x_{2:T}, y, m |x_1,a;\theta)} \\
	%&= \frac{P(\theta)}{p_{X_{2:T},Y,M|X_1,A}(x_{2:T}, y, m |x_1,a;\theta)}.
\end{align*}
%and let us note again $P_t(\theta) = p_{Z|X_{1:t}}(\cdot|~x_{1:t};\theta)$.
We take as loss function $\mathcal{L_{\text{LP-VAE}}}(x,y~|a; \theta, \phi)$ instead of $\mathcal{L_{\text{LP-VAE}}}(x,y; \theta, \phi)$, where
\begin{align}\label{lpvae-rl-loss_general}
	&\mathcal{L_{\text{LP-VAE}}}(x,y~|a; \theta, \phi)\nonumber\\
	&\qquad= \underset{t \sim~ \mathcal{U}_{[t_{\text{min}}, ~T-1]}}{\mathbb{E}}\left[ D_{KL}\left( Q_t(\theta, \phi) ~ ||~ P(\theta) \right) \right]
	%&= \underset{t \sim~ \mathcal{U}_{[t_{\text{min}}, ~T-1]}}{\mathbb{E}}\left[ D_{KL}\left( q_{Z|B_{t}}(\cdot|~b_{t};\theta, \phi) ~\bigg |\bigg |~ p_{X,Y,M,Z|X,A}(x, y, m, \cdot~|x_1, a;\theta) \right) \right]
\end{align}
This loss maximizes a lower bound of 
%$$p_{X_{2:T},Y,M|X_1,A}(x_{2:T}, y, m |x_1,a;\theta).$$
$$p_{X,Y|A}(x, y ~|a;\theta).$$
Developing the KL divergence of Eq. (\ref{lpvae-rl-loss_general}) in accordance with our new model, we get:
\begin{align*}
	&D_{KL}\left( Q_t(\theta, \phi) ~ ||~ P(\theta) \right)
	%= \underset{t_1 \sim~ \mathcal{U}_{[t_{\text{min}}, ~T-1]}}{\mathbb{E}}\left[ D_{KL}\left( q_{Z|X_{1:t_1}}(\cdot|~x_{1:t_1};\theta, \phi) ~\bigg |\bigg |~ p_{X,Y,Z}(x, y, \cdot~;\theta) \right) \right]
	\nonumber\\
	&= \underset{Z \sim Q_t(\theta, \phi)}{\mathbb{E}} \left[ \log q_{Z_{1:t}|B_{1:t}}\left(Z_{1:t} | b_{1:t};\phi\right) \right.\nonumber\\
	&\left. \qquad+ \log p_{Z_{t+1:T}|Z_{t}}\left(Z_{t+1:T} | z_{t};\theta\right) - \log p_{Z_{1:t}}\left(Z_{1:t};\theta\right) \right.\nonumber\\
	&\left.
	%-\log p_{Y_i|X_i,Z_i}\left(y_{1} |x_1, Z_{1};\theta\right) 
	\qquad- \log p_{Z_{t+1:T}|Z_{t}}\left(Z_{t+1:T} | z_{t};\theta\right) - \log p_{Y|Z}\left(y ~| Z;\theta\right) \right.\nonumber\\
	&\left. \qquad- \log p_{X|Y, Z,A}\left(x ~| y, Z,a;\theta\right) \right]\\
	&= \underset{Z \sim Q_t(\theta, \phi)}{\mathbb{E}} \left[ \log q_{Z_{1:t}|B_{1:t}}\left(Z_{1:t} | b_{1:t};\phi\right) \right.\nonumber\\
	&\left. \qquad- \log p_{Z_{1:t}}\left(Z_{1:t};\theta\right) - \log p_{Y|Z}\left(y ~| Z;\theta\right) \right.\nonumber\\
	&\left. \qquad- \log p_{X|Y, Z,A}\left(x ~| y, Z,a;\theta\right) \right]
\end{align*}
which leads to
\begin{align}\label{lpvae-rl-loss}
	&D_{KL}\left( Q_t(\theta, \phi) ~ ||~ P(\theta) \right)\nonumber\\
	&= \underset{Z \sim Q_t(\theta, \phi)}{\mathbb{E}} \left[ \vphantom{\sum_{k=2}^{T}} \log q_{Z_i|B_i}\left(Z_{t} | b_{t};\phi\right) - \log p_{Z_i}\left(Z_{1};\theta\right) \right.\nonumber\\
	&\left. + \sum_{k=1}^{t-1} \log q_{Z_{i}|B_i, Z_{i+1}}\left(Z_{k} | b_{k}, Z_{k+1};\phi\right) \right.\nonumber\\
	&\left.
	%-\log p_{Y_i|X_i,Z_i}\left(y_{1} |x_1, Z_{1};\theta\right) 
	- \sum_{k=2}^{t} \log p_{Z_{i+1}|Z_i}\left(Z_{k} | Z_{k-1};\theta\right) -\sum_{k=1}^{T} \log p_{Y_i|Z_i}\left(y_k | Z_{k};\theta\right) \right.\nonumber\\
	&\left. -\sum_{k=2}^{T} \log p_{X_i|X_{i-1},Y_i, Z_i, A_i}\left(x_k | x_{k-1}, y_k, Z_k, a_k;\theta\right)\right.\nonumber\\
	&\left. -\log p_{X_1|Y_1,Z_1}\left(x_1 | y_1, Z_1;\theta\right) \vphantom{\sum_{k=2}^{T}}\right]
\end{align}
In practice however, we will neglect the term $\\-\log p_{X_1|Y_1,Z_1}\left(x_1 | y_1, Z_1;\theta\right)$ for several reasons. First, it avoids to optimize parameters that would only be used in the learning phase, while not corresponding to an important component (the complete observation $y_1$ being already considered and containing $x_1$). But maybe more importantly, since $X_t$ keeps a memory of past observations in this formulation of the LP-VAE, $x_1$ may also contain information on actions preceding $a_{2:T}$ that should be given as well if $x_1$ is actually not the start of an episode of interactions in the environment.
Not generating $x_1$ allows us to start the inference of latent states at any point of the episode, independently from the previous actions and observations that produced $x_1$. This means that we can re-use different subsequences of the same training sequence in the learning phase, without having to make sure that $x_1$ do not contain information related to past observations and actions.

\section{Implementation as neural networks}\label{application}

%Let $M_t$ be the random variable representing the remaining information after application of this mask on $Y_t$.
%
%$D = (x_{1:T}, ~y_{1:T}, m_{2:T}, a_{2:T})_{1:N}$

%\subsection{Model}

\subsection{Belief state computation}

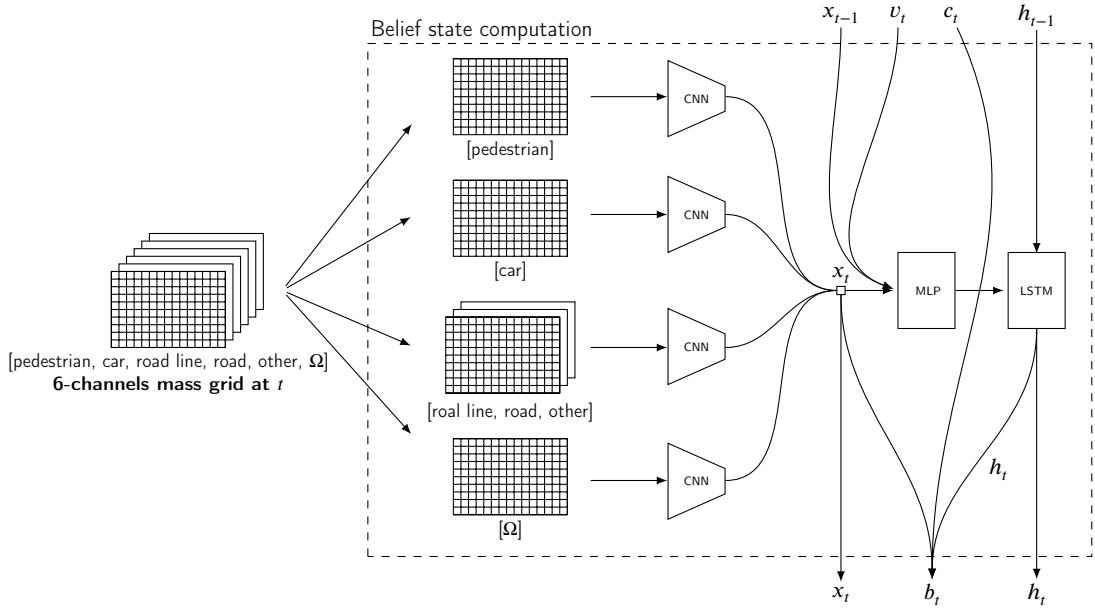
\begin{figure*}[t]
	\centering
	\begin{tikzpicture}[scale=0.5, every node/.style={transform shape},
		node/.style={draw, dot,minimum size=0.2cm, inner sep=0pt},
		det/.style={draw, diamond,minimum size=1.2cm, inner sep=1pt},
		rect/.style={draw, rectangle,minimum size=1.1cm, inner sep=2pt},
		var/.style={draw, circle, minimum size=1.1cm}
		]
		\pgfmathsetmacro{\cubex}{3}
		\pgfmathsetmacro{\cubey}{2}
		\pgfmathsetmacro{\cubez}{1}
		\pgfmathsetmacro{\offset}{0.2}
		%	\draw (0,0,0) -- ++(-\cubex,0,0) -- ++(0,-\cubey,0) -- ++(\cubex,0,0) -- cycle;
		%	\draw (0,0,0) -- ++(0,0,-\cubez) -- ++(0,-\cubey,0) -- ++(0,0,\cubez) -- cycle;
		%	\draw (0,0,0) -- ++(-\cubex,0,0) -- ++(0,0,-\cubez) -- ++(\cubex,0,0) -- cycle;
		\draw[fill=white] (5*\offset,5*\offset,0) -- ++(0,\cubey,0) -- ++(\cubex,0,0) -- ++(0,-\cubey,0) -- cycle;
		\draw[fill=white] (4*\offset,4*\offset,0) -- ++(0,\cubey,0) -- ++(\cubex,0,0) -- ++(0,-\cubey,0) -- cycle;
		\draw[fill=white] (3*\offset,3*\offset,0) -- ++(0,\cubey,0) -- ++(\cubex,0,0) -- ++(0,-\cubey,0) -- cycle;
		\draw[fill=white] (2*\offset,2*\offset,0) -- ++(0,\cubey,0) -- ++(\cubex,0,0) -- ++(0,-\cubey,0) -- cycle;
		\draw[fill=white] (\offset,\offset,0) -- ++(0,\cubey,0) -- ++(\cubex,0,0) -- ++(0,-\cubey,0) -- cycle;
		%	\draw[fill=white] (0,0,0) -- ++(0,\cubey,0) -- ++(\cubex,0,0) -- ++(0,-\cubey,0) -- cycle;
		\draw[fill=white] (0,0) grid[step=0.2] ++(\cubex,\cubey,0) rectangle (0,0);
		
		%	\draw[fill=white] (3*\cubex,3*\offset+2.5*\cubey,0) -- ++(0,\cubey,0) -- ++(\cubex,0,0) -- ++(0,-\cubey,0) -- cycle;
		\draw[fill=white] (3*\cubex,3*\offset+2.5*\cubey) grid[step=0.2] ++(\cubex,\cubey) rectangle (3*\cubex,3*\offset+2.5*\cubey);
		
		%	\draw[fill=white] (3*\cubex,1.5*\offset+1.1*\cubey,0) -- ++(0,\cubey,0) -- ++(\cubex,0,0) -- ++(0,-\cubey,0) -- cycle;
		\draw[fill=white] (3*\cubex,2*\offset+1*\cubey) grid[step=0.2] ++(\cubex,\cubey) rectangle (3*\cubex,2*\offset+1*\cubey);
		
		\draw[fill=white] (3*\cubex+\offset,-0.5*\cubey+\offset,0) -- ++(0,\cubey,0) -- ++(\cubex,0,0) -- ++(0,-\cubey,0) -- cycle;
		\draw[fill=white] (3*\cubex,-0.5*\cubey,0) -- ++(0,\cubey,0) -- ++(\cubex,0,0) -- ++(0,-\cubey,0) -- cycle;
		%	\draw[fill=white] (3*\cubex-\offset,-0.5*\cubey-\offset,0) -- ++(0,\cubey,0) -- ++(\cubex,0,0) -- ++(0,-\cubey,0) -- cycle;
		\draw[fill=white] (3*\cubex-\offset,-0.5*\cubey-\offset) grid[step=0.2] ++(\cubex,\cubey) rectangle (3*\cubex-\offset,-0.5*\cubey-\offset);
		
		%	\draw[fill=white] (3*\cubex,-2*\offset-2*\cubey,0) -- ++(0,\cubey,0) -- ++(\cubex,0,0) -- ++(0,-\cubey,0) -- cycle;
		\draw[fill=white] (3*\cubex,-2*\offset-2*\cubey) grid[step=0.2] ++(\cubex,\cubey) rectangle (3*\cubex,-2*\offset-2*\cubey);
		
		\pgfmathsetmacro{\paddingcnn}{1.325*\cubey}
		\pgfmathsetmacro{\offsetcnn}{-3.85*\cubey}
		
		\draw[fill=white] (4.95*\cubex-\offset,\offsetcnn+5*\paddingcnn,0) -- ++(0,\cubey,0) -- ++(\cubex/2,-\cubey/3,0) -- ++(0,-\cubey/3,0) -- cycle;
		\node () at (5.2*\cubex-\offset,\offsetcnn+0.5*\cubey+5*\paddingcnn) {CNN};
		
		\draw[fill=white] (4.95*\cubex-\offset,\offsetcnn+3.85*\paddingcnn,0) -- ++(0,\cubey,0) -- ++(\cubex/2,-\cubey/3,0) -- ++(0,-\cubey/3,0) -- cycle;
		\node () at (5.2*\cubex-\offset,\offsetcnn+0.5*\cubey+3.85*\paddingcnn) {CNN};
		
		\draw[fill=white] (4.95*\cubex-\offset,\offsetcnn+2.5*\paddingcnn+0.5*\offset,0) -- ++(0,\cubey,0) -- ++(\cubex/2,-\cubey/3,0) -- ++(0,-\cubey/3,0) -- cycle;
		%	\draw[fill=white] (4.95*\cubex-\offset,\offsetcnn+2.5*\paddingcnn+0.5*\offset) rectangle  ++(\cubex/2,\cubey);
		\node () at (5.2*\cubex-\offset,\offsetcnn+0.5*\cubey+2.5*\paddingcnn+0.5*\offset) {CNN};
		
		\draw[fill=white] (4.95*\cubex-\offset,\offsetcnn+1.2*\paddingcnn,0) -- ++(0,\cubey,0) -- ++(\cubex/2,-\cubey/3,0) -- ++(0,-\cubey/3,0) -- cycle;
		\node () at (5.2*\cubex-\offset,\offsetcnn+0.5*\cubey+1.2*\paddingcnn) {CNN};
		
		\node (grid) at (1.5*\cubex, 2.5*\offset+\cubey/2) {};
		%	\node () at (6*\cubex+\offset/2, 5*\cubey) {\Large locally planned path};
		%	\node[rect] () at (8*\cubex+\offset/2, 5*\cubey) {\Large 2-layers FC};
		%	\node () at (6*\cubex+\offset/2, 4.25*\cubey) {\Large[control, box]};
		\node () at (0.5*\cubex, -2*\offset) {\Large[pedestrian, car, road line, road, other, $\Omega$]};
		\node () at (0.5*\cubex, -5*\offset) {\Large\textbf{6-channels mass grid at $t$}};
		\node () at (3.5*\cubex, \offset+2.5*\cubey) {\Large[pedestrian]};
		\node () at (3.5*\cubex, 1.1*\cubey-1*\offset) {\Large[car]};
		\node () at (3.5*\cubex, -\offset -0.75*\cubey) {\Large[roal line, road, other]};
		\node () at (3.5*\cubex, -2*\cubey-4*\offset) {\Large[$\Omega$]};
		\node (ped) at (2.5*\cubex+2.5*\offset, 3.*\cubey) {};
		\node (car) at (2.5*\cubex+2.5*\offset, 2.5*\offset+1.5*\cubey) {};
		\node (lay) at (2.5*\cubex+2.5*\offset, 0) {};
		\node (om) at (2.5*\cubex+2.5*\offset, -2.*\offset-1.*\cubey) {};
		\node (pedo) at (4.*\cubex+2.5*\offset, 0.5*\offset+3.25*\cubey) {};
		\node (caro) at (4.*\cubex+2.5*\offset, 2.5*\offset+1.5*\cubey) {};
		\node (layo) at (4.*\cubex+2.5*\offset, 0) {};
		\node (omo) at (4.*\cubex+2.5*\offset, -2.5*\offset-1.5*\cubey) {};
		\node (pcnn) at (4.75*\cubex+2.5*\offset, 3*\offset+3*\cubey) {};
		\node (ccnn) at (4.75*\cubex+2.5*\offset, 2.5*\offset+1.5*\cubey) {};
		\node (lcnn) at (4.75*\cubex+2.5*\offset, 0) {};
		\node (omcnn) at (4.75*\cubex+2.5*\offset, -2.5*\offset-1.5*\cubey) {};
		
		\node (pcnno) at (5.15*\cubex+3*\offset, 3*\offset+3*\cubey) {};
		\node (ccnno) at (5.15*\cubex+3*\offset, 2.5*\offset+1.5*\cubey) {};
		\node (lcnno) at (5.15*\cubex+3*\offset, 0) {};
		\node (omcnno) at (5.15*\cubex+3*\offset, -2.5*\offset-1.5*\cubey) {};
		
		\node[draw, rectangle] (xenc) at (6*\cubex+6*\offset, 2.5*\offset+\cubey/2) {};
		\node (xenco) at (6*\cubex+6*\offset, -6.5) {\LARGE $x_{t}$};
		\node (xencprevlabel) at (6*\cubex+6*\offset, 8.75) {\LARGE $x_{t-1}$};
		\node (xenclabel) at (6*\cubex+6*\offset, 4.5*\offset+\cubey/2) {\LARGE $x_t$};
		\node (motionlabel) at (6.5*\cubex+6*\offset, 8.75) {\LARGE $v_t$};
		
		\draw[fill=white] (6.5*\cubex+6*\offset,2.5*\offset,0) -- ++(0,\cubey,0) -- ++(0.5*\cubex,0,0) -- ++(0,-\cubey,0) -- cycle;
		\node (mlpi) at (6.5*\cubex+6*\offset, 2.5*\offset+\cubey/2) {};
		\node (mlp) at (7*\cubex+2.5*\offset, 2.5*\offset+\cubey/2) {MLP};
		\node (mlpo) at (7*\cubex+5.5*\offset, 2.5*\offset+\cubey/2) {};
		
		\node (hprev) at (8*\cubex+1.75*\offset, 8.75) {\LARGE $h_{t-1}$};
		\node (lstmi0) at (8*\cubex+1.75*\offset, 6.75*\offset+\cubey/2) {};
		\node (lstmi) at (7.5*\cubex+5.5*\offset, 2.5*\offset+\cubey/2) {};
		\draw[fill=white] (7.5*\cubex+5.5*\offset,2.5*\offset,0) -- ++(0,\cubey,0) -- ++(0.5*\cubex,0,0) -- ++(0,-\cubey,0) -- cycle;
		\node (lstm) at (8*\cubex+1.75*\offset, 2.5*\offset+\cubey/2) {LSTM};
		\node (lstmo) at (8*\cubex+1.75*\offset, -2*\offset+\cubey/2) {};
		\node (lstmoh) at (8*\cubex+1.75*\offset, -6.5) {\LARGE $h_{t}$};
		
		\node (command) at (7.25*\cubex+1.75*\offset, 8.75) {\LARGE $c_t$};
		
		%\node (pcnnoo) at (8.95*\cubex+2.5*\offset, 3*\offset+3*\cubey) {\LARGE $i_p$};
		%%	\node (pccnnoo) at (9*\cubex+2.5*\offset, 4.5*\offset+1.5*\cubey) {}
		%\node (ccnnoo) at (8.95*\cubex+2.5*\offset, 1.5*\offset+0.5*\cubey) {\LARGE $i_c$};
		%%	\node (clcnnoo) at (9*\cubex+2.5*\offset, 3*\offset-1*\cubey) {};
		%\node (lcnnoo) at (8.95*\cubex+2.5*\offset, 2.5*\offset-2.25*\cubey) {\LARGE $i_l$};
		%%	\node (lpcnnoo) at (9*\cubex+2.5*\offset, 1.5*\offset-3.5*\cubey) {};
		%\node (omcnnoo) at (8.95*\cubex+2.5*\offset, 2.5*\offset-2.25*\cubey) {\LARGE $i_l$};
		\node (b) at (7*\cubex+3*\offset, -6.5) {\LARGE $b_t$};
		\draw[->,>=latex] (grid) to (ped);
		\draw[->,>=latex] (grid) to (car);
		\draw[->,>=latex] (grid) to (lay);
		\draw[->,>=latex] (grid) to (om);
		\draw[->,>=latex] (pedo) to (pcnn);
		\draw[->,>=latex] (caro) to (ccnn);
		\draw[->,>=latex] (layo) to (lcnn);
		\draw[->,>=latex] (omo) to (omcnn);
		\draw[-] (pcnno) to [out=0, in=180] (xenc);
		\draw[-] (ccnno) to [out=0, in=180] (xenc);
		\draw[-] (lcnno) to [out=0, in=180] (xenc);
		\draw[-] (omcnno) to [out=0, in=180] (xenc);
		\draw[->,>=latex] (xencprevlabel) to [out=-90, in=160] (mlpi);
		\draw[->,>=latex] (motionlabel) to [out=-90, in=160] (mlpi);
		\draw[->,>=latex] (hprev) to (lstmi0);
		\draw[->,>=latex] (xenc) to (mlpi);
		\draw[->,>=latex] (xenc) to (xenco);
		\draw[->,>=latex] (mlpo) to (lstmi);
		\draw[->,>=latex] (xenc) to [out=-90, in=90] (b);
		\draw[->,>=latex] (command) to [out=-65, in=90] (b);
		\draw[->,>=latex] (lstmo) to (lstmoh);
		\draw[->,>=latex] (lstmo) to [out=-90, in=90] node[font=\small, auto] {\LARGE $h_{t}$} (b);
		%	\draw[->,>=latex] (grid) to node[font=\small, auto] {$b_{t_1-1}$} (ped);
		\node (name) at (3.25*\cubex, 8.35) {\LARGE Belief state computation};
		\draw[draw=black, dashed] (2.25*\cubex, -5.5) rectangle (8.6*\cubex ,8);
	\end{tikzpicture}
	\caption{\small
		Illustration of the process of computing the observation $X_t$ and the belief state $B_t$ from $G_t$, $X_{t-1}$, $V_t$ and $C_t$. Four independent Convolutional VAEs are trained to learn a sufficient representation of pedestrian, car, $\{\textit{road lines, road, other}\}$ and ignorance. These encodings form $X_t$. A Multilayer perceptron (MLP) tries to learn features about the motion of road users around the ego-vehicle. The output of this MLP serves as input to a Recurrent Neural Network (RNN) composed of Long Short-Term Memory (LSTM) cells to form and update a belief over the dynamics of other road users. The concatenation of the hidden state of this RNN with $X_t$ and the driving controls $C_t$ represents the belief state $B_t$ at time $t$. 
	}\label{fig:belief-comp}
\end{figure*}

The grids $G_t$ introduced in section \ref{com:state} are not directly taken as input of our LP-VAE. Beforehand, we train a Convolutional VAE (CVAE) to learn a compressed, essentialized representation of these observations in which spatial features have been extracted. This CVAE is itself separated into 4 independent parts in order to preserve the semantics of these features: a CVAE for the pedestrian channel, another for the car channel, another for static elements (road lines, road, other) and a last one for the ignorance. The projection of $G_t$ into the latent space of this Convolutional VAE is the $X_t$ taken by our LP-VAE.
Then, we feed $X_t$, $X_{t-1}$ and the ego-motion $V_t$ to a Multilayer perceptron (MLP) in order to extract features about the motion of road users around the ego-vehicle. The output of this MLP serves as input to a Recurrent Neural Network (RNN) composed of Long Short-Term Memory (LSTM) cells to form and update a belief over the dynamics of other road users. The concatenation of the hidden state of this RNN with $X_t$ and the driving controls $C_t$ represents the belief state $B_t$ at time $t$. Fig. \ref{fig:belief-comp} visually sums up this procedure. 

\subsection{Inference of Gaussian parameters}\label{inference}

\def\x{5}
\def\xoffset{0.75}
\begin{figure}[t]
	\centering
	\begin{tikzpicture}[scale=0.74, every node/.style={transform shape},
		node/.style={draw, dot,minimum size=0.2cm, inner sep=0pt},
		det/.style={draw, diamond,minimum size=1.1cm, inner sep=0pt},
		rect/.style={draw, rectangle,minimum size=1.1cm, inner sep=2pt}
		]
		
		%	\node[draw, dashed, rectangle,minimum size=(7cm,8cm)] at (0, 0) {};
		\node (name) at (-\x+5, 3.5) {LSTM cell};
		\draw[draw=black, dashed] (-\x, -2.25) rectangle (-\x+6 ,3.25);
		\node[red] (name) at (-\x+2.1, 1.9) {\small D map};
		\draw[draw=red, dashed] (-\x+1.5, -1.35) rectangle (-\x+3.5 ,2.1);
		\node (z) at (-\x-2, 0.5) {$h$};
		\node (i) at (-\x-2, -1.75) {input};
		\node[draw, rectangle] (c0) at (-\x, 2.75) {};
		\node[draw, rectangle] (c1) at (-\x+6, 2.75) {};
		\node (c11) at (-\x+9, 2.75) {};
		\node[draw, circle] (mul) at (-\x+1, 2.75) {$\times$};
		\node[draw, circle] (add) at (-\x+3, 2.75) {$+$};
		\node[draw, rectangle] (h0) at (-\x, -1.75) {};
		\node[draw, rectangle] (h1) at (-\x+6, -1.75) {};
		\node (h11) at (-\x+9, -1.75) {};
		\node[draw, circle, minimum size=0.8cm] (s1) at (-\x+1, 0.25) {$\sigma$};
		\node[draw, rectangle] (s1n) at (-\x+1, -0.75) {FC};
		\node[draw, circle, minimum size=0.8cm] (s2) at (-\x+2, 0.25) {$\sigma$};
		\node[draw, rectangle] (s2n) at (-\x+2, -0.75) {FC};
		\node[draw, circle] (mul2) at (-\x+3, 1.5) {$\times$};
		\node[draw, circle, inner sep=0pt, minimum size=0.8cm] (tanh) at (-\x+3, 0.25) {tanh};
		\node[draw, circle, inner sep=0pt, minimum size=0.8cm] (tanho) at (-\x+5.25, 1.5) {tanh};
		\node[draw, circle] (mul3) at (-\x+5.25, 0.25) {$\times$};
		\node[draw, rectangle] (tanhn) at (-\x+3, -0.75) {FC};
		\node[draw, circle, minimum size=0.8cm] (s3) at (-\x+4, 0.25) {$\sigma$};
		\node[draw, rectangle] (s3n) at (-\x+4, -0.75) {FC};
		\node[draw, rectangle] (sigfc) at (-\x+7, -1.75) {FC};
		
		\draw[->,>=latex] (c1) to node[font=\small, auto] {$\mu_{z_t}$}  (c11);
		%	\draw[->,>=latex] (h1) to node[font=\small, auto] {$\frac{1}{S}\log(\sigma_{z_t})$}  (h11);
		\draw[->,>=latex] (h1) to (sigfc);
		\draw[->,>=latex] (sigfc) to node[font=\small, auto] {$\log(\sigma_{z_t})$}  (h11);
		\draw[->,>=latex] (c0) to (mul);
		\draw[->,>=latex] (s1) to (mul);
		\draw[->,>=latex] (s1n) to (s1);
		\draw[->,>=latex] (s2n) to (s2);
		\draw[->,>=latex] (tanhn) to (tanh);
		\draw[->,>=latex] (tanh) to (mul2);
		\draw[->,>=latex] (s3n) to (s3);
		\draw[->,>=latex] (s3) to (mul3);
		\draw[->,>=latex] (tanho) to (mul3);
		\draw[->,>=latex] (mul) to (add);
		\draw[->,>=latex] (mul2) to (add);
		\draw[-,>=latex] (add) to (c1);
		\draw[->,>=latex] (h0) to [out=0,in=-90] (s1n);
		\draw[->,>=latex] (h0) to [out=0,in=-90] (s2n);
		\draw[->,>=latex] (h0) to [out=0,in=-90] (tanhn);
		\draw[->,>=latex] (h0) to [out=0,in=-90] (s3n);
		\draw[->,>=latex] (s2) to [out=90,in=-180] (mul2);
		\draw[->,>=latex] (add) to [out=0,in=90] (tanho);
		\draw[-,>=latex] (z) to [out=0,in=-180] (c0);
		\draw[-,>=latex] (z) to [out=0,in=-180] (h0);
		\draw[-,>=latex] (mul3) to [out=-90,in=-180] (h1);
		\draw[-,>=latex] (i) to (h0);
	\end{tikzpicture}
	\caption{\small
		Proposed replacement for D maps. The \textit{FC} rectangles indicate a single Fully Connected layer. Circles indicate point-wise operations, where $\sigma$ is the sigmoïd activation function. 
	}\label{fig:Dmap}
\end{figure}
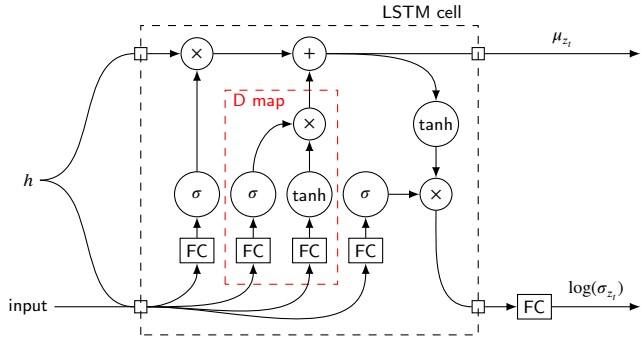

In \cite{gregor2018temporal}, they proposed to use what they called \textit{D} maps\footnote{In \cite{gregor2018temporal}, they used a 16-layer model where the information transits from layer to layer through the states of a LSTM, possibly in place of this D map, in their DeepMind Lab experiment. Note however that it is recurrent through layers, not time. This is different from what is proposed here.} to infer the Gaussian parameters of any of the distributions over the latent state $z_t$. It is a part of a LSTM cell (new features multiplied by the input gate), as indicated in Fig. \ref{fig:Dmap}, where the output is passed to two fully connected (FC) layers in parallel without activation function, one to determine $\mu_{z_t}$ and the other to determine $\log (\sigma_{z_t})$. Yet, in our sequential setting, this D map becomes a truly recurrent unit, chaining itself multiple times from $t_1$ to $1$ in the smoothing network and from $t_1$ to $t_2$ in the prediction network. As for any recurrent network, this poses the issue of vanishing gradients. Furthermore, it lacks the semantics of a transition model: some components could disappear from the frame (forget gate) and some other could become visible or simply move from their initial state (input gate, followed by an addition to the initial components). These are exactly the transformations applied to the cell state of a LSTM cell. Thus, using the cell state of a LSTM cell as latent state mean $\mu_{z_t}$ as in Fig. \ref{fig:Dmap}, where $h = z_{t+1}$ and $\text{input} = b_t$, solves both the vanishing gradient issue and the lack of model semantics. Giving $h$ as both hidden and cell states also has the effect of implementing peephole connections \cite{peepholeLSTM}, giving the cell state some control over the input, forget and output gates (the three sigmoïd layers), which better captures sporadic events. In addition, uncertainty should be encoded within the latent state to be self-sufficient for a transition model. This encourages the computation of the standard deviation $\sigma_{z_t}$ from $\mu_{z_t}$ with some filtering gate (output gate), which is exactly what a LSTM cell does to output a quantity based on its cell state. 
%Moreover, the tanh activation ensures that $\log(\sigma_{z_t})$ lies in $(-S, S)$, i.e. $\sigma_{z_t} \in (e^{-S}, e^S)$. By construction, this prevents the network from becoming deterministic (which would make it less robust to novelty) and at the same time avoids to sample too far away as the trajectory in latent space moves away from 0. 
Similarly, we use this LSTM cell in the prediction network for $p_{Z_{i+1}|Z_i}(\cdot | z_{t}; \theta)$, where $h = z_{t}$ and $\text{input} = \emptyset$. For the belief network, we keep this D map as there is no propagation in time.
% for $p_\theta(z_t | b_{t})$, where $h = W . b_t + B$ and $\text{input} = b_t$, where $W$ is a matrice of $|z_t|$ rows and $|b_t|$ columns and $B$ is a vector of size $|z_t|$, together representing a fully connected linear layer. 

\subsection{Decoding}

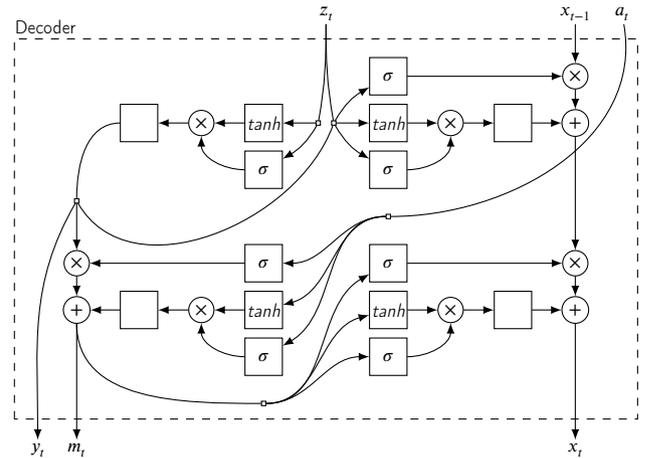
\begin{figure}[t]
	\centering
	\begin{tikzpicture}[scale=0.41, every node/.style={transform shape},
		node/.style={draw, dot,minimum size=0.2cm, inner sep=0pt},
		det/.style={draw, diamond,minimum size=1.2cm, inner sep=1pt},
		rect/.style={draw, rectangle,minimum size=1.2cm, inner sep=2pt},
		var/.style={draw, circle, minimum size=0cm},
		stoc/.style={draw, circle, minimum size=1.3cm}
		]
		\node[] (z) at (0, 5) {\LARGE $z_t$};
		\node[draw, rectangle,minimum size=0cm, inner sep=2pt] (zproxyl) at (-0.25, 1.5) {};
		\node[draw, rectangle,minimum size=0cm, inner sep=2pt] (zproxyr) at (0.25, 1.5) {};
		
		\node[] (xencprev) at (8, 5) {\LARGE $x_{t-1}$};
		\node[rect] (xencpriorf) at (2, 3) {\LARGE $\sigma$};
		\node[var] (xencpriorp) at (8, 3) {\LARGE $\times$};
		\node[rect] (xencpriorfc) at (6, 1.5) {};
		\node[var] (xencpriora) at (8, 1.5) {\LARGE $+$};
		\node[rect] (xencpriorif) at (2, 1.5) {\LARGE $\textit{tanh}$};
		\node[rect] (xencpriorig) at (2, 0) {\LARGE $\sigma$};
		\node[var] (xencpriorip) at (4, 1.5) {\LARGE $\times$};
		%	\node[] (xencprior) at (6, -1.5) {\LARGE $x_{\textit{prior}, t}$};
		
		%	\node[draw, rectangle,minimum size=0cm, inner sep=2pt] (xencfull) at (-6, -1.) {};
		%	\node[] (xencfullo) at (-9, -9) {\LARGE $y_{t}$};
		
		\node[rect] (xencfullf) at (-6, 1.5) {};
		%	\node[stoc] (xencfullp) at (-6, 3) {\LARGE $y_t$};
		%	\node[var] (xencfulla) at (-6, 1.5) {\LARGE $+$};
		\node[rect] (xencfullif) at (-2, 1.5) {\LARGE $\textit{tanh}$};
		\node[rect] (xencfullig) at (-2, 0) {\LARGE $\sigma$};
		\node[var] (xencfullip) at (-4, 1.5) {\LARGE $\times$};
		\node[draw, rectangle,minimum size=0cm, inner sep=2pt] (xencfull) at (-8, -1.) {};
		\node[] (xencfullo) at (-9.25, -9) {\LARGE $y_{t}$};
		
		\node (box) at (9.5, 5) {\LARGE $a_{t}$};
		\node[draw, rectangle,minimum size=0cm, inner sep=2pt] (boxproxy) at (2, -1.5) {};
		\node[rect] (xencmaskf) at (-2, -3) {\LARGE $\sigma$};
		\node[var] (xencmaskp) at (-8, -3) {\LARGE $\times$};
		\node[rect] (xencmaskfc) at (-6, -4.5) {};
		\node[var] (xencmaska) at (-8, -4.5) {\LARGE $+$};
		\node[rect] (xencmaskif) at (-2, -4.5) {\LARGE $\textit{tanh}$};
		\node[rect] (xencmaskig) at (-2, -6) {\LARGE $\sigma$};
		\node[var] (xencmaskip) at (-4, -4.5) {\LARGE $\times$};
		\node[] (xencmask) at (-8, -9) {\LARGE $m_{t}$};
		\node[draw, rectangle,minimum size=0cm, inner sep=2pt] (xencmaskproxy) at (-2, -7.5) {};
		
		\node[rect] (xencf) at (2, -3) {\LARGE $\sigma$};
		\node[var] (xencp) at (8, -3) {\LARGE $\times$};
		\node[rect] (xencfc) at (6, -4.5) {};
		\node[var] (xenca) at (8, -4.5) {\LARGE $+$};
		\node[rect] (xencif) at (2, -4.5) {\LARGE $\textit{tanh}$};
		\node[rect] (xencig) at (2, -6) {\LARGE $\sigma$};
		\node[var] (xencip) at (4, -4.5) {\LARGE $\times$};
		\node[] (xenc) at (8, -9) {\LARGE $x_{t}$};
		
		%		\node[draw, rectangle,minimum size=0cm, inner sep=2pt] (zproxyp) at (-0.25, 8.5) {};
		%	\node[] (xencprior) at (-6, 11.75) {\LARGE $x_{t}$};
		%	\node[] (y) at (-6, 5) {\LARGE $y_{t}$};
		%	\node[rect] (yf) at (-2, 10) {\LARGE $\sigma$};
		%	\node[var] (yp) at (-6, 10) {\LARGE $\times$};
		%	\node[var] (ya) at (-6, 8.5) {\LARGE $+$};
		%	\node[rect] (yif) at (-2, 8.5) {\LARGE $\textit{tanh}$};
		%	\node[rect] (yig) at (-2, 7) {\LARGE $\sigma$};
		%	\node[var] (yip) at (-4, 8.5) {\LARGE $\times$};
		
		%	\draw[-] (z) to [out=90, in=-80] (zproxyp);
		%	\draw[->,>=latex] (zproxyp) to [out=120, in=-30] (yf);
		%	\draw[->,>=latex] (zproxyp) to [out=180, in=0] (yif);
		%	\draw[->,>=latex] (zproxyp) to [out=-120, in=30] (yig);
		
		%	\draw[->,>=latex] (xencprior) to (yp);
		%	\draw[->,>=latex] (yf) to (yp);
		%	\draw[->,>=latex] (yif) to (yip);
		%	\draw[->,>=latex] (yig) to [out=180, in=-90] (yip);
		%	\draw[->,>=latex] (yip) to (ya);
		%	\draw[->,>=latex] (yp) to (ya);
		%	\draw[->,>=latex] (ya) to (y);
		
		\draw[->,>=latex] (xencprev) to (xencpriorp);
		\draw[->,>=latex] (xencpriora) to (xencp);
		%	\draw[->,>=latex] (xencfullprev) to (xencfullp);
		
		\draw[-] (z) to [out=-90, in=80] (zproxyl);
		\draw[-] (z) to [out=-90, in=100] (zproxyr);
		%	\draw[->,>=latex] (zproxyl) to [out=120, in=-30] (xencfullf);
		\draw[->,>=latex] (zproxyl) to [out=180, in=0] (xencfullif);
		\draw[->,>=latex] (zproxyl) to [out=-120, in=30] (xencfullig);
		
		%	\draw[->,>=latex] (xencfullf) to (xencfullp);
		\draw[->,>=latex] (xencfullif) to (xencfullip);
		\draw[->,>=latex] (xencfullig) to [out=180, in=-90] (xencfullip);
		\draw[->,>=latex] (xencfullip) to (xencfullf);
		%	\draw[->,>=latex] (xencfullp) to (xencfulla);
		\draw[-] (xencfullf) to [out=-180, in=90] (xencfull);
		
		\draw[->,>=latex] (xencfull) to [out=-120, in=90] (xencfullo);
		
		\draw[->,>=latex] (zproxyr) to [out=60, in=-150] (xencpriorf);
		\draw[->,>=latex] (zproxyr) to [out=0, in=180] (xencpriorif);
		\draw[->,>=latex] (zproxyr) to [out=-60, in=150] (xencpriorig);
		%	\draw[->,>=latex] (xencpriorcat) to [out=-90, in=180] (xencpriorf);
		%	\draw[->,>=latex] (xencpriorcat) to [out=-90, in=150] (xencpriorif);
		%	\draw[->,>=latex] (xencpriorcat) to [out=-90, in=135] (xencpriorig);
		\draw[-] (xencfull) to [out=-60, in=-110] (zproxyr);
		%	\draw[->,>=latex] (xencfull) to [out=-30, in=-135] (xencpriorf);
		%	\draw[->,>=latex] (xencfull) to [out=-30, in=-150] (xencpriorif);
		%	\draw[->,>=latex] (xencfull) to [out=-30, in=-180] (xencpriorig);
		
		\draw[->,>=latex] (xencpriorf) to (xencpriorp);
		\draw[->,>=latex] (xencpriorif) to (xencpriorip);
		\draw[->,>=latex] (xencpriorig) to [out=0, in=-90] (xencpriorip);
		\draw[->,>=latex] (xencpriorip) to (xencpriorfc);
		\draw[->,>=latex] (xencpriorfc) to (xencpriora);
		%	\draw[->,>=latex] (xencpriorip) to (xencpriora);
		\draw[->,>=latex] (xencpriorp) to (xencpriora);
		%	\draw[->,>=latex] (xencpriora) to (xencprior);
		
		\draw[-] (box) to [out=-80, in=0] (boxproxy);
		\draw[->,>=latex] (boxproxy) to [out=180, in=0] (xencmaskf);
		\draw[->,>=latex] (boxproxy) to [out=180, in=15] (xencmaskif);
		\draw[->,>=latex] (boxproxy) to [out=180, in=30] (xencmaskig);
		\draw[->,>=latex] (xencfull) to (xencmaskp);
		
		\draw[->,>=latex] (xencmaskf) to (xencmaskp);
		\draw[->,>=latex] (xencmaskif) to (xencmaskip);
		\draw[->,>=latex] (xencmaskig) to [out=180, in=-90] (xencmaskip);
		\draw[->,>=latex] (xencmaskip) to (xencmaskfc);
		\draw[->,>=latex] (xencmaskfc) to (xencmaska);
		\draw[->,>=latex] (xencmaskp) to (xencmaska);
		\draw[->,>=latex] (xencmaska) to (xencmask);
		
		\draw[-] (xencmaska) to [out=-90, in=-180] (xencmaskproxy);
		\draw[->,>=latex] (xencmaskproxy) to [out=0, in=-150] (xencf);
		\draw[->,>=latex] (xencmaskproxy) to [out=0, in=-165] (xencif);
		\draw[->,>=latex] (xencmaskproxy) to [out=0, in=-180] (xencig);
		
		\draw[->,>=latex] (xencf) to (xencp);
		\draw[->,>=latex] (xencif) to (xencip);
		\draw[->,>=latex] (xencig) to [out=0, in=-90] (xencip);
		\draw[->,>=latex] (xencip) to (xencfc);
		\draw[->,>=latex] (xencfc) to (xenca);
		%	\draw[->,>=latex] (xencip) to (xenca);
		\draw[->,>=latex] (xencp) to (xenca);
		\draw[->,>=latex] (xenca) to (xenc);
		
		%	\node (name0) at (-8.5, 11.35) {\LARGE Completion};
		%	\draw[draw=black, dashed] (-10, 6) rectangle (1.5, 11);
		
		\node (name) at (-9, 4.6) {\LARGE Decoder};
		\draw[draw=black, dashed] (-10, -8) rectangle (10, 4.2);
	\end{tikzpicture}
	\caption{\small
		Illustration of our decoding architecture. 
		%		The completion block infers the spatially complete observation $y_t$ from a partial observation $x_t$ and a latent state $z_t$.
		The decoder block infers $x_t$ the partial observation, $y_t$ the spatially complete observation and $m_t$ the masked $y_t$ (as dictated by the bounding box $a_{t}$). It takes as inputs a latent state $z_t$, a previous partial observation $x_{t-1}$ and a bounding box $a_t$. A rectangle indicates a fully connected layer, while the symbol at its center indicates the activation function applied to its output ($\sigma$ for sigmoid, \textit{tanh} for hyperbolic tangent and nothing for the identity function).
		Each updating network is composed of a forget gate (first $\sigma$) and a D map, i.e. input features (\textit{tanh}), an input gate (last $\sigma$) and a fully connected layer. 
		%		The inference of $y_t$ from $z_t$ is composed of input features, an input gate and a fully connected layer (D map). 
	}\label{fig:dec}
\end{figure}

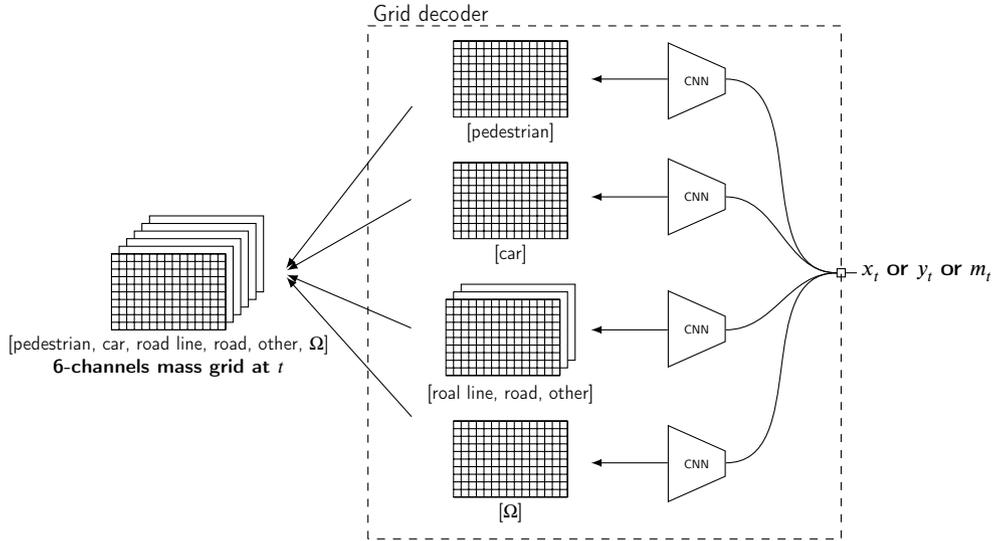
\begin{figure*}[t]
	\centering
	\begin{tikzpicture}[scale=0.5, every node/.style={transform shape},
		node/.style={draw, dot,minimum size=0.2cm, inner sep=0pt},
		det/.style={draw, diamond,minimum size=1.2cm, inner sep=1pt},
		rect/.style={draw, rectangle,minimum size=1.1cm, inner sep=2pt},
		var/.style={draw, circle, minimum size=1.1cm}
		]
		\pgfmathsetmacro{\cubex}{3}
		\pgfmathsetmacro{\cubey}{2}
		\pgfmathsetmacro{\cubez}{1}
		\pgfmathsetmacro{\offset}{0.2}
		%	\draw (0,0,0) -- ++(-\cubex,0,0) -- ++(0,-\cubey,0) -- ++(\cubex,0,0) -- cycle;
		%	\draw (0,0,0) -- ++(0,0,-\cubez) -- ++(0,-\cubey,0) -- ++(0,0,\cubez) -- cycle;
		%	\draw (0,0,0) -- ++(-\cubex,0,0) -- ++(0,0,-\cubez) -- ++(\cubex,0,0) -- cycle;
		\draw[fill=white] (5*\offset,5*\offset,0) -- ++(0,\cubey,0) -- ++(\cubex,0,0) -- ++(0,-\cubey,0) -- cycle;
		\draw[fill=white] (4*\offset,4*\offset,0) -- ++(0,\cubey,0) -- ++(\cubex,0,0) -- ++(0,-\cubey,0) -- cycle;
		\draw[fill=white] (3*\offset,3*\offset,0) -- ++(0,\cubey,0) -- ++(\cubex,0,0) -- ++(0,-\cubey,0) -- cycle;
		\draw[fill=white] (2*\offset,2*\offset,0) -- ++(0,\cubey,0) -- ++(\cubex,0,0) -- ++(0,-\cubey,0) -- cycle;
		\draw[fill=white] (\offset,\offset,0) -- ++(0,\cubey,0) -- ++(\cubex,0,0) -- ++(0,-\cubey,0) -- cycle;
		%	\draw[fill=white] (0,0,0) -- ++(0,\cubey,0) -- ++(\cubex,0,0) -- ++(0,-\cubey,0) -- cycle;
		\draw[fill=white] (0,0) grid[step=0.2] ++(\cubex,\cubey,0) rectangle (0,0);
		
		%	\draw[fill=white] (3*\cubex,3*\offset+2.5*\cubey,0) -- ++(0,\cubey,0) -- ++(\cubex,0,0) -- ++(0,-\cubey,0) -- cycle;
		\draw[fill=white] (3*\cubex,3*\offset+2.5*\cubey) grid[step=0.2] ++(\cubex,\cubey) rectangle (3*\cubex,3*\offset+2.5*\cubey);
		
		%	\draw[fill=white] (3*\cubex,1.5*\offset+1.1*\cubey,0) -- ++(0,\cubey,0) -- ++(\cubex,0,0) -- ++(0,-\cubey,0) -- cycle;
		\draw[fill=white] (3*\cubex,2*\offset+1*\cubey) grid[step=0.2] ++(\cubex,\cubey) rectangle (3*\cubex,2*\offset+1*\cubey);
		
		\draw[fill=white] (3*\cubex+\offset,-0.5*\cubey+\offset,0) -- ++(0,\cubey,0) -- ++(\cubex,0,0) -- ++(0,-\cubey,0) -- cycle;
		\draw[fill=white] (3*\cubex,-0.5*\cubey,0) -- ++(0,\cubey,0) -- ++(\cubex,0,0) -- ++(0,-\cubey,0) -- cycle;
		%	\draw[fill=white] (3*\cubex-\offset,-0.5*\cubey-\offset,0) -- ++(0,\cubey,0) -- ++(\cubex,0,0) -- ++(0,-\cubey,0) -- cycle;
		\draw[fill=white] (3*\cubex-\offset,-0.5*\cubey-\offset) grid[step=0.2] ++(\cubex,\cubey) rectangle (3*\cubex-\offset,-0.5*\cubey-\offset);
		
		%	\draw[fill=white] (3*\cubex,-2*\offset-2*\cubey,0) -- ++(0,\cubey,0) -- ++(\cubex,0,0) -- ++(0,-\cubey,0) -- cycle;
		\draw[fill=white] (3*\cubex,-2*\offset-2*\cubey) grid[step=0.2] ++(\cubex,\cubey) rectangle (3*\cubex,-2*\offset-2*\cubey);
		
		\pgfmathsetmacro{\paddingcnn}{1.325*\cubey}
		\pgfmathsetmacro{\offsetcnn}{-3.85*\cubey}
		
		\draw[fill=white] (4.95*\cubex-\offset,\offsetcnn+5*\paddingcnn,0) -- ++(0,\cubey,0) -- ++(\cubex/2,-\cubey/3,0) -- ++(0,-\cubey/3,0) -- cycle;
		\node () at (5.2*\cubex-\offset,\offsetcnn+0.5*\cubey+5*\paddingcnn) {CNN};
		
		\draw[fill=white] (4.95*\cubex-\offset,\offsetcnn+3.85*\paddingcnn,0) -- ++(0,\cubey,0) -- ++(\cubex/2,-\cubey/3,0) -- ++(0,-\cubey/3,0) -- cycle;
		\node () at (5.2*\cubex-\offset,\offsetcnn+0.5*\cubey+3.85*\paddingcnn) {CNN};
		
		\draw[fill=white] (4.95*\cubex-\offset,\offsetcnn+2.5*\paddingcnn+0.5*\offset,0) -- ++(0,\cubey,0) -- ++(\cubex/2,-\cubey/3,0) -- ++(0,-\cubey/3,0) -- cycle;
		\node () at (5.2*\cubex-\offset,\offsetcnn+0.5*\cubey+2.5*\paddingcnn+0.5*\offset) {CNN};
		
		\draw[fill=white] (4.95*\cubex-\offset,\offsetcnn+1.2*\paddingcnn,0) -- ++(0,\cubey,0) -- ++(\cubex/2,-\cubey/3,0) -- ++(0,-\cubey/3,0) -- cycle;
		\node () at (5.2*\cubex-\offset,\offsetcnn+0.5*\cubey+1.2*\paddingcnn) {CNN};
		
		\node (grid) at (1.5*\cubex, 2.5*\offset+\cubey/2) {};
		%	\node () at (6*\cubex+\offset/2, 5*\cubey) {\Large locally planned path};
		%	\node[rect] () at (8*\cubex+\offset/2, 5*\cubey) {\Large 2-layers FC};
		%	\node () at (6*\cubex+\offset/2, 4.25*\cubey) {\Large[control, box]};
		\node () at (0.5*\cubex, -2*\offset) {\Large[pedestrian, car, road line, road, other, $\Omega$]};
		\node () at (0.5*\cubex, -5*\offset) {\Large\textbf{6-channels mass grid at $t$}};
		\node () at (3.5*\cubex, \offset+2.5*\cubey) {\Large[pedestrian]};
		\node () at (3.5*\cubex, 1.1*\cubey-1*\offset) {\Large[car]};
		\node () at (3.5*\cubex, -\offset -0.75*\cubey) {\Large[roal line, road, other]};
		\node () at (3.5*\cubex, -2*\cubey-4*\offset) {\Large[$\Omega$]};
		\node (ped) at (2.5*\cubex+2.5*\offset, 3.*\cubey) {};
		\node (car) at (2.5*\cubex+2.5*\offset, 2.5*\offset+1.5*\cubey) {};
		\node (lay) at (2.5*\cubex+2.5*\offset, 0) {};
		\node (om) at (2.5*\cubex+2.5*\offset, -2.*\offset-1.*\cubey) {};
		\node (pedo) at (4.*\cubex+2.5*\offset, 0.5*\offset+3.25*\cubey) {};
		\node (caro) at (4.*\cubex+2.5*\offset, 2.5*\offset+1.5*\cubey) {};
		\node (layo) at (4.*\cubex+2.5*\offset, 0) {};
		\node (omo) at (4.*\cubex+2.5*\offset, -2.5*\offset-1.5*\cubey) {};
		\node (pcnn) at (4.75*\cubex+2.5*\offset, 3*\offset+3*\cubey) {};
		\node (ccnn) at (4.75*\cubex+2.5*\offset, 2.5*\offset+1.5*\cubey) {};
		\node (lcnn) at (4.75*\cubex+2.5*\offset, 0) {};
		\node (omcnn) at (4.75*\cubex+2.5*\offset, -2.5*\offset-1.5*\cubey) {};
		
		\node (pcnno) at (5.15*\cubex+3*\offset, 3*\offset+3*\cubey) {};
		\node (ccnno) at (5.15*\cubex+3*\offset, 2.5*\offset+1.5*\cubey) {};
		\node (lcnno) at (5.15*\cubex+3*\offset, 0) {};
		\node (omcnno) at (5.15*\cubex+3*\offset, -2.5*\offset-1.5*\cubey) {};
		
		\node[draw, rectangle] (xenc) at (6*\cubex+6*\offset, 2.5*\offset+\cubey/2) {};
		\node (xenclabel) at (6.5*\cubex+6*\offset+0.75, 2.5*\offset+\cubey/2) {\LARGE $x_t$ \textbf{or} $y_t$ \textbf{or} $m_t$};
		
		\draw[->,>=latex] (ped) to (grid);
		\draw[->,>=latex] (car) to (grid);
		\draw[->,>=latex] (lay) to (grid);
		\draw[->,>=latex] (om) to (grid);
		\draw[->,>=latex] (pcnn) to (pedo);
		\draw[->,>=latex] (ccnn) to (caro);
		\draw[->,>=latex] (lcnn) to (layo);
		\draw[->,>=latex] (omcnn) to (omo);
		\draw[-] (xenclabel) to (xenc);
		\draw[-] (pcnno) to [out=0, in=180] (xenc);
		\draw[-] (ccnno) to [out=0, in=180] (xenc);
		\draw[-] (lcnno) to [out=0, in=180] (xenc);
		\draw[-] (omcnno) to [out=0, in=180] (xenc);
		\node (name) at (2.8*\cubex, 8.35) {\LARGE Grid decoder};
		\draw[draw=black, dashed] (2.25*\cubex, -5.5) rectangle (6*\cubex+6*\offset ,8);
	\end{tikzpicture}
	\caption{\small
		Illustration of the decoding of $X_t$ or $Y_t$ or $M_t$ by the decoder of the CVAE that gave $X_t$ to get back into the observation space. The CNN blocks are Transposed CNNs.
	}\label{fig:griddec}
\end{figure*}

So far, we determined the networks outputting distribution parameters describing the latent states $Z$ used in the evaluation of $\mathcal{L_{\text{LP-VAE}}}$, both for the generative model and the recognition model. It remains to propose the decoding network that is part of the generative model and produces $X$ and $Y$. Given the conditional distributions appearing in $\mathcal{L_{\text{LP-VAE}}}$, we need 
%a completion network inferring $Y_t$ from $X_t$ and $Z_t$, as well as 
a decoder inferring $Y_t$ from $Z_t$ and another one inferring
%then $M_t$ from $Y_t$ and $A_t$, and finally 
$X_t$ from $X_{t-1}$, $Y_t$, $A_t$ and $Z_t$. 

However, since $X_t$ and $Y_t$ are not given in the original space but in a learned compressed one, extracting features from $Y_t$ according to the bounding box $A_t$ is not directly possible. One has to decode $Y_t$, extract features according to $A_t$, decode $X_t$ and then fuse it with the leaked features from $Y_t$. For the sake of efficiency, we will learn to directly extract these features that we denote by the random variable $M_t$ in the learned compressed space and to fuse them with $X_t$. Thus, in parallel to $\mathcal{L_{\text{LP-VAE}}}$, we minimize an extra loss term $$-\sum_{k=2}^{T} \log p_{M_i|A_{i},Y_i}\left(m_k | a_k, y_{k};\theta\right),$$
where $m_t$ corresponds to $y_t$ masked in accordance with $a_t$ and compressed by the same CVAE as for $y_t$. Note that our dataset becomes $D = (x_{1:T}, ~y_{1:T}, m_{2:T}, a_{2:T})_{1:N}$.

We choose to infer $Y_t$ from $Z_t$ through a D map as introduced in section \ref{inference}. All other inferences are done through an updating module that is inspired by the updating of a LSTM cell state. 
%The completion network updates $X_t$ with the knowledge of $Z_t$ to produce the spatially complete observation $Y_t$. 
The masking of $Y_t$ is orchestrated by $A_t$, producing $M_t$ by filtering. Finally, $X_{t-1}$ is updated in two steps. The first update is assumed to change its reference frame and to determine which parts of $Y_t$ are visible to the ego-vehicle. This implicitly produces the $X_t$ corresponding to the null action, i.e. the action that consists in doing nothing. We consider this transformation deterministic, given $y_t$ and $z_t$. The second update transmits the excerpt $M_t$ from $Y_t$ to this prior perception, producing the actual $X_t$ influenced by $A_t$. Fig. \ref{fig:dec} depicts these networks. In addition, Fig. \ref{fig:griddec} illustrates the decoding of $X_t$ by the decoder of the Convolutional VAE.

\section{Experiments}\label{exp}

\subsection{Data acquisition \& RL Environment}\label{data}

To conduct our experiments, we chose to work with the open-source driving simulator CARLA \cite{carla}. 
%represent the perceptual field of the ego-vehicle with a top-down semantic 4-channels plausibility grid, noted $G_t$, corresponding to the four classes of the frame of discernment $\Omega = \{ \textit{pedestrian}, \textit{car}, \\\textit{road}, \textit{other} \}$. 
Our semantic grids $G_t$ are computed online from a frontal $320\times480$ depth camera with FOV of $135^{\circ}$ and its corresponding pixel-wise semantic classification. These simulated sensors are attached to a simulated vehicle autonomously wandering in a city with other vehicles, bikes and pedestrians (see Fig. \ref{grid}). More precisely, $G_t$ is obtained by counting the number of occurrences of each class in each possible configuration of $4\times4$ consecutive pixels. All classes corresponding to static objects are merged into the class \textit{other}. Then, in each cell of the resulting $80\times120\times5$ grid, these numbers are divided by 16 and we add a channel representing ignorance (i.e. $\Omega$) to store the quantity needed to make the sum on all channels equal to 1. We also discount the resulting mass functions by a factor of $0.01$ to simulate noise, i.e. all masses are multiplied by $0.99$ and $0.01$ is added to the mass on ignorance. Finally, thanks to the depth and information about the camera, we create a 3D point cloud of this frontal perception. Thus, to get the 2D grid $G_t$, we ignore points higher than 2.5 meters and we take the highest of the remaining ones (if more than one point at the same ground coordinates). For this reason, it sometimes happens that the ground under a vehicle is perceived, but not its top, leading to \textit{road} cells surrounded by \textit{car} cells, as can be observed in Fig. \ref{grid} Left. An important road elevation may also conflict with the threshold of 2.5 meters. 
This view can be obtained by a LIDAR and a 3D semantic classifier \cite{li2020deep} as well.

Our top-down semantic grids corresponding to complete observations $y_t$ in our model are obtained with a facing ground camera above the ego-vehicle. Doing so, it contains itself some occlusions due to trees, poles, buildings, etc. Thus, it is rather a hint about the true $y_t$. This grid can also be obtained by the fusion of multiple view points, from a fleet of autonomous vehicles or infrastructure sensors, which can be acquired in the real world. A drone may be able to acquire this information as well.
%Visualization of the 12-channels top-down plausibility grid before reduction to 5 channels. It is computed online from a frontal depth camera and its corresponding pixel-wise 12-classes semantic segmentation generated by the driving simulator CARLA \cite{carla} and a simulated vehicle autonomously wandering in a city with other vehicles, bikes and pedestrians.
%Instance of top-down semantic grid corresponding to a complete observation $y_t$ in our model. Actually, this view is obtained with a facing ground camera above the ego-vehicle. Doing so, it contains itself some occlusions due to trees, poles, buildings, etc. Thus, it is rather a hint about the true $y_t$. This view can also be obtained by the fusion of multiple view points, from autonomous vehicles or infrastructure sensors.
%
%In a real-world framework, this top-down view can be the result of the fusion of multiple sensors, even other cars, to simulate knowledge coming from a vehicular network. 
In any case, this ground truth grid is in fact itself uncertain and so is computed as $G_t$ with an ignorance channel.
%, even in CARLA as there always are traffic lights, street lamps, terrain elevation changes, etc, that occlude some parts of the grid. 

We created a dataset composed of 1560 sequences of 50 timesteps (5 seconds) each, where each perception is $80\times120\times6$. There are 30 runs in each of four cities available in CARLA, including small towns, big towns and fast lanes. Each run is 35 seconds long and a sequence is recorded every 2.5 seconds, leading to 13 sequences per run, hence the size of our dataset. This dataset provides the grids corresponding to $X_t$ and $Y_t$ in the action-independent model of section \ref{ai-model}.

To provide the grids corresponding to $X_t$ as defined in the full model of section \ref{full-model}, we created a second dataset from the first one by choosing random regions of $Y_t$ to be given to $X_t$. We also added a visual memory that keeps a buffer of grid cells, transforms their coordinates according to the given motion of the ego-vehicle, discounts their mass functions to account for information ageing and fuses them with the current perception grid, resulting in this $X_t$. 
%by randomly choosing a bounding box at each time step (with the corresponding perception fusion 5 time steps later) every time a sample is taken from the dataset. 
In fact, the first dataset combined with our visual memory and our fusion procedure of Algorithm \ref{algo:fusion} for $\widetilde{G}_t$ and $G^M_t$ constitutes the environment in which our agent will learn a communication policy. %However, testing is done directly in CARLA.

%
%\newcommand{\plG}[1]{\sloppy\text{$G^{\textit{Pl}}_{#1}$}}
%
%From there, we feed a 4-channels plausibility grid $\plG{t}$, computed from the 12-channels semantic grid such that the first three channels correspond to the plausibility of respectively the pedestrian, car and road classes. The last channel corresponds to the plausibility of every other classes. We have, for each cell $c$ of the grid:
%\begin{align*}
%\forall \omega \in \Omega_\textit{Pl} \cup \{\textit{other}\},~ \plG{t}[\omega, c] = 
%\begin{cases}
%\textit{Pl}_{G_t[c]}(\{\omega\})	&\text{if $\omega \in \Omega_\textit{Pl}$}\\
%\textit{Pl}_{G_t[c]}(\Omega\backslash\Omega_\textit{Pl})	&\text{otherwise}
%\end{cases}
%\end{align*}
%where $\Omega_\textit{Pl} = \{ \textit{pedestrian}, \textit{car}, \textit{road} \}$ and $G_t[c]$ represents the mass function in cell $c$ at time $t$.
%
%We feed the TD-VAE with both $G_t$ and the $2.N + 2$ actions from the driving policy. We expect from it to be able to predict future observations and to infer the class of unkown cells. For the latter, contrary to the original TD-VAE, we train the network to reconstruct future images generated in CARLA by a camera above the ego-vehicle acquiring a top-down view of its surroundings, instead of its original input. 

\subsection{Models}

During training, we give between 8 and 10 timesteps of observations (i.e. between 0.8 and 1 second) and it is asked to predict between 5 and 10 timesteps ahead, i.e. between 0.5 and 1 second. 
We use the Mean Squared Error (MSE) loss function to compute the Gaussian negative log likelihoods of observing the grids corresponding to $x_t$ and $m_t$ given latent states. Indeed, this is analog to taking $\alpha = \frac{1}{2}$ and ignoring the constant term $\log\left(\sqrt{2\pi\alpha}\right)$.
For the negative log likelihoods on the grid corresponding to $y_t$, we binarize it by taking the class with maximum mass and use a cross-entropy loss. To account for the fact that the instances of $Y_t$ in our dataset are not perfect, we simply do a pointwise multiplication between this loss and the complement to 1 of its ignorance channel (last channel). That way, if $y_t$ does not have any information about a cell, no loss on $y_t$ is actually back-propagated.
%the cross-entropy loss mentioned in section \ref{overview} by a confidence factor $1-U$, where $U = \displaystyle\sum_{i=2}^{|\Omega|} u[i-2] . \log_{|\Omega|} (i)$ and $u$ is the vector $u$ obtained by computing the pseudo-consonant mass function corresponding to the ground truth contour grid through Algorithm \ref{algo:pltom}.
Furthermore, we weight this cross-entropy loss differently from one channel to another to account for class imbalance. We used the weight vector $[100, 10, 1, 0.2, 0.1, 1]$. Indeed, on average, there are far less cells containing pedestrians than cells containing the road or any other static class. Doing so, without weights, the network would consider pedestrian as noise and neglect them. 

In the following, we compare STD-VAE and LP-VAE for complete grid inference and prediction.

\subsubsection{Grid completion}

%- grid completion ($y_t | x_t, z_t$) : MSE
%
%['TD-VAE', 'LP-VAE'] classif_scores = [0.6879875369421585, 0.6831295350300894]
%['TD-VAE', 'LP-VAE'] classif_weighted_scores = [0.567148232747743, 0.5358727068721482]
%['TD-VAE', 'LP-VAE'] classif_per_class = [array([0.33653232, 0.72717255, 0.30700562, 0.85850399, 0.80569582,
%0.46165827]), array([0.20544966, 0.68526309, 0.28715311, 0.84319565, 0.77770921,
%0.49460436])]

In this experiment, we use the decoder network described in Fig. \ref{fig:dec} on the current latent state $Z_t$ inferred from $B_t$ to retrieve $Y_t$. Then, we use the network described in Fig. \ref{fig:griddec} to transform $Y_t$ into the complete mass grid ${G^Y}$.
To compare STD-VAE and LP-VAE, we employed two metrics: binary classification accuracy per class and a \textit{mass score}.
Our \textit{Mass score} metric is computed as the mean of $G^Y_t ~.~ \widehat{G^Y}_t$ over all cells in the grid, where $G^Y_t$ is the true binary complete grid classification and $\widehat{G^Y}_t$ is a mass grid inferred by some model. Since $G^Y_t$ is binary, it acts as an indicator function for the correct class and the mass score represents the mean mass given to the right class by the model generating $\widehat{G^Y}_t$. Results are showed in Table \ref{table:bin_classif}.

\begin{table*}[t]
	\begin{center}
		\begin{tabular}
			%	{0.8\textwidth} { 
				%		| >{\raggedright\arraybackslash}X 
				%		| >{\centering\arraybackslash}X 
				%		| >{\raggedleft\arraybackslash}X | }
			{ |p{1.6cm}||p{1cm}|p{1cm}|p{1cm}|p{1cm}|p{1cm}|p{1cm}||p{1cm}| }
			\hline
			\multicolumn{1}{|c|}{} & \multicolumn{6}{|c|}{Binary classification per class} & \multicolumn{1}{|c|}{Mass}\\
			%			\hline
			& P & C & RL & R & O & $\Omega$& ~score\\
			\hline
			LP-VAE & 20.5\% & 68.5\% & 28.7\% & 84.3\% & 77.8\% & \textbf{49.5\%} & 68.3\% \\
			STD-VAE & \textbf{33.7\%} & \textbf{72.7\%} & \textbf{30.7\%} & \textbf{85.9\%} & \textbf{80.6\%} & 46.2\% & \textbf{68.8\%} \\
			\hline
		\end{tabular}
	\end{center}
	\caption{
		Mass score and binary classification accurracy per class. \textit{P} indicates the pedestrian channel, \textit{C} the car channel, \textit{RL} the road lines channel, \textit{R} the road channel, \textit{O} the other channel and $\Omega$ the complete out-of-sight channel. It is clear that STD-VAE outperforms LP-VAE for simple grid completion, though the total mass score is not so different.
	}\label{table:bin_classif}
\end{table*}
%\begin{table}[t]
%	\begin{center}
	%		\begin{tabular}
		%			%	{0.8\textwidth} { 
			%			%		| >{\raggedright\arraybackslash}X 
			%			%		| >{\centering\arraybackslash}X 
			%			%		| >{\raggedleft\arraybackslash}X | }
		%			{ |p{1.15cm}||p{0.6cm}|p{0.6cm}|p{0.6cm}|p{0.6cm}|p{0.6cm}|p{0.6cm}||p{0.6cm}| }
		%			\hline
		%			\multicolumn{1}{|c|}{} & \multicolumn{6}{|c|}{Binary classification per class} & \multicolumn{1}{|c|}{Mass}\\
		%			%			\hline
		%			& P & C & RL & R & O & $\Omega$& ~score\\
		%			\hline
		%			\scriptsize LP-VAE &\scriptsize 20.5\% &\scriptsize 68.5\% &\scriptsize 28.7\% &\scriptsize 84.3\% &\scriptsize 77.8\% &\scriptsize \textbf{49.5\%} &\scriptsize 68.3\% \\
		%			\tiny STD-VAE &\scriptsize \textbf{33.7\%} &\scriptsize \textbf{72.7\%} &\scriptsize \textbf{30.7\%} &\scriptsize \textbf{85.9\%} &\scriptsize \textbf{80.6\%} &\scriptsize 46.2\% &\scriptsize \textbf{68.8\%} \\
		%			\hline
		%		\end{tabular}
	%	\end{center}
%	\caption{
	%		Mass score and binary classification accurracy per class. \textit{P} indicates the pedestrian channel, \textit{C} the car channel, \textit{RL} the road lines channel, \textit{R} the road channel, \textit{O} the other channel and $\Omega$ the complete out-of-sight channel. It is clear that STD-VAE outperforms LP-VAE for simple grid completion, though the total mass score is not so different.
	%	}\label{table:bin_classif}
%\end{table}

\subsubsection{Prediction}

%- prediction : qualitative analysis + quantitative
%
%Sur 49 920 pas de temps, une masse totale de 3 434 985 change positivement de façon binaire d'une frame à l'autre, soit en moyenne 68.81 cellules dont la masse sur $\{\textit{road lines}, \textit{road} \}$ change de 0 à 1.
%Sur 49 920 pas de temps, une masse totale de 3 353 706 change négativement de façon binaire d'une frame à l'autre, soit en moyenne 67.18 cellules dont la masse sur $\{\textit{road lines}, \textit{road} \}$ change de 1 à 0. 

In this experiment, we compare prediction accuracy between LP-VAE and STD-VAE. For this, we study mass variations on the super-class $\{\textit{road}, \textit{road line}\}$, i.e. the sum of the \textit{road} and \textit{road line} grid channels. Indeed, this super-class represents the road layout. Its absence in a cell indicates either road users or the \textit{other} class. Thus, its mass variations accounts for the dynamics of the whole scene, independently of classification accuracy.

In practice, for each model, we infer a prediction sequence of 10 complete grids $\hat{y}_{1:10}$ (i.e. 1 second in the future), based on 10 observations (i.e. the past second). From it, we compute the corresponding sequence of 9 grid variations $\hat{y}'_t = \hat{y}_{t+1} - \hat{y}_{t}$. We execute the same process with the true complete grids, which produces grids $y'_{1:9}$ of values ranging in $\{-1, 0, 1\}$. We test separately the accuracy on positive and negative changes. For the former, we do a pointwise multiplication between the true complete positive grids $\max(0, y'_{1:9})$ and the inferred positive ones $\max(0, \hat{y}'_{1:9})$. For the latter, we do a pointwise multiplication between the true complete negative grids $\max(0, -y'_{1:9})$ and the inferred negative ones $\max(0, -\hat{y}'_{1:9})$. We then sum all cells of each grid in the sequence, over 4992 sequences, i.e. 49 920 inferred grids and compare it to the separate sums of positive and negative true changes.
Results are displayed in the first two columns of Table \ref{table:pred}.

\begin{figure}[t]
	\centering
	\begin{subfigure}{.475\textwidth}
		\centering
		\includegraphics[width=\linewidth]{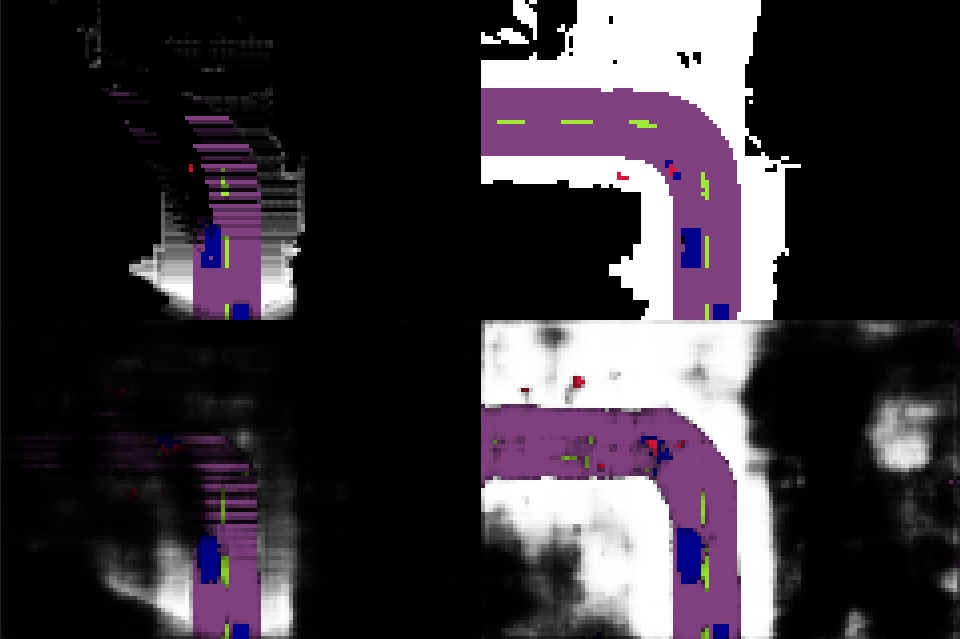}
		\caption{}
		\label{fig:pred-lpvae}
	\end{subfigure}
	\begin{subfigure}{.475\textwidth}
		\centering
		\includegraphics[width=\linewidth]{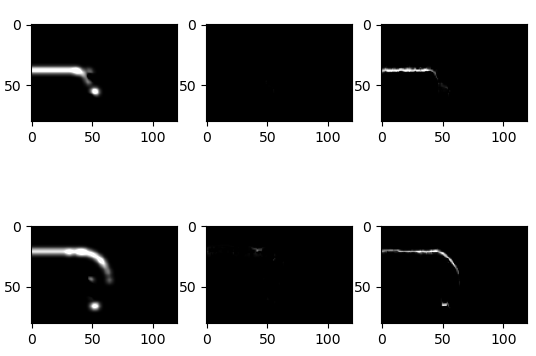}
		\caption{}
		\label{fig:dynamics}
	\end{subfigure}
	\caption{(a) Left column: partial grid $G_t$ corresponding to $X_t$. Right column: complete grid $G^Y_t$ corresponding to $Y_t$. Top row: true classification grids. Bottom row: classification grids predicted by LP-VAE from $X$ alone, 4 time steps in the future.
		(b) Prediction dynamics. Black represents the absence of variation, white some mass change in the cells of the \textit{road} and \textit{road line} channels of the grid in (a). Left column corresponds to the true variations, blurred by a 11x11 Gaussian filter. The central column corresponds to the prediction dynamics of STD-VAE, multiplied by the ones of the first column. Same for the right column but for LP-VAE. The first row represents positive changes, while the second row represents negative ones.}
	\label{fig:pred}
\end{figure}

However, note that this binary mask can be quite hard to match, as both the exact location of these changes and their amplitude must be correct. To alleviate this constraint, we repeat this test with blurring filters applied to each grid of $y'_{1:9}$. The resulting grids, noted $\tilde{y}'_{1:9}$, are then renormalized so that $\sum \max(0, y'_{1:9}) . \max(0, \tilde{y}'_{1:9}) = \sum \max(0, y'_{1:9})$ and $\sum \max(0, -y'_{1:9}) . \max(0, -\tilde{y}'_{1:9}) = \sum \max(0, -y'_{1:9})$. This allows for slight misplacements of cells in predicted grids. We repeated this test twice with Gaussian filters, with kernels 5x5 and 11x11. These experiments correspond to the last 4 columns of Table \ref{table:pred}. Our LP-VAE outperforms STD-VAE in every of these tests, no matter how hard the constraint on change location is. This means that the predicted changes of LP-VAE are not just better located, but also \textit{better shaped} than the ones of STD-VAE, as expected by design. Fig. \ref{fig:pred} illustrates this experiment.
%Fig. and illustrate what we mean by \textit{better shaped} changes.

%Ground truth
%3434985.0
%3353706.0
%
%TD-VAE
%71969.52782914718
%=> 0,020951919
%79613.57357588685
%=> 0,023738984
%
%
%LP-VAE
%234005.00857210357
%=> 0,068124026
%232762.09631487186
%=> 0,069404443
%
%3434985.0
%3353706.0
%71969.52782914718
%79613.57357588685
%49920 LP-VAE
%234005.00857210357
%232762.09631487186
%
%exp 2 avec Gaussian blur 5x5, t total = 49920:
%3434917.997479439
%3353610.998295605
%TD-VAE
%168067.0665644549
%=> 0,048928989
%181486.50885378465
%=> 0,054116744
%LP-VAE
%495111.78843807685
%=> 0,144140788
%489985.41351557226
%=> 0,146106813
%
%exp 3 avec Gaussian blur 11x11, t total = 49920:
%
%3434648.0022539496
%3353209.006379068
%TD-VAE
%297423.66317530116
%=> 0,086595093
%319202.4114273727
%=> 0,095193115
%49920 LP-VAE
%817618.4321520893
%=> 0,238050138
%816684.7064946829
%=> 0,243553177

\begin{table*}[t]
	\begin{center}
		\begin{tabular}
			%	{0.8\textwidth} { 
				%		| >{\raggedright\arraybackslash}X 
				%		| >{\centering\arraybackslash}X 
				%		| >{\raggedleft\arraybackslash}X | }
			{ |p{2cm}||p{1cm}|p{1cm}||p{1cm}|p{1cm}||p{1cm}|p{1cm}|  }
			\hline
			\multicolumn{1}{|c|}{True $y'$} & \multicolumn{2}{|c|}{No blur} & \multicolumn{2}{|c|}{Gaussian blur 5x5} & \multicolumn{2}{|c|}{Gaussian blur 11x11}\\
			\hline
			& + & - & + & - & + & -\\
			\hline
			LP-VAE $\hat{y}'$ & \textbf{6.81\%} & \textbf{6.94\%} & \textbf{14.41\%} & \textbf{14.61\%} & \textbf{23.81\%} & \textbf{24.36\%} \\
			STD-VAE $\hat{y}'$ & 2.10\% & 2.37\% & 4.89\% & 5.41\% & 8.66\% & 9.52\% \\
			\hline
		\end{tabular}
	\end{center}
	\caption{
		Prediction accurracies between STD-VAE and LP-VAE. As expected, LP-VAE significantly outperforms STD-VAE on predictions.
	}\label{table:pred}
\end{table*}

\subsection{Policy learning}

Here, we finally compare different policies learned with PPO, with and without model to test the benefits of using belief states in our case. Each policy is the best found among iterations of training with 3000 transitions amounting to 500 000 time steps in total. We used a batch size of 60, with 10 epochs on each transition dataset, with a learning rate of 0.0003 and an entropy coefficient of 0.01. We also made the time horizon vary, i.e. we made the hyperparameter $\gamma$ vary from 0 to $0.7$, in order to see if a medium/long term strategy performs better.

The network learned with PPO has two parts: one for inferring the Value of a state, representing the mean of all potential future rewards, and one for inferring the best action from this same state, representing the policy. Each of these networks is composed of two fully connected hidden layers of 128 and 64 neurons.

Different communication behaviors can be obtained by adjusting reward parameters. In particular, increasing $K$ in Eq. \ref{reward} will make requests bigger, increasing $w$ in Eq. \ref{reward_density} will make requests more focused on completely unknown areas, increasing $\eta$ will make requests more focused on \textit{pedestrians} and \textit{cars}, less rewarding in general and so less frequent. We chose the following values: $\eta = 0.3$, $K = 36$ and $w = 2$. We also added a penalty of -15 for no cooperation at all (i.e. choice of a bounding box with no pixel in it, which means no transmission cost either) to force the agent to play the game. Moreover, approximating the top-down dimensions of cars and pedestrians, we took the following reward densities per squared meter:
$r^m_{\textit{obj}} = [540 / (0.7*1.6), 540 / (3*1.8), 20, 20, 0]$. Then, we converted them into rewards per squared cell by multiplying them by our grid resolution. More precisely, we set our cameras in CARLA so that the height corresponds to 40 meters. Thus, our reward densities per squared cell are $r_{\textit{obj}} = (\frac{40}{80})^2 . r^m_{\textit{obj}}$. Our final rewards are obtained by normalizing $r_{\textit{obj}}$ to $[0, 1]$ by dividing it by its maximum.
For the spatial filter, we used the parameters of Fig. \ref{spatial-filter}, i.e. $\alpha=0.5$, $\beta_F=0.8$, $\beta_L=1$ and $\zeta=0.01$.

In order to evaluate and compare the performance of different policy learning schemes, we take as metrics the mean request size and the mean informational gain over all time steps of a test set with same size and characteristics as the training set described Section \ref{data}. We applied these metrics to 3 class groups: \textit{pedestrians} (P), \textit{cars} (C) and $\{\textit{road lines}, \textit{road} \}$ (R). 
In these conditions, we compared 3 schemes: PPO on top of the LP-VAE belief state $B_t$, PPO on top of the STD-VAE belief state $B_t$ and PPO on top of $X_t$ alone (i.e. only the features extracted from the current mass grid $G_t$ by a Convolutional VAE). Each of them has been trained with $\gamma = 0$ (i.e. only immediate rewards matter), $\gamma = 0.35$ and $\gamma = 0.7$, to see if we could benefit from medium/long term strategies. We also compare these policies with a simple random policy that has a 50\% chance of making a request and chooses uniformly random size and position of bounding box when it does. Table \ref{table:policy} presents our results, in percentage relatively to the maximal information gain and request size possible inherent to a broadcasting policy.

All of our learned policies only ask for about 5\% of the space around the ego-vehicle, while receiving about 25\% of the relevant information the agent lacks. Requiring about 2.5 times more information from the vehicular network for about the same relevant information gain or lower, the random policy is vastly less efficient. It only outperforms the others for pedestrians, which is consistent with the highly random behavior of pedestrians in CARLA. However, PPO + $X_t$ alone and $\gamma = 0$ (i.e. greedy policy) is the policy that performs best overall. Surprisingly enough, taking into account future rewards actually harms performance in our case. A lower discounting factor in the memory module (i.e. observations that are kept longer in memory) would probably make policies perform best with $\gamma > 0$. Furthermore, note that LP-VAE always performs better than the other learned policies when $\gamma > 0$. This is consistent with the fact that LP-VAE has better prediction capabilities and thus provides useful information in its belief state for predicting future rewards.

\begin{table*}
	\begin{center}
		\begin{tabular}
			%	{0.8\textwidth} { 
				%		| >{\raggedright\arraybackslash}X 
				%		| >{\centering\arraybackslash}X 
				%		| >{\raggedleft\arraybackslash}X | }
			{ |p{1.3cm}|p{2.1cm}||p{0.85cm}|p{0.85cm}|p{0.85cm}|p{1.2cm}|}%p{0.85cm}|p{0.85cm}|p{0.85cm}|p{1.5cm}|  }
		%	\hline
		%	\multicolumn{1}{|c|}{} &\multicolumn{4}{|c|}{$\gamma = 0$} &\multicolumn{4}{|c|}{$\gamma = 0.7$} \\
		\hline
		& & \multicolumn{3}{|c|}{Information gain} & Request\\% & \multicolumn{3}{|c|}{Info gain} &  Request\\
		& & P & C & R & \quad size\\% & P & C & R & size\\
		\hline
		& Random& \textbf{26.2\%} & 22\% & 22.9\% & 13\% \\%& 26.2\% & 22\% & 22.9\% & 13\%\\
		\hline
		& LP-VAE $B_t$ & \textbf{22}\% & 27.6\% & 26.5\% & 6\% \\%& \textbf{15.7\%} & \textbf{18.2\%} & \textbf{19.5\%} & 5\%\\
		\centering$\gamma = 0$ & STD-VAE $B_t$ & 19.9\% & 26.5\% & 24.4\% & 5\% \\%& 13.6\% & 16.3\% & 17.2\% & \textbf{4\%}\\
		& $X_t$ alone & 21.7\% & \textbf{29.2\%} & \textbf{27.6\%} & 6\% \\%& 14.3\% & 17.8\% & 18.6\% & \textbf{4\%}\\
		\hline
		& LP-VAE $B_t$ & \textbf{20.6\%} & \textbf{25.7\%} & \textbf{24.8\%} & 6\%\\
		\centering$\gamma = 0.35$ & STD-VAE $B_t$ & 18.2\% & 22.8\% & 23.3\% & 5\%\\
		& $X_t$ alone & 17.8\% & 23.4\% & 22.3\% & 5\%\\
		\hline
		& LP-VAE $B_t$ & \textbf{15.7\%} & \textbf{18.2\%} & \textbf{19.5\%} & 5\%\\
		\centering$\gamma = 0.7$ & STD-VAE $B_t$ & 13.6\% & 16.3\% & 17.2\% & 4\%\\
		& $X_t$ alone & 14.3\% & 17.8\% & 18.6\% & 4\%\\
		\hline
	\end{tabular}
\end{center}
\caption{
	Learned communication policy performances relatively to a broadcasting policy. The information gain is a mean percentage representing the mass actually gained after request, over the total mass that can be gained, at each time step.
}\label{table:policy}
\end{table*}

%gamma = 0.35
%total = (2.86, 41.72, 1562.60)
%PPO-0 || PPO-1 || PPO-2 | GAIN: (0.52, 9.53, 363.60) || (0.59, 10.71, 388.19) || (0.51, 9.78, 348.82), COST: 0.13 || 0.05 || 0.06 || 0.05
%
%exp 2 :
%'PPO-logs/best_model_cell_cost_0.3_no_req_15_tdvae_b',
%'PPO-logs/best_model_cell_cost_0.3_no_req_15_lpvae_b',
%'PPO-logs/best_model_cell_cost_0.3_no_req_15_lpvae_b_bigger_network',
%'PPO-logs/best_model_cell_cost_0.3_no_req_15_lpvae_rm',
%'PPO-logs/best_model_cell_cost_0.3_no_req_15_x',
%
%Random || PPO-0 || PPO-1 || PPO-2 || PPO-3 || PPO-4 | GAIN: (2.86, 41.72, 1562.60) || (0.75, 9.06, 352.29) || (0.57, 11.04, 380.72) || (0.63, 11.49, 414.44) || (0.59, 11.88, 390.45) || (0.65, 9.92, 394.06) || (0.62, 12.22, 431.10), COST: 0.12 || 0.05 || 0.06 || 0.05 || 0.06 || 0.06

\section{Conclusions}\label{com:conclusion}

In this paper, we tried to elaborate an efficient peer-to-peer communication policy for collaborative perception. For this, we made agents learn what could be hidden in their blind spots through a generative sequence model that we proposed, named Locally Predictable VAE (LP-VAE). We compared its performance with another generative sequence model for RL applications called TD-VAE that we slightly adapted to our problem by making it both jumpy and sequential, referring to it as STD-VAE. We demonstrated that LP-VAE produces better predictions than STD-VAE, which translated into better performance for policies learned on top of its belief state. However, we discovered in the end that our best communication policy was a greedy one, i.e. one that does not need prediction capabilities. Combined with the fact that we augmented each observation with the discounted memories of past observations, it followed that only a state-less Convolutional VAE was needed for this greedy policy.
Overall, our best learned policies only require about 5\% of the space around the ego-vehicle, while gaining about 25\% of the relevant information the agent lacks. Thus, we proved that learning to value the unknown is much more efficient than employing a broadcasting policy. It is also more efficient than blindly asking for random areas around the ego-vehicle since it requires about 13\% of the total information, while gaining less than 25\% of the relevant information the agent lacks.
In addition, we defined interpretable hyperparameters shaping the reward function corresponding to our problem. This makes it possible to obtain various communication policies, with different trade-offs between request size and information gain, as well as different class valuations, spatial priorities and valuation of ignorance (i.e. more or less emphasis on total ignorance).
For future works, it would be interesting to compare LP-VAE and STD-VAE in RL tasks where future rewards are more important. Also, we would like to test our communication policies in a truly multi-agent context, where the agent would need to take into account the availability of nearby communicating vehicles, and with real sensor data.

\begin{appendices}
	%============================================================================
	
	\section{LP-VAE loss}\label{app:lpvae-loss}
	
	\subsection{Minimization of $D_{KL}\left( Q_t(\theta, \phi) ~||~ P_t(\theta) \right)$}
	\begin{proof}
		Indeed, we have, for some instant $t$:
		\begin{align*}
			&D_{KL}\left( Q_t(\theta,\phi) ~ | |~ P(\theta) \right)\\
			%		&= D_{KL}\left( Q_t(\theta,\phi) ~ | |~ p_{Z}(\cdot~;\theta) \right) - \underset{Z \sim Q_t(\theta,\phi)}{\mathbb{E}}\bigg[\log p_{X,Y|Z} (x, y |Z;\theta)\bigg]\\
			%		&= D_{KL}\left( q_{Z_{1:t}|X_{1:t}}(\cdot| x_{1:t};\phi) ~.~ p_{Z_{t+1:T}|Z_{t}}(\cdot|\cdot~;\theta) ~\bigg |\bigg |~ p_{Z_{1:t}}(\cdot~;\theta) ~.~ p_{Z_{t+1:T}|Z_{t}}(\cdot|\cdot~;\theta) \right)\\
			%		&\qquad- \underset{Z \sim Q_t(\theta,\phi)}{\mathbb{E}}\bigg[\log p_{X|Z} (x |Z;\theta)\bigg] - \underset{Z \sim Q_t(\theta,\phi)}{\mathbb{E}}\bigg[\log p_{Y|X,Z} (y |x,Z;\theta)\bigg]\\
			&= \underset{Z \sim Q_t(\theta,\phi)}{\mathbb{E}}\left[ \log q_{Z_{1:t}|X_{1:t}}(\cdot| x_{1:t};\phi) + \log p_{Z_{t+1:T}|Z_{t}}(\cdot|\cdot~;\theta) \right.\\
			&\qquad\left.- \log p_{Z_{1:t}} (Z_{1:t};\theta) - \log p_{Z_{t+1:T}|Z_{t}}(Z_{t+1:T}|Z_{t};\theta) \right.\\
			&\qquad\left. - \log p_{X,Y|Z} (x, y |Z;\theta) \right]\\
			&= \underset{Z \sim Q_t(\theta,\phi)}{\mathbb{E}}\left[ \log q_{Z_{1:t}|X_{1:t}}(\cdot| x_{1:t};\phi) \right.\\
			&\qquad\left.- \log p_{Z_{1:t}} (Z_{1:t};\theta)  \right.\\
			&\qquad\left. - \log p_{X|Z} (x |Z;\theta) - \log p_{Y|X,Z} (y |x,Z;\theta) \right]\\
			&= \underset{Z \sim Q_t(\theta,\phi)}{\mathbb{E}}\left[ \log q_{Z_{1:t}|X_{1:t}}(\cdot| x_{1:t};\phi) \right.\\
			&\qquad\left.- \log p_{Z_{1:t}} (Z_{1:t};\theta) - \log p_{X_{1:t}|Z_{1:t}} (x_{1:t} |Z_{1:t};\theta) \right.\\
			&\qquad\left. - \log p_{X_{t+1:T}|Z_{t+1:T}} (x_{t+1:T} |Z_{t+1:T};\theta) \right.\\
			&\qquad\left.- \log p_{Y|X,Z} (y |x,Z;\theta) \right]\\
			&= \underset{Z \sim Q_t(\theta,\phi)}{\mathbb{E}}\left[ \log q_{Z_{1:t}|X_{1:t}}(\cdot| x_{1:t};\phi) \right.\\
			&\qquad\qquad\qquad\left.- \log p_{X_{1:t},Z_{1:t}} (X_{1:t},Z_{1:t};\theta) \right.\\
			&\qquad\qquad\qquad\left. - \log p_{X_{t+1:T}|Z_{t+1:T}} (x_{t+1:T} |Z_{t+1:T};\theta) \right.\\
			&\qquad\qquad\qquad\left.- \log p_{Y|X,Z} (y |x,Z;\theta) \right]
			%		&= \underset{Z_{1:t} \sim Q_t(\theta,\phi)}{\mathbb{E}}\left[\log q_{Z_{1:t}|X_{1:t}}(\cdot| x_{1:t};\phi) - \log p_{Z_{1:t}} (\cdot~;\theta) - \log p_{X|Z} (x_{1:t} |Z_{1:t};\theta)\right]\\
			%		&\qquad- \underset{Z \sim Q_t(\theta,\phi)}{\mathbb{E}}\bigg[\log p_{X|X,Z} (x_{t+1:T} |x_{1:t}, Z;\theta)\bigg] - \underset{Z \sim Q_t(\theta,\phi)}{\mathbb{E}}\bigg[\log p_{Y|X,Z} (y |x,Z;\theta)\bigg]\\
			%		&= D_{KL}\left(q_{Z_{1:t_1}|X_{1:t_1}}(\cdot| x_{1:t_1};\phi) ~\bigg|\bigg|~ p_{X_{1:t_1},Z_{1:t_1}} (x_{1:t_1}, \cdot~;\theta)\right)\\
			%		&\qquad- \underset{Z \sim q_{Z|X}(\cdot|x_{1:t_1};\theta, \phi)}{\mathbb{E}}\bigg[\log p_{X|X,Z} (x_{t_1+1:T} |x_{1:t_1}, Z;\theta)\bigg] - \underset{Z \sim q_{Z|X}(\cdot|x_{1:t_1};\theta, \phi)}{\mathbb{E}}\bigg[\log p_{Y|X,Z} (y |x,Z;\theta)\bigg]
			%&=D_{KL}\left(q_{Z_{1:t_1}|X_{1:t_1}}(\cdot| \cdot~;\phi) ~\bigg|\bigg|~ p_{Z_{1:t_1}} (\cdot~;\theta)\right) - \underset{Z \sim q_{Z|X}(\cdot|~x_{1:t_1};\theta, \phi)}{\mathbb{E}}\bigg[\log p_{X,Y|Z} (x, y |Z;\theta)\bigg]\\
			%&=D_{KL}\left(q_{Z_{1:t_1}|X_{1:t_1}}(\cdot| \cdot~;\phi) ~\bigg|\bigg|~ p_{X_{1:t_1},Z_{1:t_1}} (x_{1:t_1}, \cdot~;\theta)\right) - \underset{Z \sim q_{Z|X}(\cdot|~x_{1:t_1};\theta, \phi)}{\mathbb{E}}\bigg[\log p_{X,Y|Z} (x, y |Z;\theta)\bigg]
		\end{align*}
		Suppose that both $p_{X_{t+1:T}|Z_{t+1:T}} (x_{t+1:T} |Z_{t+1:T};\theta)$ and $\\p_{Y|X,Z} (y |x,Z;\theta)$ range in $[0, 1]$. This can be easily verified if they can be written as a factorization of probability density functions that each ranges in $[0, 1]$, e.g. Gaussian distributions with diagonal covariance matrices where each term of the diagonal is in $\big[\frac{1}{2\pi}, +\infty\big)$. Then, both $-\log p_{X_{t+1:T}|Z_{t+1:T}} (x_{t+1:T} |Z_{t+1:T};\theta)$ and $\\-\log p_{Y|X,Z} (y |x,Z;\theta)$ are nonnegative, i.e.
		%	Recall that for any $t \in [1:T]$, we have $p_{X_i|Z_i}(\cdot | z_{t}; \theta) = \mathcal{N}(\mu_{x}(z_{t};\theta),~ \alpha . I_{|X_t|})$
		%	and $p_{Y_i|Z_i}(\cdot | z_{t}; \theta) = \mathcal{N}(\mu_{y}(z_{t}; \theta),~ \alpha . I_{|X_t|})$. This means that for any $\alpha \in [\frac{1}{2\pi}, +\infty)$, both $p_{X_i|Z_i}(\cdot | z_{t}; \theta) $ and $p_{Y_i|Z_i}(\cdot | z_{t}; \theta) $ are nonpositive, i.e. 
		\begin{align*}
			&D_{KL}\left( Q_t(\theta,\phi) ~ | |~ P(\theta) \right)\\
			&\quad\geq D_{KL}\left(q_{Z_{1:t}|X_{1:t}}(\cdot| x_{1:t};\phi) ~||~ p_{X_{1:t},Z_{1:t}} (x_{1:t}, \cdot~;\theta)\right).
		\end{align*}
		Thus, by minimizing $D_{KL}\left( Q_t(\theta,\phi) ~ | |~ P(\theta) \right)$, we minimize an upper bound of $\\D_{KL}\left(q_{Z_{1:t}|X_{1:t}}(\cdot| x_{1:t};\phi) ~||~ p_{X_{1:t},Z_{1:t}} (x_{1:t}, \cdot~;\theta)\right)$. 
		
		Furthermore, since we have
		\begin{align*}
			&D_{KL}\left(q_{Z_{1:t}|X_{1:t}}(\cdot| x_{1:t};\phi) ~||~ p_{X_{1:t},Z_{1:t}} (x_{1:t}, \cdot~;\theta)\right)\\
			&= D_{KL}\left( q_{Z_{1:t}|X_{1:t}}(\cdot|~x_{1:t};\phi) ~||~ p_{Z_{1:t}|X_{1:t}}(\cdot|~x_{1:t};\theta) \right)\\
			&\qquad\qquad\qquad\qquad\qquad\qquad- \log p_{X_{1:t}}(x_{1:t};\theta)\\
			&= D_{KL}\left( Q_t(\theta,\phi) ~||~ P_t(\theta) \right)- \log p_{X_{1:t}}(x_{1:t};\theta),
		\end{align*}
		we know that by optimizing $\phi$ to minimize $\\D_{KL}\left(q_{Z_{1:t}|X_{1:t}}(\cdot| x_{1:t};\phi) ~||~ p_{X_{1:t},Z_{1:t}} (x_{1:t}, \cdot~;\theta)\right)$, we minimize $D_{KL}\left( Q_t(\theta,\phi) ~||~ P_t(\theta) \right)$. To sum up, minimizing $\\D_{KL}\left( Q_t(\theta,\phi) ~ | |~ P(\theta) \right)$ w.r.t. $\phi$ minimizes an upper bound of $D_{KL}\left( Q_t(\theta,\phi) ~ | |~ P_t(\theta) \right)$.
	\end{proof}
	
	\subsection{Maximization of $p_{X,Y}(x, y;\theta)$ 
		%	and minimization of $\underset{t \sim~ \mathcal{U}_{[t_{\text{min}}, ~T-1]}}{\mathbb{E}}\big[ D_{KL}\big( Q_t(\theta, \phi) ~\big |\big |~ \frac{P(\theta)}{p_{X,Y}(x,y;\theta)} \big) \big]$
	}
	
	\begin{proof}
		Replacing $P_t(\theta)$ by $Q_t(\theta, \phi)$ in Eq. (\ref{UB}), we get:
		\begin{align*}
			&\underset{t \sim~ \mathcal{U}_{[t_{\text{min}}, ~T-1]}}{\mathbb{E}}\left[ D_{KL}\left( Q_t(\theta, \phi) ~ | |~ P(\theta) \right) \right]\nonumber\\
			&= -\log p_{X,Y}(x, y;\theta) \\
			&\qquad+ \underset{t \sim~ \mathcal{U}_{[t_{\text{min}}, ~T-1]}}{\mathbb{E}}\left[ D_{KL}\left( Q_t(\theta, \phi) ~\bigg |\bigg |~ \frac{P(\theta)}{p_{X,Y}(x,y;\theta)} \right) \right]\nonumber\\
			&\geq -\log p_{X,Y}(x, y;\theta)
		\end{align*}
		Therefore, by optimizing $\phi$ to minimize $\\\underset{t \sim~ \mathcal{U}_{[t_{\text{min}}, ~T-1]}}{\mathbb{E}}\left[ D_{KL}\left( Q_t(\theta, \phi) ~ | |~ P(\theta) \right) \right]$, we minimize $\\\underset{t \sim~ \mathcal{U}_{[t_{\text{min}}, ~T-1]}}{\mathbb{E}}\left[ D_{KL}\left( Q_t(\theta, \phi) ~\bigg |\bigg |~ \frac{P(\theta)}{p_{X,Y}(x,y;\theta)} \right) \right]$, and by optimizing $\theta$ to minimize $\underset{t \sim~ \mathcal{U}_{[t_{\text{min}}, ~T-1]}}{\mathbb{E}}\left[ D_{KL}\left( Q_t(\theta, \phi) ~ | |~ P(\theta) \right) \right]$, we maximize a lower bound of $p_{X,Y}(x, y;\theta)$.
	\end{proof}
\end{appendices}

%\bibliography{../CSI_scientifique-v2}{}
\bibliography{./complexity-reduction}
\bibliographystyle{ieeetr}

\printcredits

%%% Loading bibliography style file
%%\bibliographystyle{model1-num-names}
%\bibliographystyle{cas-model2-names}
%
%% Loading bibliography database
%\bibliography{}

%% Biography


\end{document}